\documentclass{article}
\PassOptionsToPackage{numbers, compress}{natbib}
\usepackage[preprint]{neurips_2026}

\usepackage{xspace}
\usepackage{arydshln}   % Provides \dashvline for broken/dashed lines (must load after array)
\usepackage{array}      % Required for advanced tabular features

\newcommand{\paper}{\textbf{SpheRoPE}\xspace}

\usepackage[utf8]{inputenc} % allow utf-8 input
\usepackage[T1]{fontenc}    % use 8-bit T1 fonts
\usepackage{hyperref}       % hyperlinks
\usepackage{url}            % simple URL typesetting
\usepackage{booktabs}       % professional-quality tables
\usepackage{amsfonts}       % blackboard math symbols
\usepackage{nicefrac}       % compact symbols for 1/2, etc.
\usepackage{microtype}      % microtypography
\usepackage{amsmath}
\usepackage{graphicx}
\usepackage{multirow}

\usepackage[table]{xcolor}
\usepackage[export]{adjustbox} 

\title{SpheRoPE: Zero-Shot Optimization-Free 360\boldmath$^\circ$\unboldmath ~Panorama Generation with Spherical RoPE}

\author{
Or Hirschorn$^{1,2}$,
Aaron Olender$^{1,3}$,
Eli Alshan$^1$,
Ianir Ideses$^1$,
Lior Fritz$^{*1}$,
Sagie Benaim$^{*1,3}$\\
\\
$^1$Amazon Prime Video
\quad \quad
$^2$Tel-Aviv University\\
$^3$Hebrew University of Jerusalem\\
\\
\texttt{\url{https://orhir.github.io/SpheRoPE}}
}

\begin{document}

\maketitle

\def\thefootnote{$*$}\footnotetext{Denotes equal advisory}

\begin{abstract}
We present a zero-shot, training-free and optimization-free framework for generating 360$^\circ$ panoramic images and videos by directly injecting spherical priors into pre-trained diffusion transformers. Existing methods either rely on costly fine-tuning on scarce panoramic data that limits generalization, or leverage multi-step optimization that incurs prohibitive inference latency. 
We observe that contemporary generative models natively exhibit some panoramic priors from large-scale training. 
However, these emergent capabilities are insufficient, as the models fundamentally fail to satisfy the rigorous topological constraints imposed by equirectangular projection (ERP).
We introduce a zero-shot and optimization-free approach that resolves these constraints at inference time. \textbf{Spherical RoPE} replaces standard rotary position embeddings: low-frequency channels are re-parameterized as 3D Cartesian coordinates to natively encode the spherical manifold, while high-frequency channels are harmonically quantized to enforce exact $2\pi$ periodicity. 
Coupled with complementary Semantic Distortion classifier-free guidance (CFG) that explicitly steers geometry, 
we avoid retraining and inherit the full creative breadth of state-of-the-art models. 
Our approach generalizes across diverse backbones and 360\boldmath$^\circ$\unboldmath generation modalities. We demonstrate this across text-to-panorama using Flux.1, Flux.2, and LTX-Video backbones, achieving competitive performance against baselines, all while remaining training-free. 
% Our code is publicly available.
% Furthermore, we achieve the first zero-shot $360^\circ$ audio-to-video generation.
\end{abstract}

\begin{figure*}[t]
\centering
\setlength{\tabcolsep}{1pt}
\newlength{\cropw}\setlength{\cropw}{0.099\textwidth}
\newlength{\panow}\setlength{\panow}{0.297\textwidth}

% === Image Generation ===
\noindent
\rotatebox[origin=c]{90}{\small\textbf{Image Generation}}%
\hspace{8pt}%
\begin{minipage}[c]{0.95\textwidth}
\begin{tabular}{@{}r@{\hspace{10pt}}ccc@{}}
\smash{\rotatebox[origin=c]{90}{\small\textit{FLUX.1}}} &
\includegraphics[width=\panow]{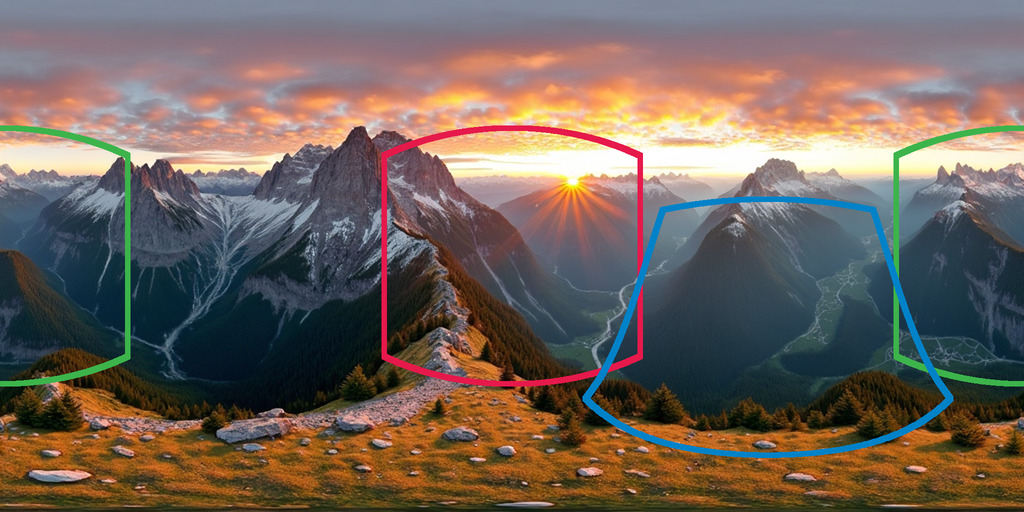} &
\includegraphics[width=\panow]{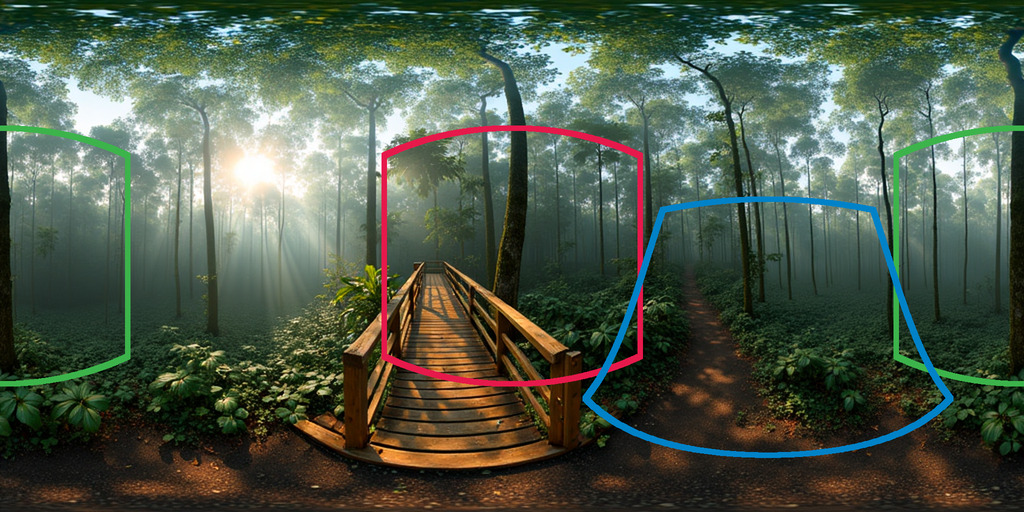} &
\includegraphics[width=\panow]{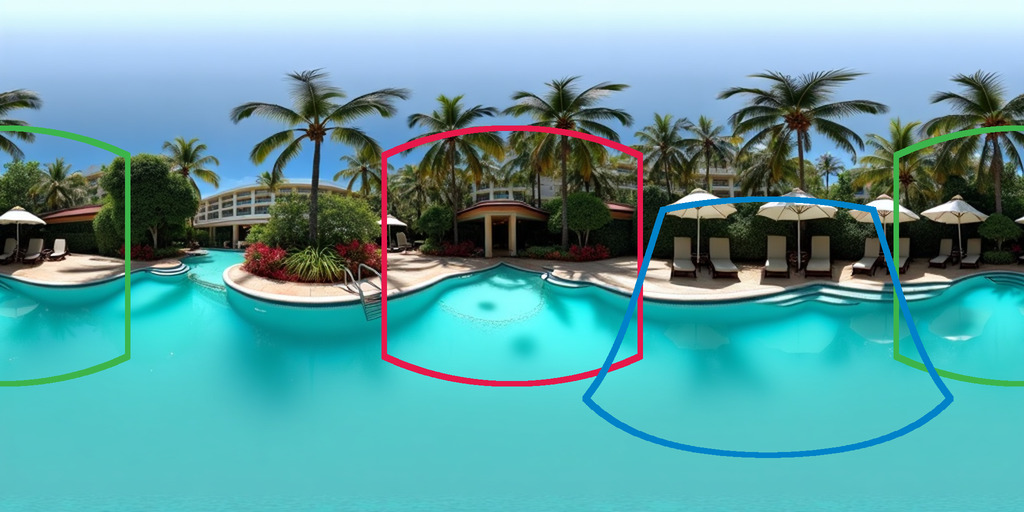} \\[1pt]
&
\includegraphics[width=\cropw]{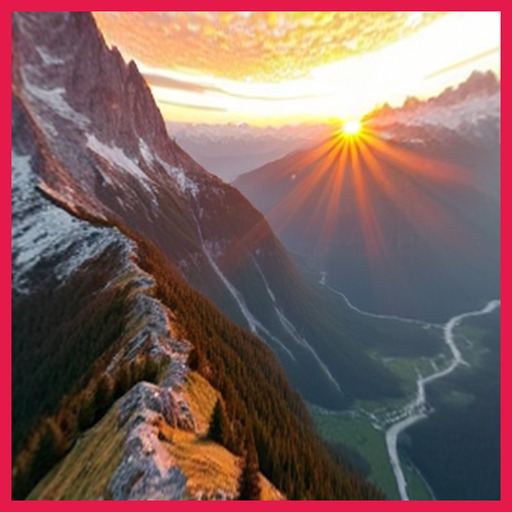}\includegraphics[width=\cropw]{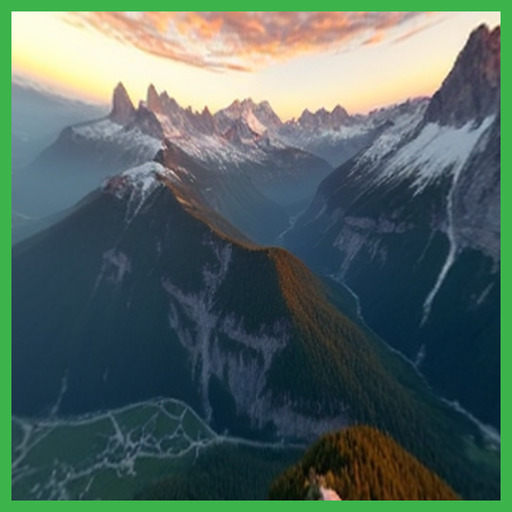}\includegraphics[width=\cropw]{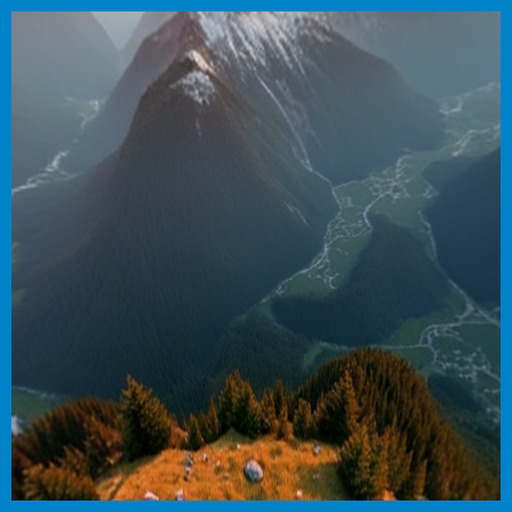} &
\includegraphics[width=\cropw]{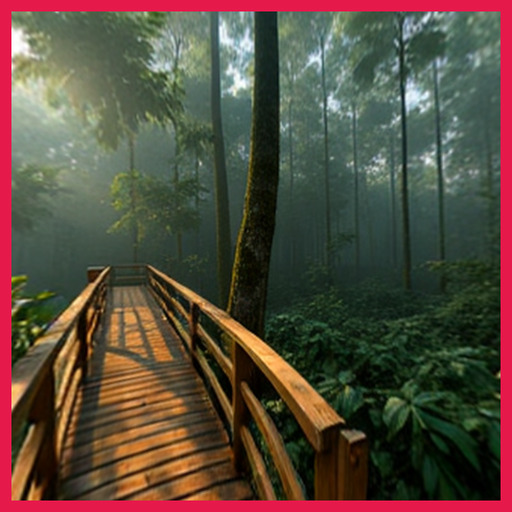}\includegraphics[width=\cropw]{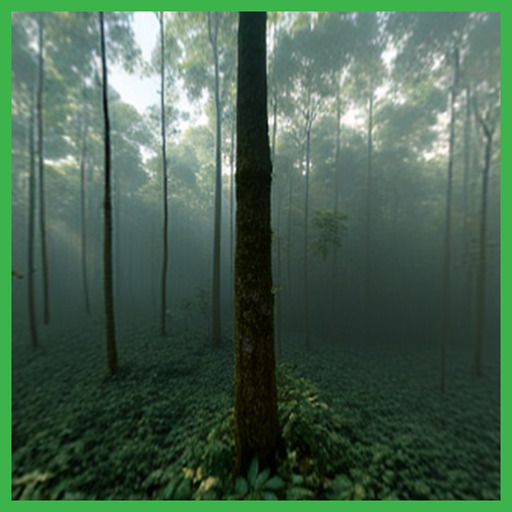}\includegraphics[width=\cropw]{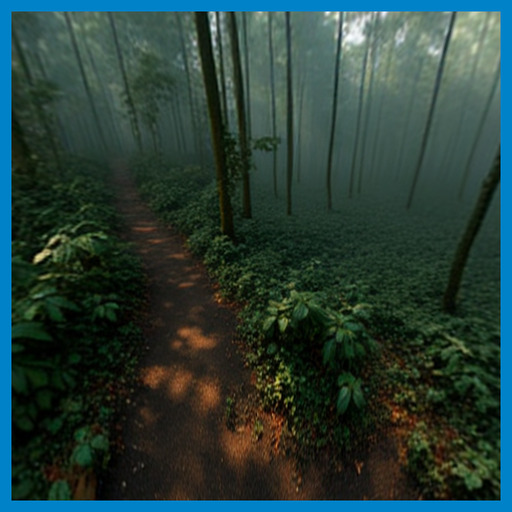} &
\includegraphics[width=\cropw]{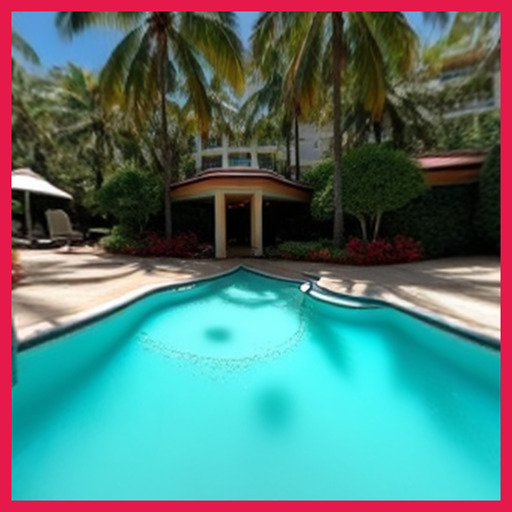}\includegraphics[width=\cropw]{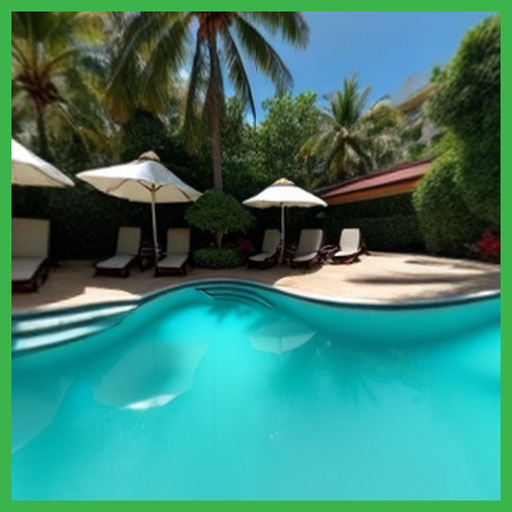}\includegraphics[width=\cropw]{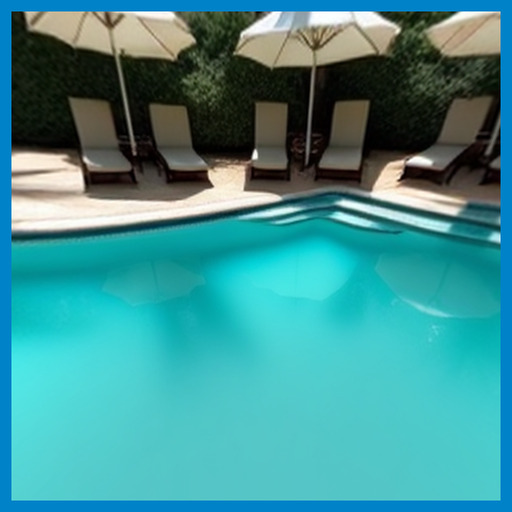} \\[3pt]
\smash{\rotatebox[origin=c]{90}{\small\textit{FLUX.2}}} &
\includegraphics[width=\panow]{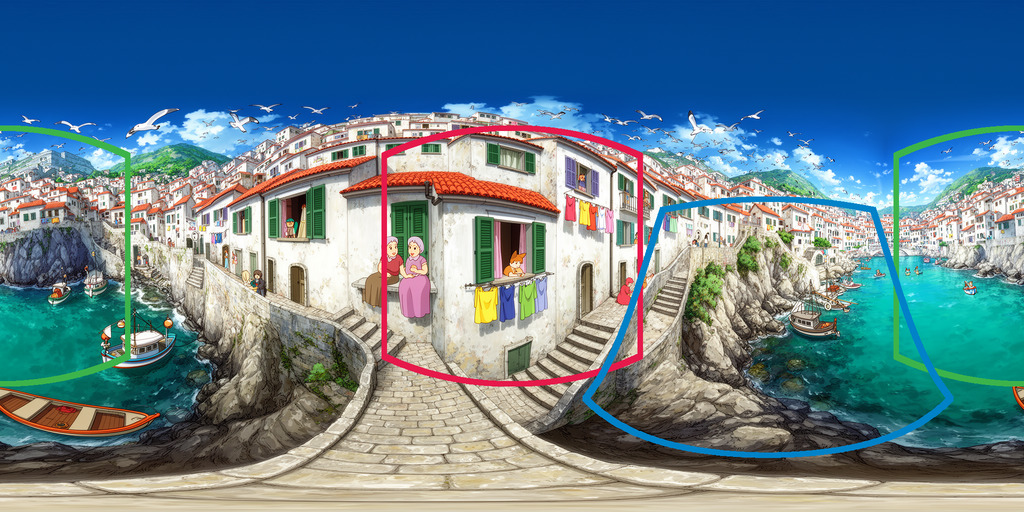} &
\includegraphics[width=\panow]{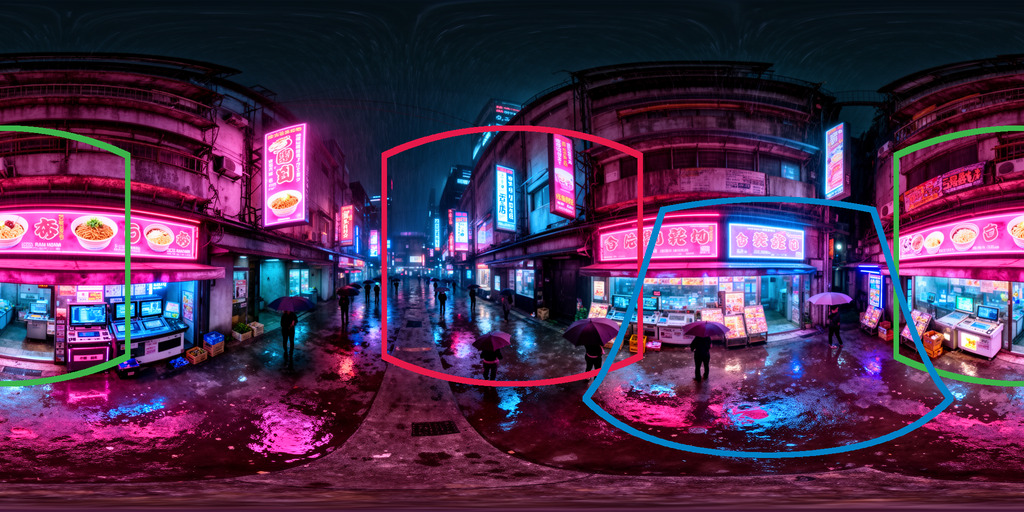} &
\includegraphics[width=\panow]{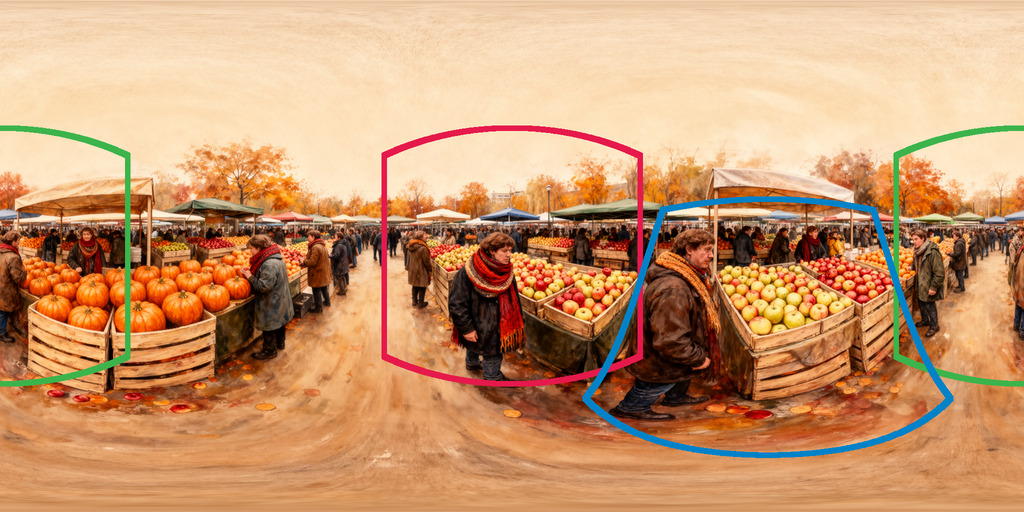} \\[1pt]
&
\includegraphics[width=\cropw]{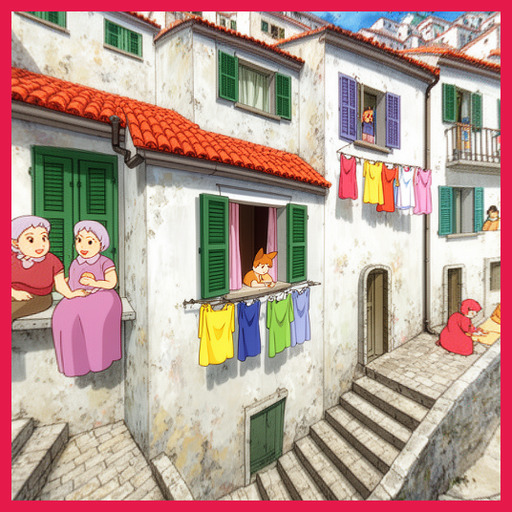}\includegraphics[width=\cropw]{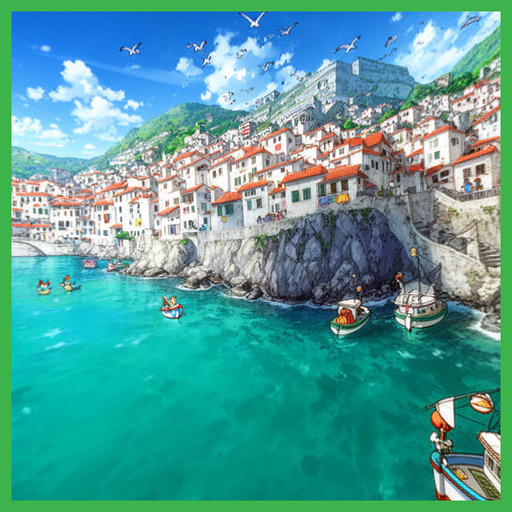}\includegraphics[width=\cropw]{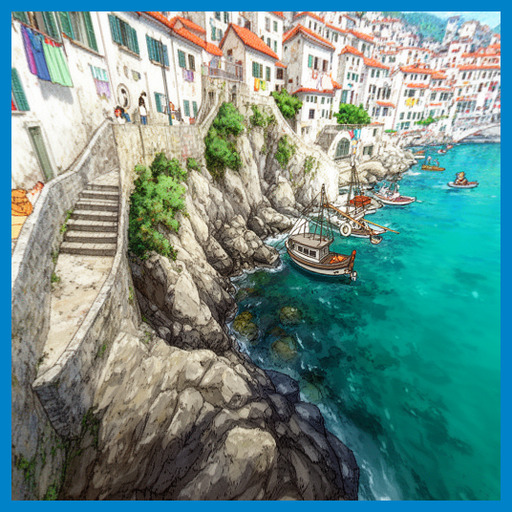} &
\includegraphics[width=\cropw]{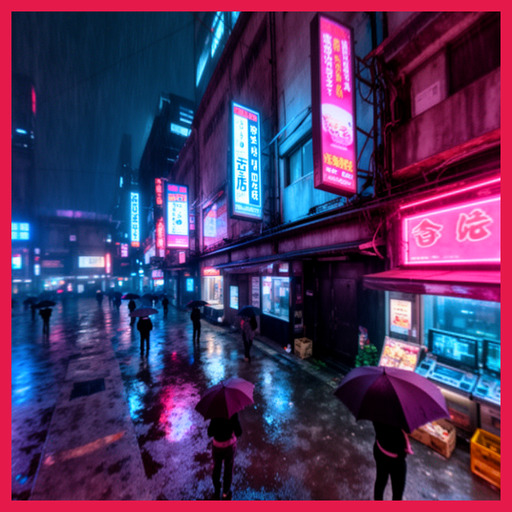}\includegraphics[width=\cropw]{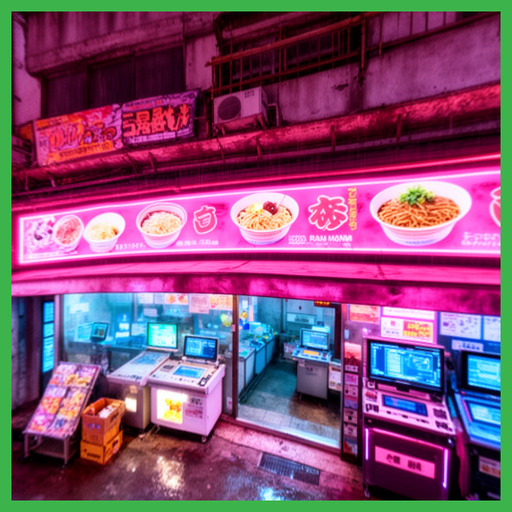}\includegraphics[width=\cropw]{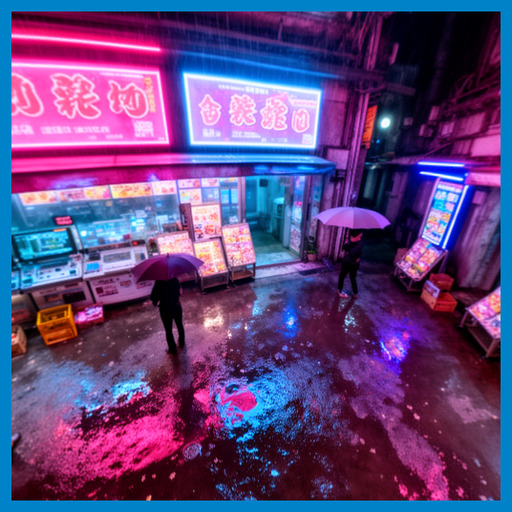} &
\includegraphics[width=\cropw]{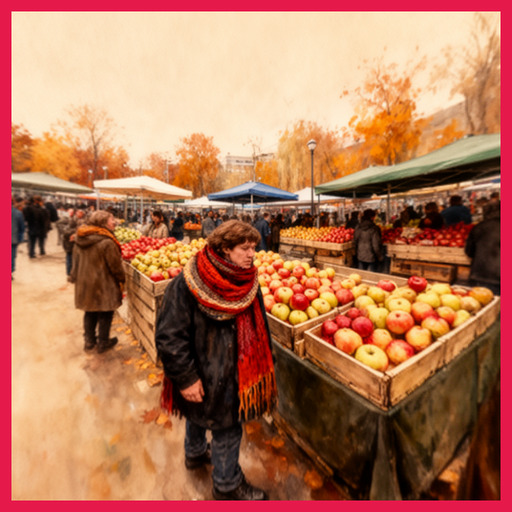}\includegraphics[width=\cropw]{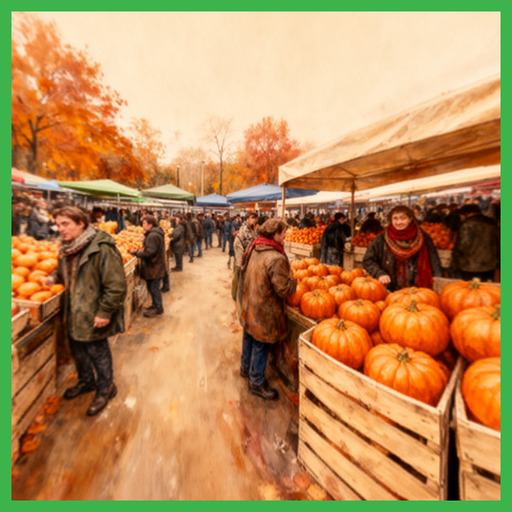}\includegraphics[width=\cropw]{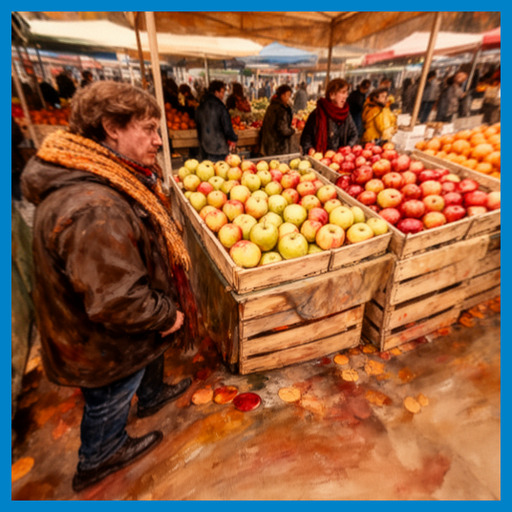} \\
\end{tabular}
\end{minipage}

\vspace{1pt}\hrule\vspace{1pt}

% === Video Generation ===
\noindent
\rotatebox[origin=c]{90}{\small\textbf{Video-Audio Generation}}%
\hspace{8pt}%
\begin{minipage}[c]{0.95\textwidth}
\begin{tabular}{@{}r@{\hspace{10pt}}ccc@{}}
& {\small Frame 1} & {\small Frame 120} & {\small Frame 241} \\[1pt]
\smash{\rotatebox[origin=c]{90}{\small\textit{LTX 2.3}}} &
\includegraphics[width=\panow]{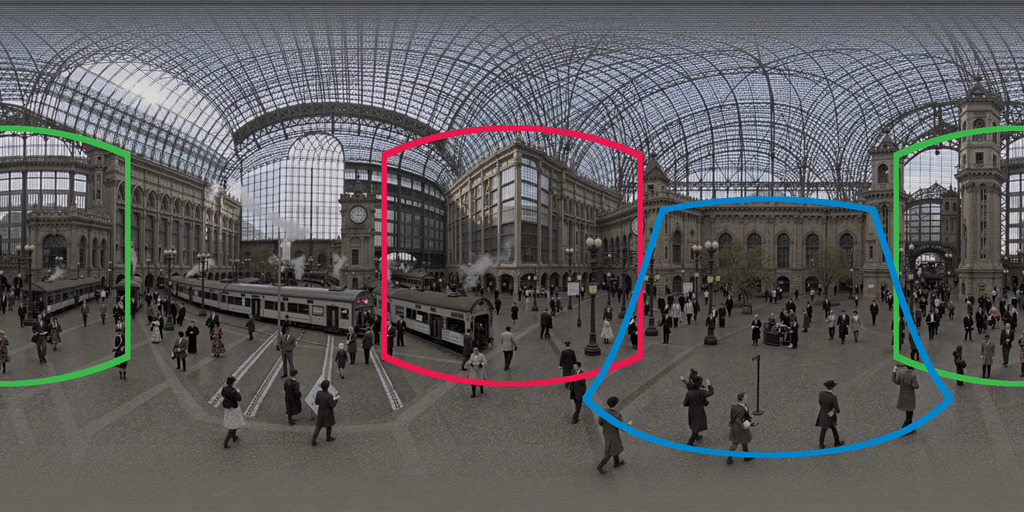} &
\includegraphics[width=\panow]{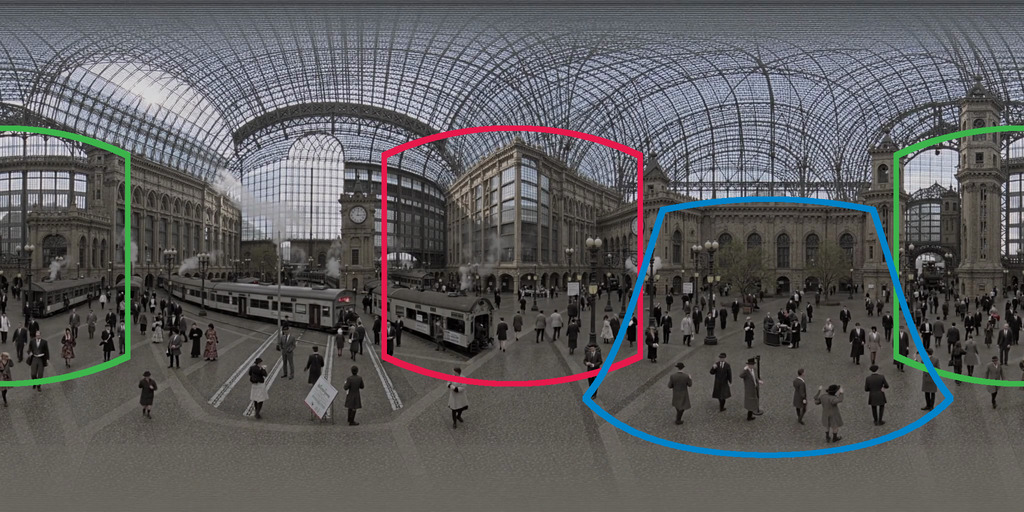} &
\includegraphics[width=\panow]{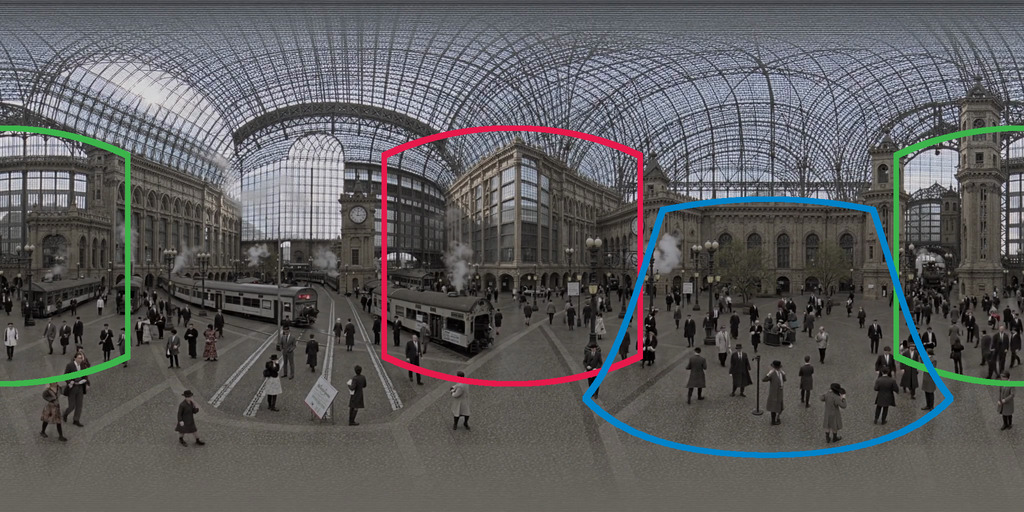} \\[1pt]
&
\includegraphics[width=\cropw]{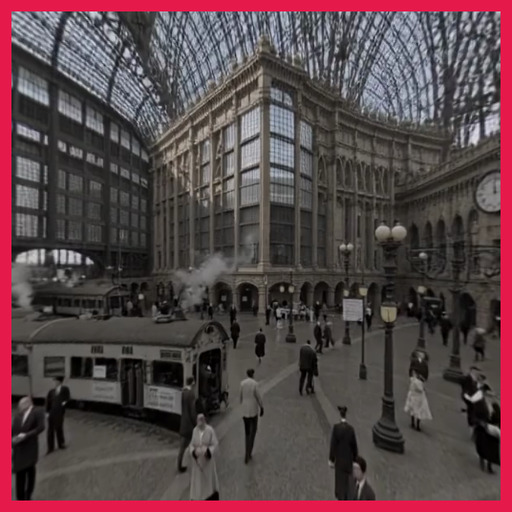}\includegraphics[width=\cropw]{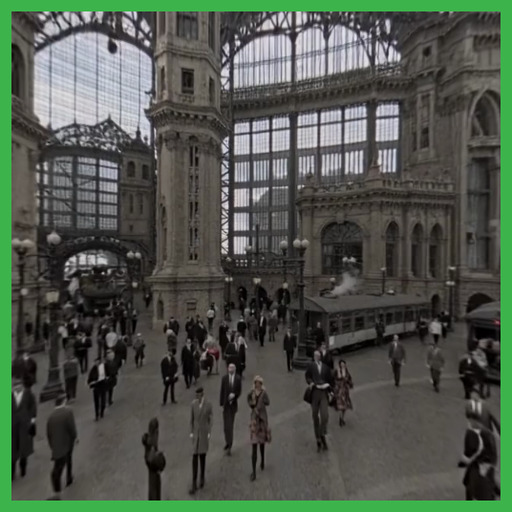}\includegraphics[width=\cropw]{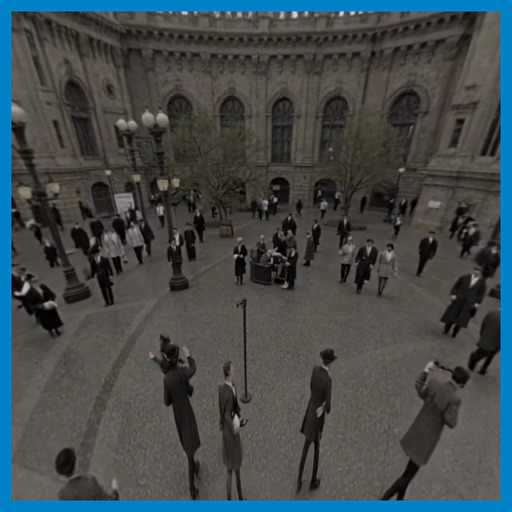} &
\includegraphics[width=\cropw]{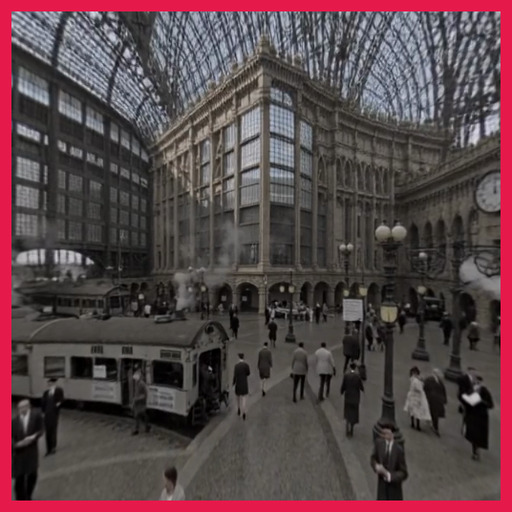}\includegraphics[width=\cropw]{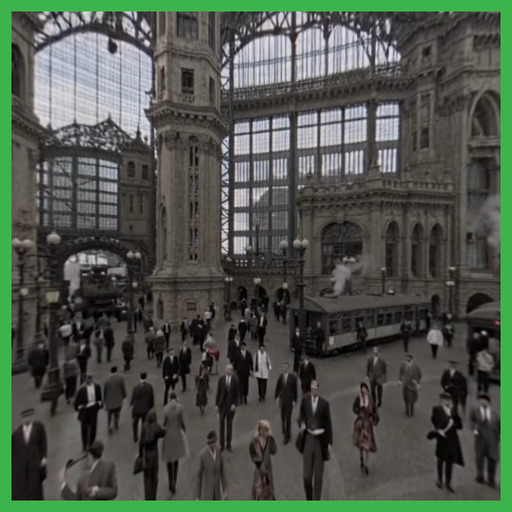}\includegraphics[width=\cropw]{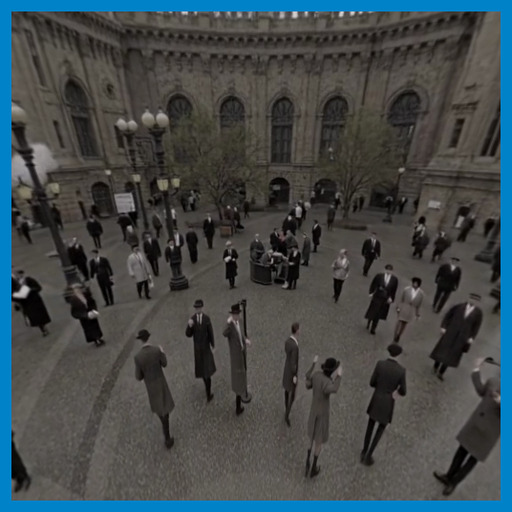} &
\includegraphics[width=\cropw]{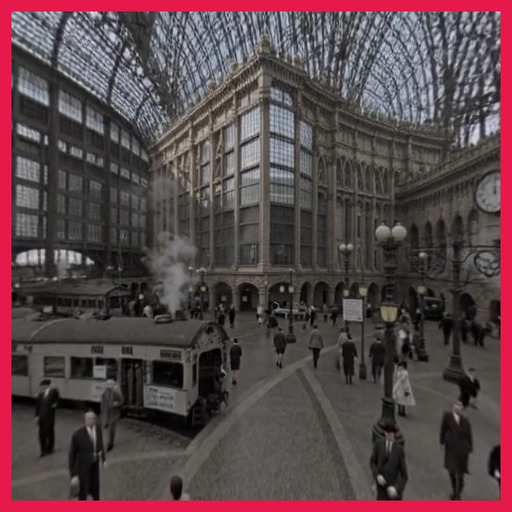}\includegraphics[width=\cropw]{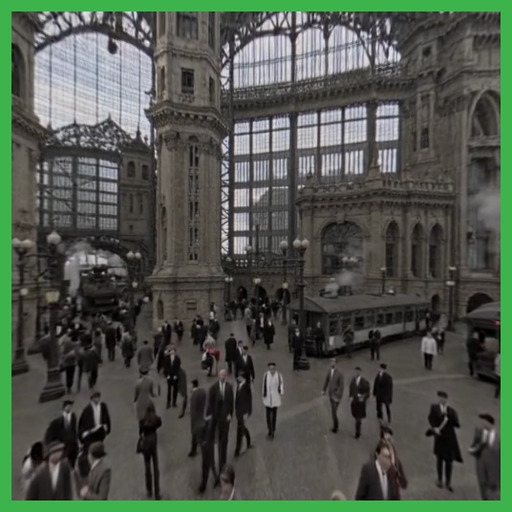}\includegraphics[width=\cropw]{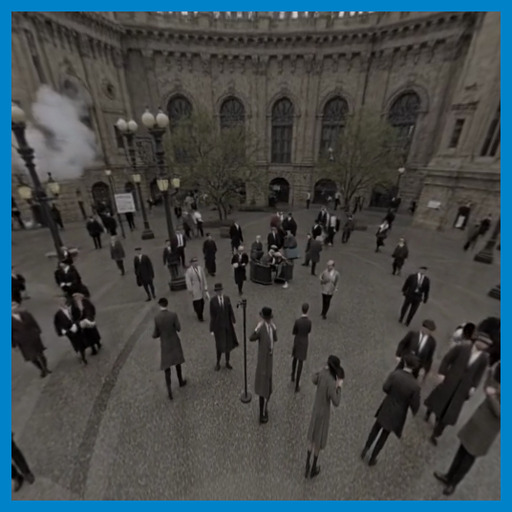} \\
\end{tabular}
\end{minipage}
\caption{\textbf{Zero-shot 360\boldmath$^\circ$\unboldmath Generation.} Our inference-only approach enables seamless text-to-panorama synthesis across image (FLUX.1, FLUX.2) and audio-video (LTX~2.3) backbones. Each section displays the generated ERP panorama (top) and perspective crops re-projected from the highlighted regions (bottom), demonstrating strict geometric consistency.}
% \caption{\textbf{Zero-shot 360\boldmath$^\circ$\unboldmath Generation.} Our method modifies only the inference-time spatial priors of pre-trained diffusion transformers, enabling seamless text-to-panorama synthetic across diverse backbones. We demonstrate image generation (FLUX.1 and FLUX.2) and audio-video generation (LTX~2.3). Perspective crops highlight geometric consistency across different field of views. }

% \sagie{Full results, as well as corresponding audio output is provided in the supplementary. }

\label{fig:teaser}
\end{figure*}

\section{Introduction}

The rapid evolution of image and video diffusion models~\cite{rombach2022ldm, saharia2022imagen, ramesh2022dalle2, ho2022video} has fundamentally shifted the landscape of content creation, enabling the synthesis of high-fidelity visual assets from natural language descriptions. While these models have achieved remarkable success in generating planar, perspective-frame images and videos, there is an increasing demand for omnidirectional generative frameworks capable of producing immersive 360$^\circ$ panoramic environments. Such content is critical for applications in virtual reality (VR)~\cite{brivio2021vr360} and robotics simulation, where a narrow field-of-view is insufficient to capture the global context of a scene~\cite{wang2025survey}. In this work we address zero-shot and optimization-free text-to-panorama generation across different modalities, including static images, dynamic videos, and integrated audio-video environments.

360$^\circ$ panoramas are typically represented using an equirectangular projection (ERP), which maps spherical imagery onto a 2D rectangular plane. Diffusion transformers (DiTs)~\cite{peebles2023dit}, as employed in recent models such as Flux~\cite{blackforestlabs2024flux} and LTX-Video~\cite{ltxvideo2025}, rely on rotary position embeddings (RoPE)~\cite{su2024roformer} and planar attention. These mechanisms inherently assume a Euclidean grid, leading to visible seams at the boundaries and geometric incoherence at the poles when associated with ERP-based generation. 
Existing research attempting to bridge this gap generally follows two paradigms. 
\textit{Training-based methods} fine-tune pre-trained models on specialized panoramic datasets, either through full fine-tuning~\cite{feng2023diffusion360}, parameter-efficient adaptation via LoRA~\cite{hu2022lora, wang2024stitchdiffusion, zhang2024panfusion, wang2024diffpano}, or specialized architectures such as spherical manifold convolutions~\cite{sun2025smgd}.
However, these approaches are constrained by the scarcity of high-quality 360$^\circ$ data.
They further suffer from the immense computational cost of retraining large-scale models, degraded generalization in out-of-distribution (OOD) scenarios~\cite{wang2025survey}, and the need to re-train whenever the backbone is updated or extended to new modalities. 
\textit{Optimization-based methods}, such as PanoFree~\cite{liu2024panofree} and SphereDiff~\cite{park2025spherediff}, avoid retraining but rely on iterative warping-inpainting pipelines or patch-based MultiDiffusion~\cite{bartal2023multidiffusion} in spherical latent spaces, introducing substantial inference-time overhead that makes them impractical for real-time or video applications.

We propose \paper, a zero-shot, training-free and optimization-free framework that generates high-fidelity ERP panoramas by realigning the internal inductive biases of pre-trained DiTs with spherical geometry. 
% Our approach builds upon the insight that pre-trained foundational models already capture some of the underlying distribution of panoramic environments, evidenced by their tendency to produce panoramic-like patterns when prompted accordingly. However, they fail at adhering to the geometric constraints imposed by ERP.
Our approach builds upon the insight that pre-trained foundational models already possess rich priors for panoramic environments. When conditioned on panoramic prompts, these models natively generate equirectangular characteristics, yet their standard architectures fail to enforce the rigorous topological constraints imposed by equirectangular projection (ERP).
To bridge this gap, we introduce \textbf{Spherical RoPE}, a frequency-aware reformulation of the standard transformer RoPE~\cite{su2024roformer}. Under this formulation, high-frequency channels are harmonically quantized to guarantee $2\pi$ horizontal periodicity, whereas low-frequency channels utilize 3D Cartesian coordinates on the unit sphere to enforce both polar convergence and periodicity.
In addition, we leverage Semantic Distortion CFG that extends standard CFG~\cite{ho2022cfg} to a three-way scheme using an anchored geometric prompt, steering the denoising process toward valid ERP projections without sacrificing semantic detail. Figure~\ref{fig:teaser} shows results of our method.

A defining advantage of our framework is its architectural generality. By isolating modifications to the positional encoding and guidance logic, our approach generalizes across modern diffusion transformers without requiring task-specific adaptation. 
We demonstrate this versatility by leveraging Flux~\cite{blackforestlabs2024flux} for high-fidelity static environments and LTX~2.3~\cite{ltxvideo2025} for 360$^\circ$ video generation.
Furthermore, because we do not alter the underlying model weights, the framework strictly preserves the foundational capabilities of the backbone. It seamlessly inherits built-in conditioning pipelines, enabling advanced functionalities such as zero-shot image-to-panorama translation, or creating both video and audio for a scene.
Moreover, operating as a zero-shot paradigm, our method yields highly robust out-of-distribution (OOD) performance across diverse styles and visual domains.

We evaluate our method on both image and video generation tasks. For static 360$^\circ$ panorama synthesis, we benchmark on ODI-SR~\cite{Deng_2021_CVPR} against training and optimization-based methods. For video generation, we assess performance on two prompt sets using VBench~\cite{vbench2024}. Additionally, we conduct an LLM-based perceptual evaluation and a user study, showing a clear preference for the results of our method. Our zero-shot approach achieves state-of-the-art results on several metrics and competitive performance on others, all without any training, optimization, or model-specific adaptation. Finally, we validate our core design choices through extensive ablation studies.

Our contributions are summarized as follows:
\begin{itemize}
    \item We present the first zero-shot, training and optimization-free framework for seamless 360$^\circ$ image and video generation. Our approach modifies only the inference-time spatial priors of pre-trained DiTs, making it plug-and-play for diverse backbones.
    \item We introduce \textbf{Spherical RoPE} to geometrically induce the topological invariants of the sphere ($S^2$) via spectral decomposition, alongside \textbf{Semantic Distortion CFG} to amplify the native panoramic priors of the model during inference.
    % \item We introduce Spherical RoPE, a spectral decomposition strategy that mathematically enforces the topological invariants of the sphere ($S^2$).
    % \item We propose Semantic Distortion CFG, an inference strategy that enhances panoramic generation for pre-trained diffusion models.
    \item We demonstrate the efficacy of our zero-shot approach across image and video benchmarks, achieving highly competitive performance against fine-tuned baselines while consistently outperforming them in panoramic coherence and human preference.
\end{itemize}

\section{Related Work}
\label{sec:related}
\paragraph{Text-driven 360 panorama generation.}
Following early VQGAN-CLIP models~\cite{chen2022text2light}, latent diffusion~\cite{rombach2022ldm} for panorama generation split into training and optimization paradigms.
Training-based methods fine-tune on panoramic datasets via DreamBooth~\cite{ruiz2022dreambooth, feng2023diffusion360} or LoRA~\cite{ni2025makes, wang2025conditional}. To improve spatial consistency, architectures have been extended with MultiDiffusion stitching~\cite{bartal2023multidiffusion, wang2024stitchdiffusion}, dual-branch networks~\cite{zhang2024panfusion}, specialized attention (epipolar~\cite{wang2024diffpano}, cubemap~\cite{kalischek2025cubediff}), spherical convolutions~\cite{sun2025smgd}, distortion decoupling~\cite{zheng2025panodecouple}, and hybrid DiTs~\cite{feng2025dit360}. However, these remain bottlenecked by data scarcity, retraining costs, and limited out-of-distribution generalization.
Optimization-based methods avoid fine-tuning. Approaches include factor-graph inference over patches (DiffCollage~\cite{zhang2023diffcollage}), iterative warping (PanoFree~\cite{liu2024panofree}), and distortion-aware MultiDiffusion in spherical latent space (SphereDiff~\cite{park2025spherediff}). While 360PanT~\cite{wang2025360pant} tackles tiling for panorama translation, these training-free methods generally incur high latency and risk global structural inconsistencies due to their patch-based or iterative nature.
The extension to 360$^\circ$ video is rapidly emerging. Training-based models introduce adapters (360DVD~\cite{wang2024360dvd}), DiT backbones (PanoDiT~\cite{zhang2025panodit}), latitude-aware sampling (PanoWan~\cite{xia2026panowan}), dual-branch lifting (Imagine360~\cite{tan2025imagine360}), and geometry-aware conditioning (Argus~\cite{luo2025argus}). Training-free video methods include offset-shifting (DynamicScaler~\cite{liu2025dynamicscaler}) and temporal extensions of spherical latents (SphereDiff~\cite{park2025spherediff}).
To the best of our knowledge, we are the first to achieve panorama synthesis that is both training-free and optimization-free. Our approach relies solely on inference-time modifications to positional encoding and guidance, bypassing resource-intensive training and slow multi-step optimization.

\paragraph{Positional encoding adaptation in transformers.}
Our training-free approach modifies rotary position embedding (RoPE)~\cite{su2024roformer} at inference time to encode spherical geometry. Several prior works have explored modifications to RoPE structure. This includes frequency rescaling for NLP contexts~\cite{chen2023pi, peng2023ntk, peng2023yarn}, adaptations for DiT spatial/temporal axes~\cite{lu2024fit, wang2024fitv2, zhao2025riflex}, and dynamic adjustments across denoising timesteps~\cite{zhuo2024luminanext, issachar2025dype}. For non-Euclidean spaces, RoPE has been adapted for geolocation tokens in NLP~\cite{unlu2023spherical}, and classification~\cite{vandegeijn2025circular}.
Recently, IaaW~\cite{gui2025iaaw}, incorporated a spherical RoPE formulation for image-to-panoramic-video generation. Their approach maps the entire spatial domain to a 3D sphere. Consequently, this requires fine-tuning to accommodate the shift in positional structure.
In contrast, our \textbf{SpheRoPE} encodes positions within the 2D ERP image plane while respecting its underlying spherical topology. By keeping the vertical RoPE axis unchanged and modifying only the horizontal axis, we preserve the 2D structure expected by pre-trained models. This, combined with our spectral partitioning (where spherical encoding is restricted to low-frequency channels), enables true zero-shot generation without retraining.

\section{Method}

% \sagie{give a quick overview of the method. We start with XXX in Sec. XXX. Then we...}
We present a zero-shot optimization-free framework for generating ERP panoramas from pre-trained diffusion transformers, requiring no fine-tuning or architectural changes. 
We begin with our problem formulation and setting in Section~\ref{sec:formulation}. We then introduce SpheRoPE, our spherical rotary position embedding that encodes horizontal periodicity and polar convergence directly into the attention mechanism in Section~\ref{sec:spherope}. Finally, we describe Semantic Distortion CFG, a dual-guidance scheme that steers generation toward geometrically valid panoramic structure in Section~\ref{sec:sdcfg}.
In the supplementary, Section~\ref{supp:preliminary}, we provide detailed formulation of CFG and RoPE.

% \subsection{Preliminaries}
% \paragraph{Rotary Position Embedding.} Diffusion transformers model spatial dependencies using Rotary Position Embedding (RoPE), which encodes relative positions directly into query-key interactions. For 2D inputs, RoPE operates axially. It rotates pairs of hidden dimensions for a spatial coordinate (e.g., column $c$) by an angle $\alpha_i(c) = c \cdot \omega_i$, where $\omega_i$ defines a geometric series of frequencies. These block-diagonal rotations are applied to queries and keys prior to the attention inner product.
% We provide further details in section~\ref{supp:preliminary}

\subsection{Problem Formulation}
\label{sec:formulation}

% To leverage 2D image priors for $360^\circ$ scene generation, spherical content must be projected from 2D manifolds, primarily via cubemaps or equirectangular projection (ERP)~\cite{wang2025survey}. 
% Cubemaps minimize distortion by distributing the sphere across six planar faces, but introduce edge discontinuities that disrupt global continuity, requiring specialized inter-face attention mechanisms~\cite{kalischek2025cubediff}. In contrast, ERP maps the sphere to a single rectangular plane parameterized by longitude $\phi$ and latitude $\theta$. While ERP is the industry standard for panoramic storage and display, it imposes severe geometric demands on the generative process.
% Specifically, a valid ERP panorama must satisfy two rigorous topological constraints that standard diffusion models are not designed to handle. First, \textit{horizontal periodicity} requires that the left and right boundaries be perfectly continuous, as they represent the same longitudinal meridian. Second, \textit{polar convergence} necessitates that all columns converge to a single point at the poles, creating a coordinate singularity.
To leverage 2D priors for 360$^\circ$ generation, spherical content is typically projected via ERP~\cite{wang2025survey}. Although standard for panoramic formats, ERP requires mapping a sphere to a 2D plane, introducing rigorous geometric demands. Standard diffusion models struggle with two core topological invariants imposed by ERP: \textit{horizontal periodicity} (perfect continuity between the left and right edges) and \textit{polar convergence} (a coordinate singularity where all columns meet at the poles).
% Third, the projection introduces \textit{non-uniform spatial density}, where the resolution per solid angle increases dramatically toward the poles. 

Formally, let $\mathbf{x} \in \mathbb{R}^{H \times W \times C}$ denote an ERP panoramic image (or video frame) with height $H$, width $W = 2H$, and $C$ channels. The ERP maps spherical coordinates $(\theta, \phi)$, latitude $\theta \in [-\pi/2, \pi/2]$ and longitude $\phi \in [-\pi, \pi)$, to pixel coordinates $(r, c)$ via
\begin{equation}
    r = \frac{\theta + \pi/2}{\pi} \cdot (H - 1), \quad c = \frac{\phi + \pi}{2\pi} \cdot W
\end{equation}
This mapping introduces two geometric constraints that any valid ERP must satisfy:
\begin{itemize}
    \item[\textbf{C1.}] \textbf{Horizontal periodicity}: $\mathbf{x}[r, c] = \mathbf{x}[r, c \bmod W]$ for all rows $r$, since longitudes $-\pi$ and $\pi$ correspond to the same meridian.
    \item[\textbf{C2.}] \textbf{Polar convergence}: $\mathbf{x}[0, c_1] = \mathbf{x}[0, c_2]$ and $\mathbf{x}[H{-}1, c_1] = \mathbf{x}[H{-}1, c_2]$ for all columns $c_1, c_2$, since all longitudes converge to a single point at each pole.
\end{itemize}

Given a text prompt $\mathbf{p}$ and a pre-trained diffusion model, our goal is to generate $\mathbf{x}$ that is both semantically faithful to $\mathbf{p}$ and satisfies constraints \textbf{C1} and \textbf{C2}, without modifying any model parameters.

\subsection{SpheRoPE: Spherical RoPE}
\label{sec:spherope}
The standard linear encoding $\alpha_i(c) = c \cdot \omega_i$ for $\omega_i = \theta_{\text{base}}^{-2i/d}$ in RoPE fundamentally violates both ERP constraints. It assigns different embeddings to columns $0$ and $W$ (breaking horizontal periodicity, \textbf{C1}), and distinct embeddings to all columns at the poles (breaking polar convergence, \textbf{C2}). We propose to replace the width-axis channels of RoPE with a geometry-aware encoding, preserving the height and temporal axes identically to the original model.

A na\"ive approach would apply a uniform geometric transformation across all channels. 
However, the wide spectral range of RoPE dictates that different channels govern distinct spatial behaviors within the self-attention mechanism. This spectral diversity creates a natural functional division: high-frequency channels specialize in fine-grained local texture coherence, while low-frequency channels act as a global compass for the spatial layout. 
To avoid disrupting these specialized roles with a one-size-fits-all geometric fix, we partition the channels based on their harmonic alignment with the image width $W_{\text{tokens}}$. 
This allows us to transform the RoPE components to encode spherical geometry while preserving the unique spatial characteristics of each frequency band.
% A na\"ive approach would apply a uniform geometric transformation across all frequency channels, yet the wide spectral range of RoPE implies that different channels govern distinct spatial behaviors. 
% Within the self-attention mechanism, the score between a query $\mathbf{q}$ and key $\mathbf{k}$ at a horizontal distance $t$ is modulated by the sum of orthogonal projections $\sum_i \cos(t\omega_i)$. This spectral diversity creates a functional division where high-frequency channels, which exhibit rapid oscillations and destructive interference over distance, specialize in fine-grained local texture coherence. 
% Alternatively, low-frequency channels decay slowly and act as a global compass for the macroscopic spatial layout. 
% Because a one-size-fits-all geometric fix compromises these specialized roles, we partition the channels based on their harmonic alignment with the image width $W_{\text{tokens}}$, transforming the RoPE components to encode spherical geometry while preserving the unique spatial characteristics of each frequency band. 

\begin{figure}[t]
\centering
% \setlength{\tabcolsep}{1pt}
% \renewcommand{\arraystretch}{0.5}
% \begin{tabular}{@{}cc@{}}
\includegraphics[width=0.8\linewidth]{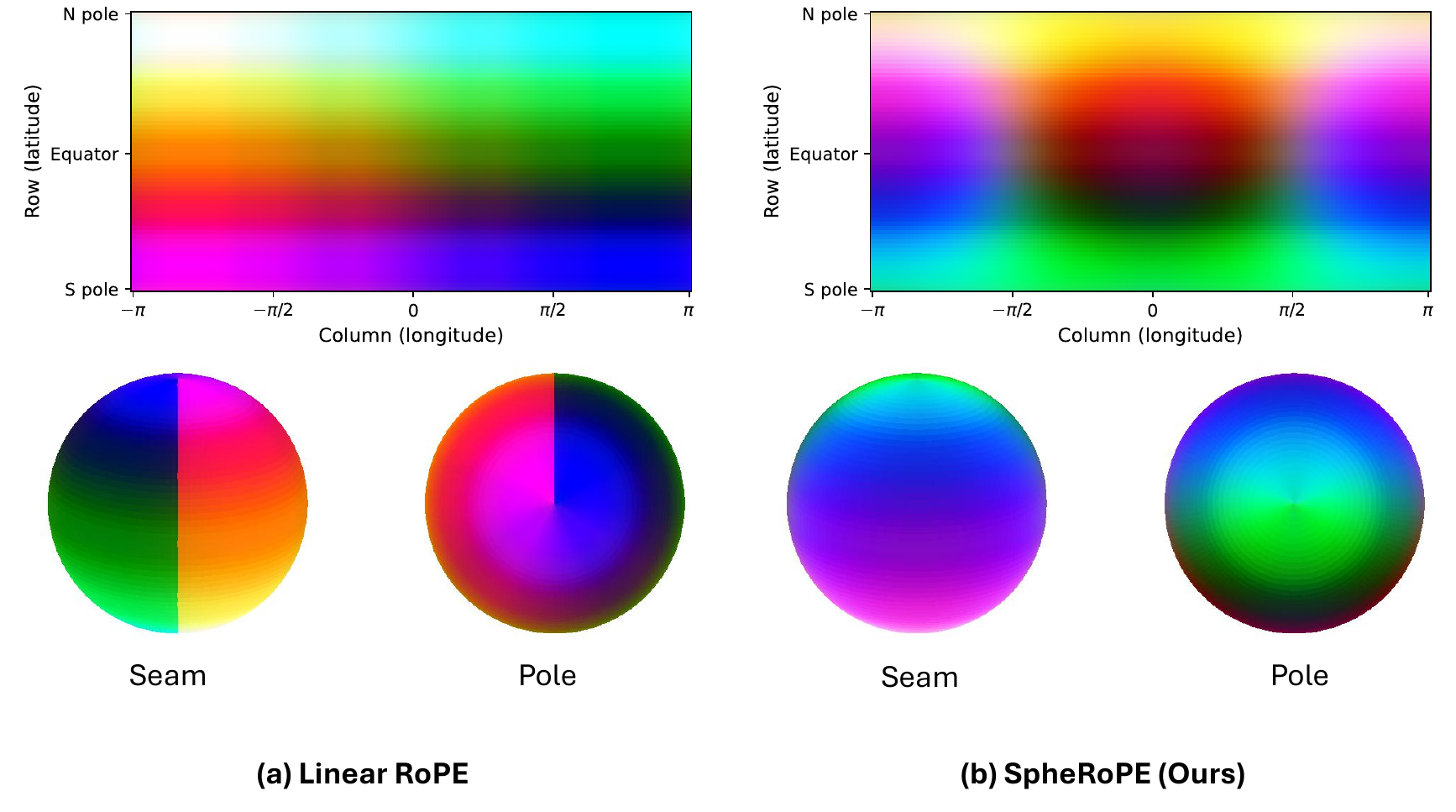}
% \includegraphics[width=0.49\linewidth]{figures/RoPE//rope_stock_pca.pdf} &
% \includegraphics[width=0.49\linewidth]{figures/RoPE//rope_spherical_pca.pdf} \\ [-5pt] 
% \includegraphics[width=0.49\linewidth]{figures/RoPE//sphere_stock.pdf} &
% \includegraphics[width=0.49\linewidth]{figures/RoPE//sphere_ours.pdf} \\
% {\small \textbf{(a)} Linear RoPE} & {\small \textbf{(b)} SpheRoPE (Ours)}
% \end{tabular}
\caption{\textbf{PCA visualization of RoPE.} (a) Linear RoPE creates a seam at the $\pm\pi$ meridian and disjoint polar embeddings. (b) Our method wraps seamlessly with uniform polar convergence.}
\label{fig:rope_vis}
\end{figure}

Specifically, for each RoPE channel $i$, we compute the ratio $k_i = \omega_i / \omega_{\text{fund}}$, where $\omega_{\text{fund}} = 2\pi / W_{\text{tokens}}$ is the fundamental frequency for horizontal wrap-around. A frequency is quantizable if it completes at least one full cycle ($k_i \ge 1$) and is near-integer ($|k_i - \text{round}(k_i)| / k_i \leq \varepsilon$). Rather than masking individual channels, we identify the first channel index $i_{\text{split}}$ that violates this condition, partitioning the spectrum into two continuous bands. For high-frequency channels ($i < i_{\text{split}}$), we enforce strict cyclic periodicity by using Cyclic Linear Encoding.
For the remaining low-frequency channels ($i \ge i_{\text{split}}$), we apply the spherical Cartesian encoding to anchor the global layout. Our solution, illustrated in Figure~\ref{fig:rope_vis}, allows the model to respect spherical topology without sacrificing high-frequency priors.

\paragraph{Spherical Cartesian Encoding (Low-Frequency Channels).}
Low-frequency channels vary slowly across the image, dictating the global layout of the scene. They cannot be made cyclic through simple quantization: because they accumulate less than one full oscillation across $W_{\text{tokens}}$, snapping them to the nearest harmonic would require a massive spectral shift, shattering the global distance metric of the model.
Instead, we abandon the linear parameterization for these channels and replace it with Cartesian coordinates on the unit sphere. For a token at row $r$ and column $c$, we compute:
\begin{equation}
\label{eq:spherical_rope}
\begin{aligned}
    \theta(r) &= \frac{r}{H_{tokens}-1}\pi - \frac{\pi}{2}, \quad \phi(c) = \frac{2\pi c}{W_{tokens}} - \pi \\
    X(r,c) &= \left( \cos\theta(r) \cos\phi(c) + 1 \right) R \\
    Y(r,c) &= \left( \cos\theta(r) \sin\phi(c) + 1 \right) R
\end{aligned}
\end{equation}
where $R$ is a scale-dependent radius.
To encode the full circular topology and resolve symmetric ambiguity (e.g., distinguishing $\phi$ from $-\phi$), we replace the scalar column index $c$ in the encoding function $\alpha_i(c)$ with Cartesian $X$ and $Y$ coordinates, interleaved into even and odd frequency slots.

This encoding mathematically satisfies the topological invariants of the $S^2$ manifold. For periodicity (\textbf{C1}), the trigonometric components trace a closed circle as longitude wraps from $-\pi$ to $\pi$, guaranteeing $X(r, 0) = X(r, W)$ and $Y(r, 0) = Y(r, W)$. 
For polar convergence (\textbf{C2}), as latitude approaches the poles ($\theta \to \pm\pi/2$), $X$ and $Y$ converge to $R$, independent of the column index.

% The attention mechanism correctly perceives all polar tokens as occupying a singular physical coordinate.
% \sagie{the content is good, but again a bit too dense. sparse it out and explain a bit more. also we should note an ablation here about where is the cutoff between low and high frquencies}

\noindent \textbf{Cyclic Linear Encoding (High-Frequency Channels).}
While spherical encoding captures global topology, applying it uniformly to high-frequency channels introduces severe spatial aliasing. Pre-trained models rely on constant phase shifts between adjacent pixels for local texture. Multiplying a non-linear spherical projection by high frequencies catastrophically amplifies phase variance, destroying local coherence and introducing moir\'e artifacts. To preserve local structure, we maintain a linear Euclidean parameterization for high-frequency channels but enforce strict cyclicity by snapping each to the nearest integer harmonic:
\begin{equation}
\hat{\omega}_i = \text{round}(k_i) \cdot \omega_{\text{fund}}, \quad \alpha_i(c) = c \cdot \hat{\omega}_i.
\end{equation}
Because high frequencies destructively interfere over long distances, they do not impact global layout. Harmonically quantizing them ensures phase alignment modulo $2\pi$ at the boundaries (\textbf{C1}). This flawlessly stitches local textures without disrupting the global spherical structure.
While this linear form technically violates polar convergence (\textbf{C2}) in the high-frequency subspace, we observe that generation in these areas is dominated by the low frequency correspondences. This formulation allows the model to maintain consistent local texture density across the sphere.
% While this linear form technically violates polar convergence (C2) in the high-frequency subspace, it allows the model to maintain consistent local texture density across the sphere, preventing the destructive aliasing that would occur if high-frequency embeddings were forced to converge at the poles.

\subsection{Semantic Distortion CFG}
\label{sec:sdcfg}
We observe that pre-trained diffusion models natively exhibit equirectangular projection (ERP) characteristics, such as polar stretching and horizon curvature, when prompted with 360$^\circ$ descriptions. To amplify this inherent prior and complement the hard geometry of Spherical RoPE, we extend standard classifier-free guidance (CFG) to a three-way formulation. We introduce a geometric prompt $\mathbf{p}_{\text{geo}}$ that is anchored to the user prompt $\mathbf{p}$ via concatenation, $\mathbf{p}_{\text{anchor}} = [\mathbf{p}, \, \mathbf{p}_{\text{geo}}]$. This anchoring isolates the pure effect of ERP geometry without introducing conflicting semantic content.

At each denoising step, we compute three noise predictions: $\epsilon_{\text{cond}}$ (user prompt), $\epsilon_{\text{uncond}}$ (empty prompt), and $\epsilon_{\text{geo}}$ (anchored prompt). The final prediction combines these directions:
\begin{equation}\hat{\epsilon} = \epsilon_{\text{uncond}} + w_{\text{sem}} \cdot (\epsilon_{\text{cond}} - \epsilon_{\text{uncond}}) + \gamma \cdot (\epsilon_{\text{geo}} - \epsilon_{\text{cond}})
\end{equation}
where $w_{\text{sem}}$ and $\gamma$ independently control the semantic and geometric scales. This orthogonal decomposition provides precise control over the trade-off between prompt fidelity and geometric validity, gracefully recovering standard CFG when $\gamma = 0$.

\section{Experiments}
\label{sec:experiments}
We evaluate our framework on two 360$^\circ$ generation tasks: text-to-image and text-to-video.
We first describe the experimental setup, then present qualitative comparisons highlighting out-of-distribution generalization and image-to-panorama conditioning, followed by quantitative evaluations on both image and video benchmarks.
Finally we present an ablation study validating each component of our method. 
Extended results including a LLM-based perceptual assessment, ablations, limitations, along with our project page which includes interactive panorama viewers are in the supplementary.

% \sagie{give a quick overview of the experiments section. Also mention all results will be shown in supp webpage}

% StitchDiffusion~\cite{wang2024stitchdiffusion}, PanFusion~\cite{zhang2024panfusion}, DiffPano~\cite{wang2024diffpano}, and CubeDiff~\cite{kalischek2025cubediff} for images, and 360DVD~\cite{wang2024360dvd} and PanoDiT~\cite{zhang2025panodit} for video. Training-free baselines include PanoFree~\cite{liu2024panofree} and SphereDiff~\cite{park2025spherediff} for both images and video. \textbf{[Placeholder: Confirm final baseline list and versions used.]}
\subsection{Experimental Setup}
% For image generation, we use Flux.1 and Flux.2~\cite{blackforestlabs2024flux} as backbone architectures. 
% For video generation, we use LTX~2.3~\cite{ltxvideo2025}.  
\paragraph{Datasets.}
For image evaluation, following~\cite{wang2025survey}, we use the ODI-SR dataset~\cite{Deng_2021_CVPR}: 1,200 ERP panoramas spanning diverse indoor and outdoor scenes, captioned with Qwen3-VL~\cite{Qwen3-VL}. Captions are constrained to short descriptions for compatibility with CLIP-based text encoders. None of the evaluated methods were trained on this dataset, ensuring a fair out-of-domain generalization test~\cite{Deng_2021_CVPR}. 
For video evaluation, we use two prompt sets: \textit{SphereDiff-20}, a 20-scene benchmark from~\cite{park2025spherediff}, and \textit{Stress-20}, a set of 20 prompts generated by an LLM (Claude Opus~4.7) to stress-test panoramic generation with high-motion, dynamic content (construction protocol in Section~\ref{sec:supp_stress20}). 
Prompts are converted to match each baseline's native conditioning format (Section~\ref{sec:supp_prompt_conversion}).

\paragraph{Baselines.}
For both text-to-image and text-to-video, we benchmark against recent training and optimization-based methods. However, due to its inference latency, we exclude SphereDiff~\cite{park2025spherediff} from the ODI-SR benchmark, evaluating it instead via VLM and a user study on a reduced prompt set.
% For text-to-image, we compare against training-based methods: Text2Light~\cite{chen2022text2light}, Diffusion360~\cite{feng2023diffusion360}, PanFusion~\cite{zhang2024panfusion}, UniPano~\cite{ni2025makes}, PAR~\cite{wang2025conditional}, SMGD~\cite{sun2025smgd}, and DiT360~\cite{feng2025dit360}.
% We additionally evaluate SphereDiff~\cite{park2025spherediff} as a representative optimization-based method. Due to its prohibitive inference latency, we exclude it from the full quantitative benchmark and compare via LLM-based evaluation and user study on a smaller prompt set, following their suggested protocol.
% For text-to-video, we compare against training-based methods: 360DVD~\cite{wang2024360dvd} and PanoWAN~\cite{xia2026panowan}, as well as optimization-based methods: DynamicScaler~\cite{liu2025dynamicscaler} and SphereDiff~\cite{park2025spherediff}.

\paragraph{Evaluation metrics.}
For image generation~\cite{wang2025survey}, we compute universal metrics (FID~\cite{heusel2017gans}, KID~\cite{binkowski2018demystifying}, IS~\cite{salimans2016improved}, CS~\cite{radford2021learning}) on perspective crops to evaluate quality and diversity. We assess panoramic fidelity using distortion-aware features (FAED~\cite{oh2022bips}) and quantify wrap-boundary artifacts via DS~\cite{christensen2024geometry}. For video generation, we evaluate six VBench~\cite{vbench2024} dimensions: imaging quality~\cite{ke2021musiq}, text-alignment (CLIP Mean~\cite{radford2021learning}), temporal stability (temporal flickering, motion smoothness~\cite{li2023amt}), and semantic persistence (subject~\cite{caron2021dino} and background consistency). 
% For a fair comparison, all methods are scored against a unified per-scene reference prompt (Sec.~\ref{sec:supp_prompt_conversion}).
% For image generation, we report both universal and panorama-specific metrics~\cite{wang2025survey}. Universal metrics (FID~\cite{heusel2017gans}, KID~\cite{binkowski2018demystifying}, IS~\cite{salimans2016improved}, CS~\cite{radford2021learning}) are computed on perspective crops extracted via gnomonic projection to avoid ERP distortion bias. For panoramic fidelity, we report FAED~\cite{oh2022bips} and OmniFID~\cite{christensen2024geometry}, which evaluate perceptual quality using distortion-aware features, and DS~\cite{christensen2024geometry} to quantify seam artifacts at the wrap boundary. For video generation, we assess six complementary dimensions using VBench~\cite{vbench2024}: Imaging Quality (MUSIQ-based perceptual score~\cite{ke2021musiq}), CLIP Mean (per-frame text--video alignment~\cite{radford2021learning}), Temporal Flickering (frame-to-frame intensity stability), Motion Smoothness (movement continuity via frame-interpolation prior~\cite{li2023amt}), Subject Consistency (foreground appearance persistence via DINO~\cite{caron2021dino} features), and Background Consistency (global scene stability via CLIP features). To isolate generation quality from prompt verbosity, every method is scored against the same per-scene reference prompt (Sec.~\ref{sec:supp_prompt_conversion}).
Full metric details are provided in Section~\ref{sec:supp_metrics}.

\subsection{Qualitative Results}
% Qualitative comparison figure
% Requires: graphicx, xcolor, booktabs, adjustbox

\newcommand{\figpath}{figures/qualitative}

\newcommand{\panoW}{0.31\linewidth}
\newcommand{\cropW}{0.1033\linewidth}

% Stack pano + crops vertically in one cell
\newcommand{\methodcell}[2]{% #1=figpath prefix, #2=method
 \parbox[c]{10cm}{%
   \includegraphics[width=10cm]{#1/#2/pano.jpg}\\[0pt]
   \includegraphics[width=3.333cm]{#1/#2/crop_0.jpg}%
   \includegraphics[width=3.333cm]{#1/#2/crop_1.jpg}%
   \includegraphics[width=3.333cm]{#1/#2/crop_2.jpg}%
 }%
}

\begin{figure}[t]
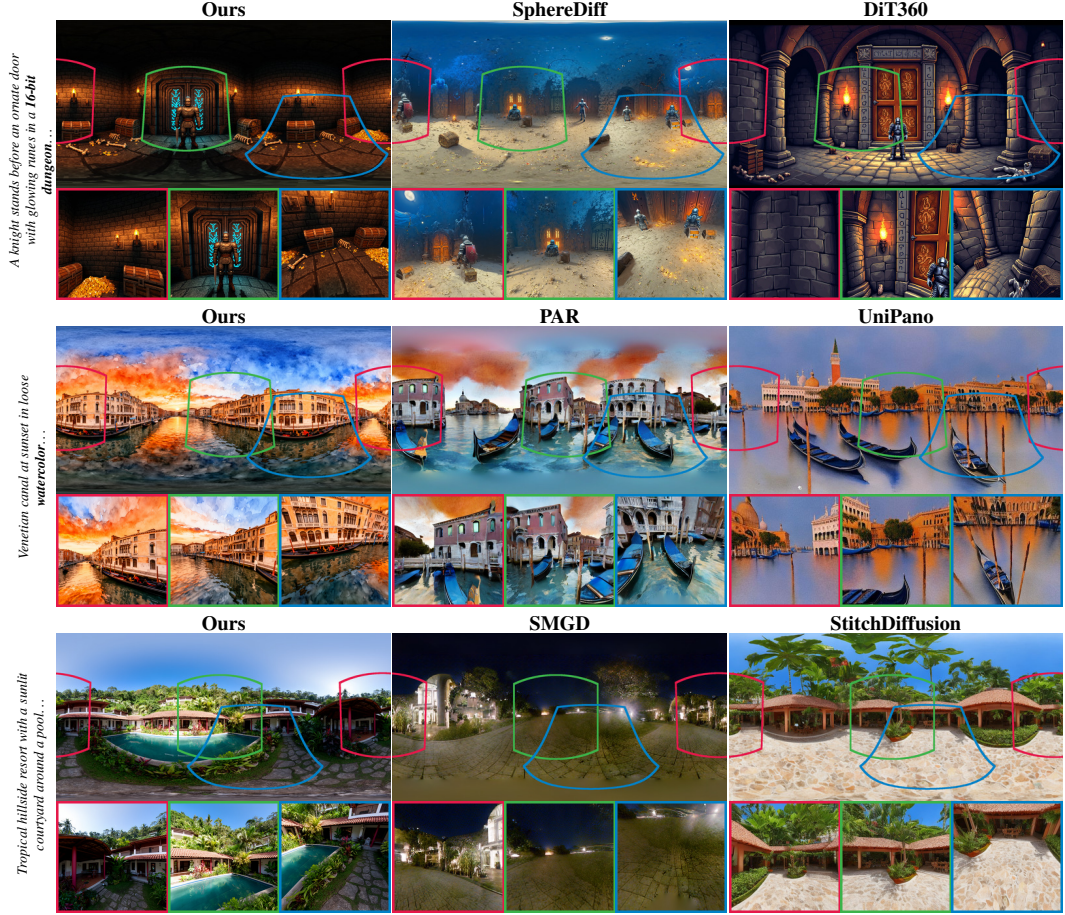

 \centering
 \setlength{\tabcolsep}{1pt}
 \resizebox{\linewidth}{!}{%
 \begin{tabular}{c  ccc}
 & {\LARGE\textbf{Ours}} & {\LARGE\textbf{SphereDiff}} & {\LARGE\textbf{DiT360}} \\
 %
 % \rotatebox[origin=c]{90}{\parbox[c]{7cm}{\large\centering\itshape A young girl in a straw hat\ldots in a \textbf{Studio Ghibli-style}\ldots}}
 % & \methodcell{\figpath/studio_ghibli_countryside}{Ours}
 % & \methodcell{\figpath/studio_ghibli_countryside}{SphereDiff}
 % & \methodcell{\figpath/studio_ghibli_countryside}{DiT360} \\
 % \noalign{\vspace{8pt}}

 \rotatebox[origin=c]{90}{\parbox[c]{7cm}{\large\centering\itshape A knight stands before an ornate door with glowing runes in a \textbf{16-bit dungeon}\ldots}}
 & \methodcell{\figpath/pixel_art_dungeon}{Ours}
 & \methodcell{\figpath/pixel_art_dungeon}{SphereDiff}
 & \methodcell{\figpath/pixel_art_dungeon}{DiT360} \\
 \noalign{\vspace{8pt}}
 
 % \midrule
 %
 % \rotatebox[origin=c]{90}{\parbox[c]{7cm}{\large\centering\itshape Floating islands in a pink-violet sky, linked by rope bridges\ldots}}
 % & \methodcell{\figpath/3d_render_floating_islands}{Ours}
 % & \methodcell{\figpath/3d_render_floating_islands}{UniPano}
 % & \methodcell{\figpath/3d_render_floating_islands}{SphereDiff} \\
 % \midrule
 %
 % \rotatebox[origin=c]{90}{\parbox[c]{7cm}{\large\centering\itshape Venetian canal at sunset in loose watercolor\ldots}}
 % & \methodcell{\figpath/watercolor_venetian_canal}{Ours}
 % & \methodcell{\figpath/watercolor_venetian_canal}{UniPano}
 % & \methodcell{\figpath/watercolor_venetian_canal}{SphereDiff} \\
 % \midrule
 %
 % \rotatebox[origin=c]{90}{\parbox[c]{7cm}{\large\centering\itshape Golden hour mountain vista with snow-capped peaks glowing in warm light\ldots}}
 % & \methodcell{\figpath/mountain_sunset}{Ours}
 % & \methodcell{\figpath/mountain_sunset}{UniPano}
 % & \methodcell{\figpath/mountain_sunset}{SphereDiff} \\
 % \midrule
 %

  & {\LARGE\textbf{Ours}} & {\LARGE\textbf{PAR}} & {\LARGE\textbf{UniPano}} \\

 \rotatebox[origin=c]{90}{\parbox[c]{7cm}{\large\centering\itshape Venetian canal at sunset in loose \textbf{watercolor}\ldots}}
 & \methodcell{\figpath/watercolor_venetian_canal}{Ours}
 & \methodcell{\figpath/watercolor_venetian_canal}{PAR}
 & \methodcell{\figpath/watercolor_venetian_canal}{UniPano} \\

 % \rotatebox[origin=c]{90}{\parbox[c]{7cm}{\large\centering\itshape A serene \textbf{Japanese garden} with a curved stone bridge over a koi pond\ldots}}
 % & \methodcell{\figpath/japanese_garden}{Ours}
 % & \methodcell{\figpath/japanese_garden}{PAR}
 % & \methodcell{\figpath/japanese_garden}{PanFusion} \\

  \noalign{\vspace{8pt}}

 % & {\LARGE\textbf{Ours}} & {\LARGE\textbf{SMGD}} & {\LARGE\textbf{StitchDiffusion}} \\
 %  \rotatebox[origin=c]{90}{\parbox[c]{7cm}{\large\centering\itshape A serene \textbf{lakeside} with limestone rocks, clear water, and two boats\ldots}}
 %  & \methodcell{\figpath/odisr_test_092}{Ours}
 %  & \methodcell{\figpath/odisr_test_092}{SMGD}
 %  & \methodcell{\figpath/odisr_test_092}{StitchDiffusion} \\
 
 & {\LARGE\textbf{Ours}} & {\LARGE\textbf{SMGD}} & {\LARGE\textbf{StitchDiffusion}} \\
  \rotatebox[origin=c]{90}{\parbox[c]{7cm}{\large\centering\itshape Tropical hillside resort with a sunlit courtyard around a pool\ldots}}
  & \methodcell{\figpath/odisr_test_041}{Ours}
  & \methodcell{\figpath/odisr_test_041}{SMGD}
  & \methodcell{\figpath/odisr_test_041}{StitchDiffusion} \\
 
 \end{tabular}%
 }
 \caption{\textbf{Text-to-360 Image Qualitative Results.} Each scene shows the ERP panorama (top) and three perspective crops (bottom): seam region (\textcolor{red}{red}), horizon (\textcolor{green}{green}), and ground (\textcolor{blue}{blue}). Our method, while being zero-shot, produces more coherent structure and successfully handles OOD prompts.}
 \label{fig:qualitative}
\end{figure}

% -------------------------------------------------------
% Figure 2: LAU Annotation (Ours vs UniPano vs PAR)
% -------------------------------------------------------
% \begin{figure*}[t]
%  \centering
%  \setlength{\tabcolsep}{1pt}
%  \resizebox{\linewidth}{!}{%
%  \begin{tabular}{c ccc}
%  & \textbf{Ours} & \textbf{UniPano} & \textbf{PAR} \\
%  %
%  \rotatebox[origin=c]{90}{\parbox[c]{7cm}{\large\centering\itshape A serene European garden surrounds a classical yellow stucco manor with red roof\ldots}}
%  & \methodcell{\figpath/odisr_test_050}{Ours}
%  & \methodcell{\figpath/odisr_test_050}{UniPano}
%  & \methodcell{\figpath/odisr_test_050}{PAR} \\
%  \midrule
%  %
%  \rotatebox[origin=c]{90}{\parbox[c]{7cm}{\large\centering\itshape A formal garden with gravel paths, boxwood hedges, and stone statues\ldots}}
%  & \methodcell{\figpath/odisr_test_048}{Ours}
%  & \methodcell{\figpath/odisr_test_048}{UniPano}
%  & \methodcell{\figpath/odisr_test_048}{PAR} \\
%  \midrule
%  %
%  \rotatebox[origin=c]{90}{\parbox[c]{7cm}{\large\centering\itshape A serene lakeside with limestone rocks, clear water, and two boats\ldots}}
%  & \methodcell{\figpath/odisr_test_092}{Ours}
%  & \methodcell{\figpath/odisr_test_092}{UniPano}
%  & \methodcell{\figpath/odisr_test_092}{PAR} \\
%  \bottomrule
%  \end{tabular}%
%  }
%  \caption{Qualitative comparison on the LAU benchmark. Each scene shows the full ERP panorama (top) and three perspective crops (bottom): seam region (\textcolor{red}{red}), horizon (\textcolor{green}{green}), and ground (\textcolor{blue}{blue}).}
%  \label{fig:qual_lau}
% \end{figure*}

\paragraph{Text-to-360 Image.} Figure~\ref{fig:qualitative} qualitatively compares our text-to-image method against baselines. Our approach consistently generates seamless, globally coherent panoramas, excelling on out-of-distribution stylized prompts that exceed typical training data. In contrast, training-based methods exhibit architectural bias, ignoring stylistic instructions. Although the optimization-based SphereDiff avoids these overfitting artifacts, its patch-based synthesis lacks global coherence.
% Our framework maintains a unified global composition by inheriting the full generative diversity of the diffusion model while enforcing spherical topology through Spherical RoPE and Semantic Distortion CFG.

\begin{figure*}

  \centering

  \setlength{\tabcolsep}{4pt} % Spacing for the Pano/Crops gap

  \def\totalheight{0.2\textwidth}

  \def\cropheight{0.098\textwidth}

  % The @{\hspace{30pt}} creates the large gap only between column 1 and 2

  \begin{tabular}{c@{\hspace{40pt}}cc}

    \small Reference Image & Crops & \small 360$^\circ$ Panorama \\

    % --- Row 1: Corgi ---

    \includegraphics[height=\totalheight, valign=t]{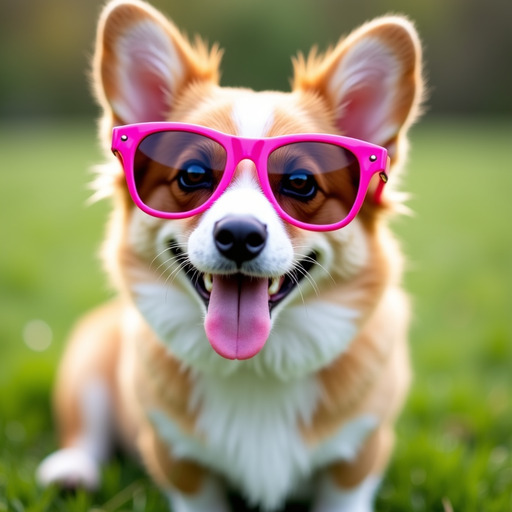} & 

        \begin{tabular}[t]{@{}c@{}}

        \includegraphics[height=\cropheight, valign=t]{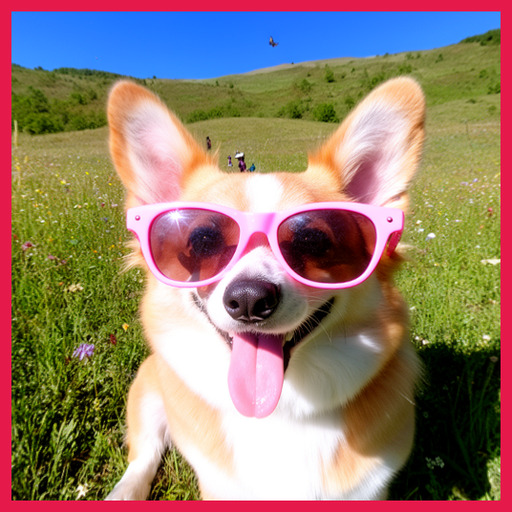} \\

        \noalign{\vspace{2pt}}

        \includegraphics[height=\cropheight, valign=t]{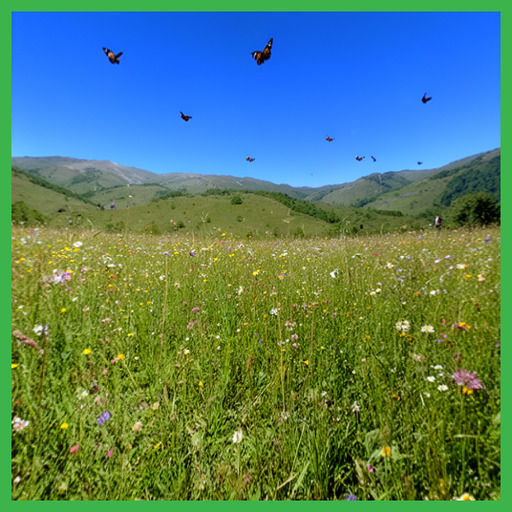}

    \end{tabular}  &

    \includegraphics[height=\totalheight, valign=t]{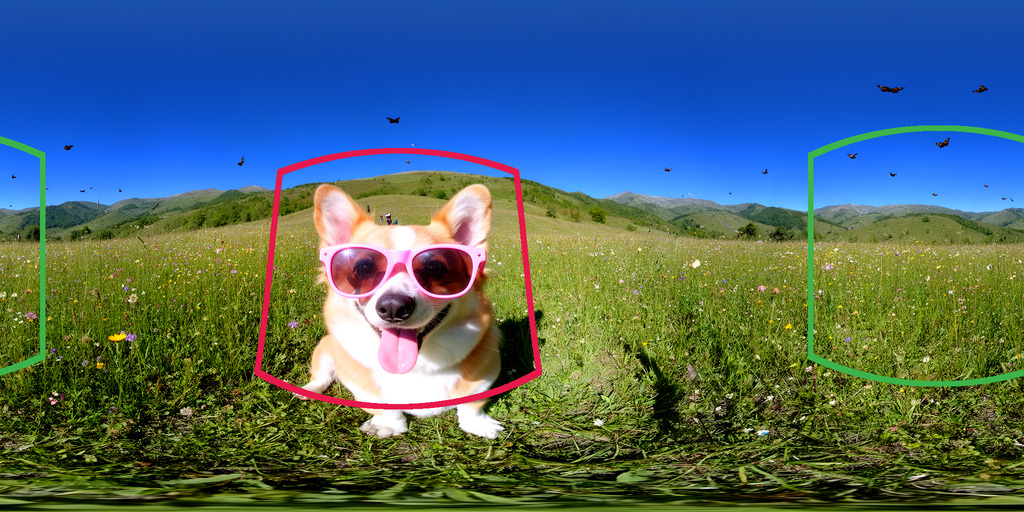}  \\

    \noalign{\vspace{8pt}} % Vertical separation between rows

    % --- Row 2: Astronaut ---

    \includegraphics[height=\totalheight, valign=t]{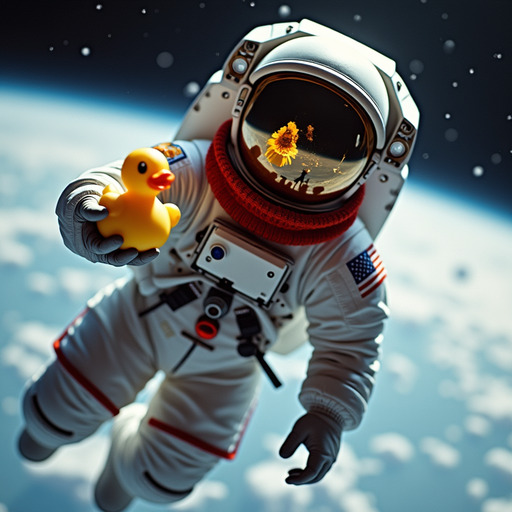} & 

    \begin{tabular}[t]{@{}c@{}}

        \includegraphics[height=\cropheight, valign=t]{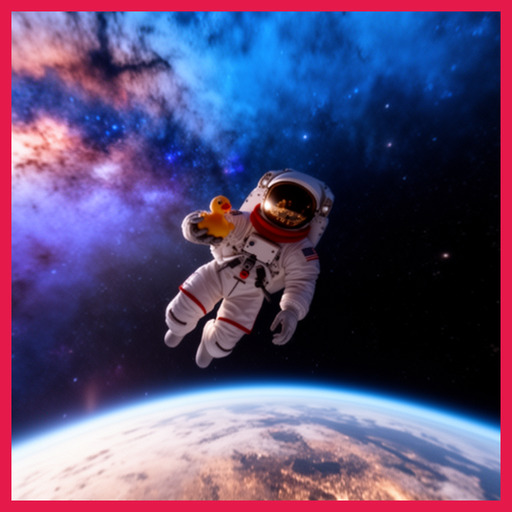} \\

        \noalign{\vspace{2pt}}

        \includegraphics[height=\cropheight, valign=t]{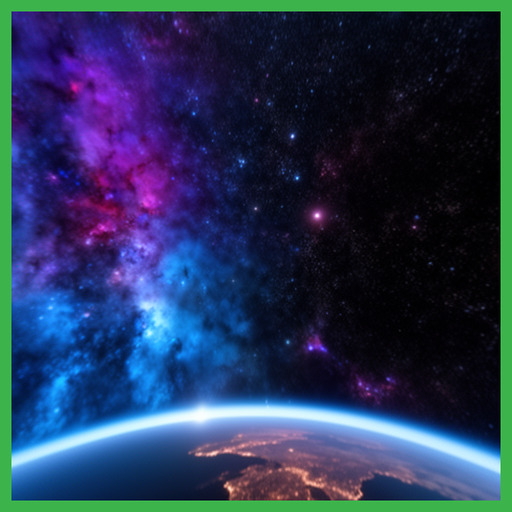}

    \end{tabular} &

    \includegraphics[height=\totalheight, valign=t]{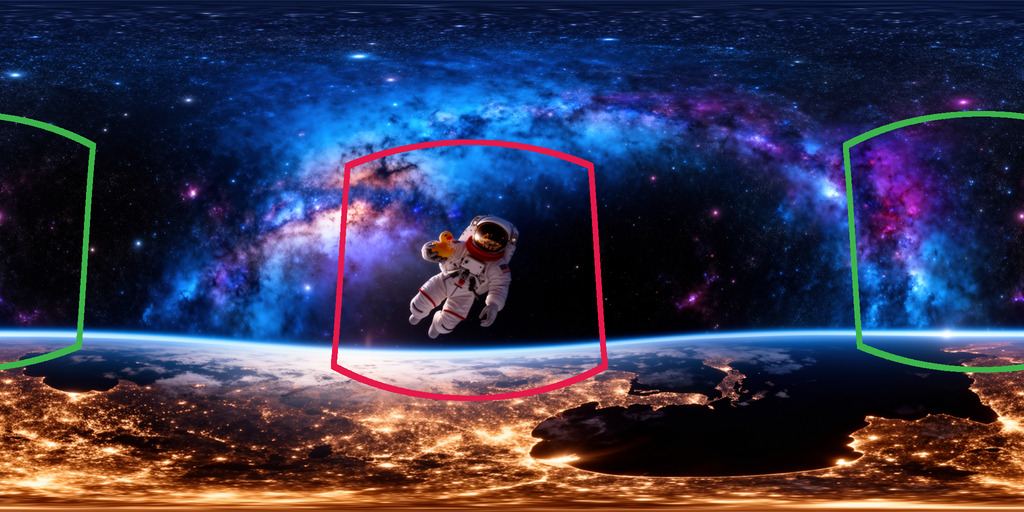} \\

  \end{tabular}

  \caption{\textbf{Image-conditioned 360 generation.} Our zero-shot approach natively inherits built-in model functionalities, such as image-conditioned generation. Using reference images, our framework generates seamless 360$^\circ$ panoramas that preserve identity and style across the entire sphere.}

  \label{fig:image2image}

\end{figure*}
\paragraph{Plug-and-Play Conditioning.} A primary advantage of our zero-shot approach is preserving the full versatility of the foundation model. By modifying only the inference process, we enable seamless 360$^\circ$ generation using built-in conditioning pipelines. Unlike training-based methods that require retraining adapters for new modalities, our framework natively inherits all supported functionalities. Figure~\ref{fig:image2image} demonstrates image-conditioned generation, where our approach maintains high-fidelity identity preservation, accurately capturing fine-grained reference details like the corgi's glasses or the astronaut's rubber duck. Beyond images, applying our pipeline to LTX~2.3 unlocks the first training-free 360$^\circ$ text-to-video-audio generation, yielding panoramic video with synchronized audio.

\subsection{Quantitative Results}

   \begin{table*}[t]
   \centering
   \caption{\textbf{Quantitative comparison}. We compare our zero-shot results with training-based methods on the ODISR dataset.
   We show both panorama-level metrics and multi-view metrics (computed from 8 equi-spaced horizontal perspective crops).
   Best in \textbf{bold}, second best \underline{underlined}.}
   \label{tab:quantitative}
   \resizebox{\textwidth}{!}{%
   \begin{tabular}{l|l|cc|cccc}
   \toprule
   & & \multicolumn{2}{c|}{\textbf{Panorama-Level Metrics}} & \multicolumn{4}{c}{\textbf{Multi-View Metrics}} \\
   \textbf{Method} & \textbf{Type} & FAED $\downarrow$ & DS $\downarrow$ & FID $\downarrow$ & KID$_{\times 10^2}$ $\downarrow$ & IS $\uparrow$ & CS $\uparrow$ \\
   \midrule
   Text2Light~\cite{chen2022text2light} & Full Fine-Tune & \cellcolor[rgb]{0.91,0.73,0.73}77.94 & \cellcolor[rgb]{0.91,0.73,0.73}3.82 & \cellcolor[rgb]{0.91,0.73,0.73}81.01 & \cellcolor[rgb]{0.91,0.73,0.73}4.55 & \cellcolor[rgb]{0.91,0.75,0.73}6.25 & \cellcolor[rgb]{0.73,0.91,0.73}\textbf{18.60} \\
   Diffusion360~\cite{feng2023diffusion360} & Full Fine-Tune & \cellcolor[rgb]{0.91,0.78,0.73}71.02 & \cellcolor[rgb]{0.74,0.90,0.73}0.87 & \cellcolor[rgb]{0.91,0.90,0.73}53.17 & \cellcolor[rgb]{0.91,0.90,0.73}2.82 & \cellcolor[rgb]{0.91,0.89,0.73}8.92 & \cellcolor[rgb]{0.76,0.91,0.73}18.46 \\
   StitchDiffusion~\cite{wang2024stitchdiffusion} & LoRA & \cellcolor[rgb]{0.91,0.83,0.73}63.10 & \cellcolor[rgb]{0.75,0.89,0.73}1.01 & \cellcolor[rgb]{0.86,0.91,0.73}43.73 & \cellcolor[rgb]{0.83,0.91,0.73}1.84 & \cellcolor[rgb]{0.83,0.91,0.73}10.88 & \cellcolor[rgb]{0.75,0.91,0.73}18.50 \\
   PanFusion~\cite{zhang2024panfusion} & LoRA & \cellcolor[rgb]{0.83,0.91,0.73}39.29 & \cellcolor[rgb]{0.75,0.89,0.73}1.08 & \cellcolor[rgb]{0.84,0.91,0.73}40.37 & \cellcolor[rgb]{0.86,0.91,0.73}2.12 & \cellcolor[rgb]{0.90,0.91,0.73}9.66 & \cellcolor[rgb]{0.91,0.80,0.73}17.14 \\
   UniPano~\cite{ni2025makes} & LoRA & \cellcolor[rgb]{0.87,0.91,0.73}46.19 & \cellcolor[rgb]{0.75,0.89,0.73}1.11 & \cellcolor[rgb]{0.80,0.91,0.73}34.76 & \cellcolor[rgb]{0.82,0.91,0.73}1.78 & \cellcolor[rgb]{0.89,0.91,0.73}9.73 & \cellcolor[rgb]{0.91,0.73,0.73}16.81 \\
   SMGD~\cite{sun2025smgd} & Full Fine-Tune & \cellcolor[rgb]{0.79,0.91,0.73}33.55 & \cellcolor[rgb]{0.73,0.91,0.73}\underline{0.79} & \cellcolor[rgb]{0.89,0.91,0.73}49.02 & \cellcolor[rgb]{0.85,0.91,0.73}2.01 & \cellcolor[rgb]{0.91,0.73,0.73}5.77 & \cellcolor[rgb]{0.75,0.91,0.73}\underline{18.52} \\
   DiT360~\cite{feng2025dit360} & LoRA & \cellcolor[rgb]{0.86,0.91,0.73}43.42 & \cellcolor[rgb]{0.84,0.80,0.73}2.59 & \cellcolor[rgb]{0.73,0.91,0.73}\textbf{23.28} & \cellcolor[rgb]{0.74,0.91,0.73}0.92 & \cellcolor[rgb]{0.73,0.91,0.73}\textbf{12.90} & \cellcolor[rgb]{0.75,0.91,0.73}18.52 \\
   PAR~\cite{wang2025conditional} & Full Fine-Tune & \cellcolor[rgb]{0.80,0.91,0.73}34.79 & \cellcolor[rgb]{0.73,0.91,0.73}\textbf{0.75} & \cellcolor[rgb]{0.75,0.91,0.73}25.80 & \cellcolor[rgb]{0.73,0.91,0.73}\textbf{0.85} & \cellcolor[rgb]{0.86,0.91,0.73}10.47 & \cellcolor[rgb]{0.75,0.91,0.73}18.51 \\
   \midrule
   Ours (FLUX.1) & \textbf{Zero-Shot} & \cellcolor[rgb]{0.76,0.91,0.73}\underline{29.94} & \cellcolor[rgb]{0.74,0.90,0.73}0.89 & \cellcolor[rgb]{0.87,0.91,0.73}45.22 & \cellcolor[rgb]{0.91,0.91,0.73}2.63 & \cellcolor[rgb]{0.86,0.91,0.73}10.32 & \cellcolor[rgb]{0.75,0.91,0.73}18.52 \\
   \textbf{Ours (FLUX.2)} & \textbf{Zero-Shot} & \cellcolor[rgb]{0.73,0.91,0.73}\textbf{25.40} & \cellcolor[rgb]{0.74,0.90,0.73}0.94 & \cellcolor[rgb]{0.75,0.91,0.73}\underline{25.33} & \cellcolor[rgb]{0.74,0.91,0.73}\underline{0.90} & \cellcolor[rgb]{0.74,0.91,0.73}\underline{12.75} & \cellcolor[rgb]{0.77,0.91,0.73}18.43 \\
   \bottomrule
   \end{tabular}%
   }
   \end{table*}

\paragraph{Text-to-360 image generation.}

Table~\ref{tab:quantitative} shows our zero-shot approach matches or outperforms trained baselines on ODI-SR. At the panorama level, we achieve the best FAED score without task-specific training, demonstrating that our outputs most accurately capture the global distribution and structural integrity of real $360^\circ$ scenes. This global coherence is further reinforced by a competitive Discontinuity Score (DS), as our method effectively avoid boundary seams. On crop-level metrics, our individual perspective views retain high local realism: we match PAR on KID and IS, and perform competitively alongside DiT360. Ultimately, SpheRoPE delivers strong performance entirely zero-shot, bypassing the massive computational bottlenecks of dataset collection and model fine-tuning.

\label{sec:video_eval}

\begin{table*}
\centering
\setlength{\tabcolsep}{4pt}
\caption{\textbf{Text-to-360 Video Comparison.} We compare our zero-shot results with training and optimization-based methods. Values $\times 100$
except CLIP Mean. Best in \textbf{bold},
second-best \underline{underlined}.}
\label{tab:vbench_new}
\scriptsize
\begin{tabular}{l|lc|c|cccccc}
\textbf{Prompt Set} & \textbf{Method} & \textbf{Type}
& \shortstack{\textbf{Sec./}\\\textbf{frame} $\downarrow$}
& \shortstack{\textbf{Img.}\\\textbf{Qual.} $\uparrow$}
& \shortstack{\textbf{CLIP}\\\textbf{Mean} $\uparrow$}
& \shortstack{\textbf{Temp.}\\\textbf{Flicker} $\uparrow$}
& \shortstack{\textbf{Motion}\\\textbf{Smooth.} $\uparrow$}
& \shortstack{\textbf{Subj.}\\\textbf{Cons.} $\uparrow$}
& \shortstack{\textbf{BG}\\\textbf{Cons.} $\uparrow$} \\
\midrule
\multirow{5}{*}{\textit{SphereDiff-20}}
& 360DVD~\cite{wang2024360dvd} & LoRA & \textbf{0.95} & 47.99 & 26.02 & 99.18 & 98.98 & 95.11 & 96.13 \\
& PanoWan~\cite{xia2026panowan} & LoRA & 1.59 & 39.10 & 27.32 & 99.13 & 99.43 & \underline{97.40} & \underline{96.59} \\
& DynamicScaler~\cite{liu2025dynamicscaler} & Optimization & 51.56 & \underline{53.21} & \underline{27.38} & 97.13 & 98.11 & 93.79 & 94.80 \\
& SphereDiff~\cite{park2025spherediff} & Optimization & 6.08 & 40.08 & 27.03 & \underline{99.35} & \underline{99.51} & 95.90 & 96.33 \\
& \textbf{Ours (LTX~2.3)} & \textbf{Zero-Shot} & \underline{1.11} & \textbf{53.64} & \textbf{27.95} & \textbf{99.58} & \textbf{99.62} & \textbf{98.22} & \textbf{97.71} \\
\midrule
\multirow{5}{*}{\textit{Stress-20}}
& 360DVD~\cite{wang2024360dvd} & LoRA & \textbf{0.95} & 50.19 & 28.62 & \underline{98.66} & \underline{98.42} & \underline{93.30} & \underline{94.82} \\
& PanoWan~\cite{xia2026panowan} & LoRA & 1.59 & 39.49 & 28.42 & 97.11 & 98.58 & 91.64 & 93.66 \\
& DynamicScaler~\cite{liu2025dynamicscaler} & Optimization & 51.56 & \underline{50.65} & \textbf{32.39} & 94.45 & 96.17 & 91.58 & 93.25 \\
& SphereDiff~\cite{park2025spherediff} & Optimization & 6.08 & 33.82 & 28.20 & 96.52 & 97.72 & 88.55 & 93.34 \\
& \textbf{Ours (LTX~2.3)} & \textbf{Zero-Shot} & \underline{1.11} & \textbf{51.76} & \underline{28.91} & \textbf{98.91} & \textbf{99.46} & \textbf{96.51} & \textbf{96.02} \\
\bottomrule
\end{tabular}
\end{table*}

\paragraph{Text-to-360 video generation.}
Table~\ref{tab:vbench_new} reports our results. On SphereDiff-20, our zero-shot approach leads all metrics (temporal stability, imaging quality, and CLIP Mean). On the high-motion Stress-20 set, our method continues to dominate all temporal metrics and imaging quality. While DynamicScaler achieves a higher CLIP Mean via multi-pass stitching, it suffers the worst temporal coherence and an order of magnitude slower inference time. Meanwhile, SphereDiff's patch-based synthesis struggles severely with dynamic scenes, yielding the lowest imaging quality and subject consistency.

\begin{table}
\centering
\caption{\textbf{Human preference study.} We collect 320 blind pairwise judgments from 18 annotators using interactive image panorama viewers. For each pair, raters make a three-way choice (A, B, or tie) on overall preference and text alignment. Our method (using Flux.2) is consistently preferred across all baselines and both criteria.}
\label{tab:user_study}
\small
\begin{tabular}{l ccc ccc}
\toprule
& \multicolumn{3}{c}{Overall Preference} & \multicolumn{3}{c}{Text Alignment} \\
\cmidrule(lr){2-4} \cmidrule(lr){5-7}
Baseline & \textbf{Ours} & Tie & Base & \textbf{Ours} & Tie & Base \\
\midrule
PanFusion~\cite{zhang2024panfusion}  & \textbf{88.5\%} &  3.3\% &  8.2\% & \textbf{83.6\%} &  6.6\% &  9.8\% \\
UniPano~\cite{ni2025makes}    & \textbf{82.7\%} &  3.8\% & 13.5\% & \textbf{65.4\%} & 17.3\% & 17.3\% \\
SMGD~\cite{sun2025smgd}       & \textbf{95.2\%} &  2.4\% &  2.4\% & \textbf{95.2\%} &  2.4\% &  2.4\% \\
SphereDiff~\cite{park2025spherediff} & \textbf{93.2\%} &  3.4\% &  3.4\% & \textbf{86.4\%} & 10.2\% &  3.4\% \\
DiT360~\cite{feng2025dit360}     & \textbf{56.5\%} & 10.9\% & 32.6\% & \textbf{50.0\%} & 41.3\% &  8.7\% \\
PAR~\cite{wang2025conditional}        & \textbf{71.7\%} & 15.0\% & 13.3\% & \textbf{73.3\%} & 21.7\% &  5.0\% \\
\bottomrule
\end{tabular}
\end{table}

\paragraph{User Preference Study.}
Table~\ref{tab:user_study} summarizes our blind pairwise user study (full protocol in Section~\ref{sec:supp_user_study}) among the six most competitive text-to-image panorama generation. Users consistently prefer our panoramas over all baselines for both overall quality and text alignment. We observe the widest preference margins against other methods. Even against DiT360, the most competitive baseline, our method secures a clear preference advantage in overall quality.
% We conduct a blind pairwise user study comparing our panoramas against each baseline. 
% For each prompt, participants are shown two anonymized panoramas in randomized left-right order and asked two questions: 
% ``Which panorama do you prefer?'' (overall quality) and ``Which one has better text alignment?'', each with an additional tie option. 
% We report the fraction of comparisons in which our method is preferred, ties, and baseline preference for both criteria.

% As shown in Table~\ref{tab:user_study}, users consistently prefer our results over all baselines on both overall quality and text alignment, 
% with the strongest margins against SMGD and SphereDiff. 
% The only relatively competitive baseline is DIT360, where our method still obtains a clear preference advantage on overall quality. Full study protocol details are provided in Sec.~\ref{sec:supp_user_study}.

\subsection{Ablation Study}
We evaluate the quantitative impact of each core component below. Supplementary Section~\ref{supp:ablation} provides extensive additional qualitative and quantitative ablations such as our RoPE components, Semantic Distortion CFG, and frequency quantization tolerance.

  \begin{table}[t]
  \centering
  \caption{\textbf{Components Contribution.} We ablate key components of our method: spherical RoPE and Semantic Distortion CFG.
  Best in \textbf{bold}, second best \underline{underlined}.}
  \label{tab:ablation}
  \resizebox{\columnwidth}{!}{%
  \begin{tabular}{l|cc|cccc}
  & \multicolumn{2}{c|}{\textbf{Panorama-Level Metrics}} & \multicolumn{4}{c}{\textbf{Multi-View Metrics}} \\
  \textbf{Configuration} & FAED $\downarrow$ & DS $\downarrow$ & FID $\downarrow$ & KID$_{\times 10^2}$ $\downarrow$ & IS $\uparrow$ & CS $\uparrow$ \\
  \midrule
  Vanilla FLUX.2 & \underline{33.38} & 1.37 & \underline{26.05} & \underline{1.06} & \textbf{14.24} & \underline{18.42} \\
  (+) SpheRoPE & 40.90 & \textbf{0.92} & 28.62 & 1.53 & \underline{13.77} & 18.25 \\
(+) SD-CFG & 27.41 & 1.22 & 28.86 & 1.16 & 12.98 & 18.42 \\
  \textbf{(+) SpheRoPE (+) SD-CFG (Ours)} & \textbf{25.40} & \underline{0.94} & \textbf{25.33} & \textbf{0.90} & 12.75 & \textbf{18.43} \\
  \bottomrule
  \end{tabular}%
  }
  \end{table}
   
\paragraph{Component Contributions.}
Table~\ref{tab:ablation} ablates our framework components on FLUX.2. To ensure a fair comparison, the vanilla baseline is explicitly guided with panoramic text prompts. While scoring reasonably well on unconstrained multi-view metrics, it lacks global structure, evidenced by boundary seams (high Discontinuity Score). 
When applied independently, Spherical RoPE successfully enforces topological invariants, resolving hard discontinuities and improving DS. However, this geometric constraint alone degrades global image fidelity (FAED increases) as the model struggles with the enforced spherical layout. 
Conversely, Semantic Distortion CFG (SD-CFG) applied independently compensates for ERP distortion and improves global fidelity (FAED), but cannot close boundary seams. 
Combining SpheRoPE's spatial topology with SD-CFG's distortion-aware guidance cleanly resolves this tension, achieving the best overall FAED and multi-view metrics while maintaining strict panoramic coherence. Qualitative examples are provided in the supplementary (Figure~\ref{fig:ablation_components}).

\section{Conclusion}
\label{sec:conclusion}
% We present a zero-shot, optimization-free, framework for synthesizing seamless 360$^\circ$ panoramas of images and videos from pre-trained diffusion transformers. By introducing Spherical RoPE and semantic distortion CFG, we enforce horizontal periodicity and polar convergence natively within the latent space. Our approach preserves the full semantic and stylistic versatility of foundation models, enabling high-fidelity panoramic generation without task-specific fine-tuning or costly optimization. We demonstrate state-of-the-art results across multiple backbones. On the ODI-SR benchmark, our zero-shot paradigm outperforms established training-based methods in panoramic coherence and human preference. We further demonstrate the effectiveness of our approach in 360-video generation.
% Our framework is designed to scale along with the rapid evolution of foundation models. This decoupling of geometric grounding from model training provides a robust path for future immersive media synthesis. We anticipate that extending this spherical encoding will unlock new possibilities for VR and spatial computing.
We present a zero-shot, optimization-free framework for synthesizing seamless 360$^\circ$ images and videos using pre-trained diffusion models. By introducing Spherical RoPE and Semantic Distortion CFG, we natively enforce spherical topological invariants within the latent space. This approach preserves the full versatility of foundation models without requiring fine-tuning. Despite being entirely training-free, our method achieves highly competitive performance against fully fine-tuned baselines, consistently outperforming them in panoramic coherence and human preference.
By decoupling geometric grounding from model training, our framework scales directly with future foundation models, paving the way for immersive media and VR applications.

{
    \small
    \bibliographystyle{plainnat}
    \bibliography{main}
}

\newpage
\appendix
% \section{Supplementary}

\section{Preliminaries}
\label{supp:preliminary}
\noindent \textbf{Diffusion Models and CFG.} \quad 
Diffusion models~\cite{rombach2022ldm} learn to reverse a gradual noising process: given a clean sample $\mathbf{x}_0$, Gaussian noise is progressively added over $T$ timesteps to produce $\mathbf{x}_T \sim \mathcal{N}(\mathbf{0}, \mathbf{I})$, and a neural network $\epsilon_\theta$ is trained to predict the noise at each step. Generation proceeds by iteratively denoising from pure noise back to a clean sample. To condition generation on a text prompt $\mathbf{p}$, classifier-free guidance (CFG)~\cite{ho2022cfg} interpolates between an unconditional prediction $\epsilon_{\text{uncond}}$ and a conditional prediction $\epsilon_{\text{cond}}$
\begin{equation}
\label{eq:cfg}
    \hat{\epsilon} = \epsilon_{\text{uncond}} + w_{sem} \cdot (\epsilon_{\text{cond}} - \epsilon_{\text{uncond}}),
\end{equation}
where $w_{sem} > 1$ amplifies the influence of the text condition. This formulation operates along a single guidance direction, from unconditional to semantically conditioned, and encodes no geometric priors about the output domain. 
The same principle applies to flow-matching models that predict velocity rather than noise, where guidance is performed in the velocity or $\mathbf{x}_0$-prediction space.

\noindent \textbf{Rotary Position Embedding.} \quad 
Modern diffusion transformers~\cite{peebles2023dit} encode spatial structure through rotary position embedding (RoPE)~\cite{su2024roformer}. The transformer block is permutation-equivariant by design. A positional encoding mechanism is therefore necessary to model the strong spatial dependencies present in natural images. While early approaches used fixed sinusoidal~\cite{vaswani2017attention} or learned absolute embeddings~\cite{devlin2019bert}, RoPE has emerged as a more effective alternative that encodes \textit{relative} positions directly into the query-key interaction in the attention mechanism.
Concretely, RoPE represents a position coordinate as a set of rotations at different frequencies. For a token at spatial position $(r, c)$ in a 2D image, RoPE is applied axially: one subset of the hidden dimension is rotated according to the row coordinate $r$, and the other according to the column coordinate $c$, enabling the model to encode relative offsets along each spatial axis independently. Focusing on the width (column) axis, the rotation angle for frequency channel $i$ is
\begin{equation}
    \alpha_i(c) = c \cdot \omega_i, \quad \text{for} \quad \omega_i = \theta_{\text{base}}^{-2i/D_w}
\end{equation}
where $D_w$ is the number of width frequency channels, and $\theta_{\text{base}}$ determines the geometric series of frequencies. 
Each frequency channel $i$ produces a 2D rotation matrix $\mathbf{R}(\alpha_i)$ applied to the corresponding pair of dimensions in the query and key vectors
\begin{equation}
    \mathbf{R}(\alpha_i) = \begin{pmatrix} \cos \alpha_i & -\sin \alpha_i \\ \sin \alpha_i & \cos \alpha_i \end{pmatrix}.
\end{equation}
The full RoPE transformation applies these block-diagonal rotations to the query $\mathbf{q}$ and key $\mathbf{k}$ before computing the attention inner product. 

\section{Additional Experiments}
\subsection{Implementation Details}
All experiments are conducted on NVIDIA H100 GPUs. 
For image generation, we produce ERP panoramas at a resolution of $1024 \times 2048$ using 50 denoising steps. We use a harmonic quantization tolerance of $\varepsilon = 0.06$. 
For the spherical Cartesian encoding, the radius $R$ is set as $R = W_{\text{tokens}} / 2$, and $R = W_{\text{span}} / 2$ for LTX~2.3, where $W_{\text{span}}$ is the coordinate-normalized width extent. 
The RoPE dimensionality follows native structure of each backbone. In all cases, only the width axis is modified; the temporal and height axes remain identical to the stock model.
The Semantic Distortion CFG scale is set to $\gamma = 6.0$ across all experiments. 
The semantic guidance scale $w_{\text{sem}}$ follows the default CFG value of each backbone. 

All assets are used under their original licenses: FLUX.1/2 [dev] under Black Forest Labs' Non-Commercial License, LTX-Video 2.3 under Lightricks' Community License, ODI-SR and VBench under Apache 2.0. Use is strictly for non-commercial academic research.
   
All baselines are evaluated using their official codebases and released pretrained weights. For StitchDiffusion, we use the diffusers-based reimplementation by~\cite{wang2024stitchdiffusion} with the official LoRA weights, as the original codebase does not provide a runnable inference script. Since the original Stable Diffusion 2.1-base weights were removed from HuggingFace by Stability AI, we use a community mirror (\texttt{Manojb/stable-diffusion-2-1-base}) for methods that depend on it. All other methods use their default inference hyperparameters as specified in their respective repositories. Generated panoramas are resized to $512 \times 1024$ before evaluation when the native resolution differs.

\paragraph{Circular Latent Encoding}
To address boundary discontinuities introduced by standard zero-padding in convolution-based VAEs, we follow 360Anything~\cite{wu2026360anything}. We apply circular padding symmetrically to the VAE decoder, horizontally extending the tensors to simulate periodic boundaries prior to the decoding pass. By subsequently cropping the results, we ensure that the convolutional receptive fields perceive a continuous spherical manifold. This procedure effectively eliminates latent-space seams at the decoding stage with zero computational overhead to the diffusion process.

\subsection{Evaluation Metrics}
\label{sec:supp_metrics}

We evaluate $360^\circ$ panorama generation using both universal image quality metrics and panorama-specific metrics, as standard measures often fail to account for the global layout and geometric properties unique to equirectangular projection (ERP)~\cite{wang2025survey}.

\paragraph{Universal metrics.}
We report Fr\'echet Inception Distance (FID)~\cite{heusel2017gans} and Kernel Inception Distance (KID)~\cite{binkowski2018demystifying}, which measure the distributional distance between generated and real images using Inception-v3 features. We also report Inception Score (IS)~\cite{salimans2016improved}, which evaluates both quality and diversity via the KL divergence between conditional and marginal class distributions, and CLIP Score (CS)~\cite{radford2021learning}, the cosine similarity between CLIP text and image embeddings.

Because these metrics rely on encoders trained on perspective images, they may penalize geometrically correct panoramic features as artifacts~\cite{wang2025survey}. To mitigate this, we compute all universal metrics on perspective crops extracted via gnomonic projection from the ERP panoramas, ensuring the feature extractors operate on undistorted patches consistent with their training distribution.

We further note that distributional metrics (FID, KID, FAED) inherently favor training-based methods, whose fine-tuning data overlaps with the evaluation distribution. Our zero-shot approach generates from the broader, unconstrained distribution of the foundation model, placing it at a structural disadvantage on these metrics despite producing visually competitive results.

\paragraph{Panorama-specific metrics.}
FAED~\cite{oh2022bips} computes Fr\'echet distances using an autoencoder trained specifically on $360^\circ$ panoramic images, providing a distortion-aware alternative to standard FID that better captures the perceptual and geometric quality of panoramic content.
Discontinuity Score (DS)~\cite{christensen2024geometry} quantifies seam artifacts at the horizontal wrap boundary ($\pm\pi$ meridian) using kernel-based edge detection. Lower DS indicates better perceived continuity across the seam - a direct measure of whether horizontal periodicity (\textbf{C1}) is satisfied in the generated output.

\subsection{Additional Quantitative Results}
\begin{table*}
\centering
\setlength{\tabcolsep}{4pt}
\caption{\textbf{LLM-based evaluation on $360^\circ$ panorama generation.} Following SphereDiff~\cite{park2025spherediff}, we use GPT-4o to score 14 perspective views per scene on a 1--5 scale across panoramic and image criteria, using their 20-prompt benchmark. We additionally report inference time per scene (seconds) on an NVIDIA H100. Best in \textbf{bold}, second best \underline{underlined}.}
\label{tab:panorama_eval}
\scriptsize
\begin{tabular}{l|l|cc|cc|c}
% \toprule
& & \multicolumn{2}{c|}{Panoramic Criteria}
& \multicolumn{2}{c|}{Image Criteria}
& \multirow{2}{*}{Time/scene (s)} \\
\cline{3-6}
Method & Type
& Distortion $\uparrow$ & End Continuity $\uparrow$
& Image Quality $\uparrow$ & Aesthetic Appearance $\uparrow$
& \\
\midrule

360 LoRA & LoRA & 2.03 & 3.42 & 2.97 & 3.49 & -- \\
Text2Light & Fine-Tune & 2.38 & 3.45 & 2.42 & 2.78 & \underline{36} \\
PanFusion & LoRA & 1.97 & 3.70 & 2.82 & 3.45 & \textbf{28} \\
DynamicScaler & Optimization & 2.85 & 3.99 & \textbf{4.50} & \textbf{4.58} & -- \\

SphereDiff (Flux1) & Optimization & \textbf{3.15} & \underline{4.60} & 3.76 & 4.24 & 1274 \\
\midrule
\textbf{SpheRoPE (Flux.1)} & \textbf{Zero-Shot} & 3.05 & 4.56 & \underline{4.37} & \underline{4.49} & 62 \\
\textbf{SpheRoPE (Flux.2)} & \textbf{Zero-Shot} & \underline{3.13} & \textbf{4.69} & 4.18 & 4.41 & 189 \\

\bottomrule
\end{tabular}
\vspace{2pt}\\
{\scriptsize
\textsuperscript{\dag}Scores for 360\,LoRA, Text2Light, PanFusion, and DynamicScaler are taken from~\cite{park2025spherediff}. We additionally time all methods on our hardware for fair comparison, except 360\,LoRA and DynamicScaler, which are reimplemented by~\cite{park2025spherediff} and do not have publicly available inference code. All other scores and timings are ours (NVIDIA H100).
}
\end{table*}

% 360 LoRA & 2.027 & 3.423 & 2.965 & 3.492 & -- \\
% Text2Light & 2.381 & 3.454 & 2.415 & 2.777 & -- \\
% PanFusion & 1.965 & 3.696 & 2.819 & 3.450 & -- \\
% DynamicScaler & 2.854 & 3.985 & 4.496 & 4.577 & -- \\

% SphereDiff (SANA, High)
% & 2.825 & 4.721 & 3.843 & 4.336 & 284 \\
% SphereDiff (SANA, Low)
% & 3.025 & 4.768 & 4.129 & 4.514 & 284 \\

% SphereDiff (Flux1, High)
% & 3.146 & 4.596 & 3.764 & 4.236 & 1274 \\
% SphereDiff (Flux1, Low)
% & 3.236 & 4.650 & 3.914 & 4.286 & 1274 \\

% \textbf{SpheRoPE (Flux1, High)}
% & 3.045 & 4.562 & 4.365 & 4.494 & 62 \\
% \textbf{SpheRoPE (Flux1, Low)}
% & 3.125 & 4.664 & 4.379 & 4.504 & 62 \\

% \textbf{SpheRoPE (Flux2, High)}
% & 3.132 & 4.689 & 4.182 & 4.414 & 189 \\
% \textbf{SpheRoPE (Flux2, Low)}
% & 3.239 & 4.714 & 4.282 & 4.464 & 189 \\
\paragraph{LLM-based Evaluation.}
Following the established protocol of SphereDiff~\cite{park2025spherediff}, we evaluate each panorama with GPT-4o~\cite{hurst2024gpt4o} as an LLM judge~\cite{zheng2023judging}, scoring 14 perspective views on a 1-5 scale across four criteria: \textit{distortion} (whether the image resembles a natural photograph), \textit{end continuity} (seamlessness at the wrap boundary), \textit{image quality}, and \textit{aesthetic appearance}. The exact prompt used, taken directly from SphereDiff~\cite{park2025spherediff}, is reproduced in Section~\ref{sec:supp_llm_prompt}. Table~\ref{tab:panorama_eval} reports the results: we achieve the best end continuity, are competitive on distortion, and substantially outperform SphereDiff on image quality and aesthetics, while running over $20\times$ faster.

\subsection{Additional Qualitative Results}
   \begin{figure}
     \centering
     \setlength{\tabcolsep}{2pt}
     \renewcommand{\arraystretch}{0.8}
     \newcommand{\panowidth}{0.48\linewidth}
     \newcommand{\cropwidth}{0.155\linewidth}
     \begin{tabular}{cc}
       {\textbf{Base Model}} & {\textbf{Ours}} \\[2pt]
       % --- (a) Niagara Falls rotated (seam) ---
       \includegraphics[width=\panowidth]{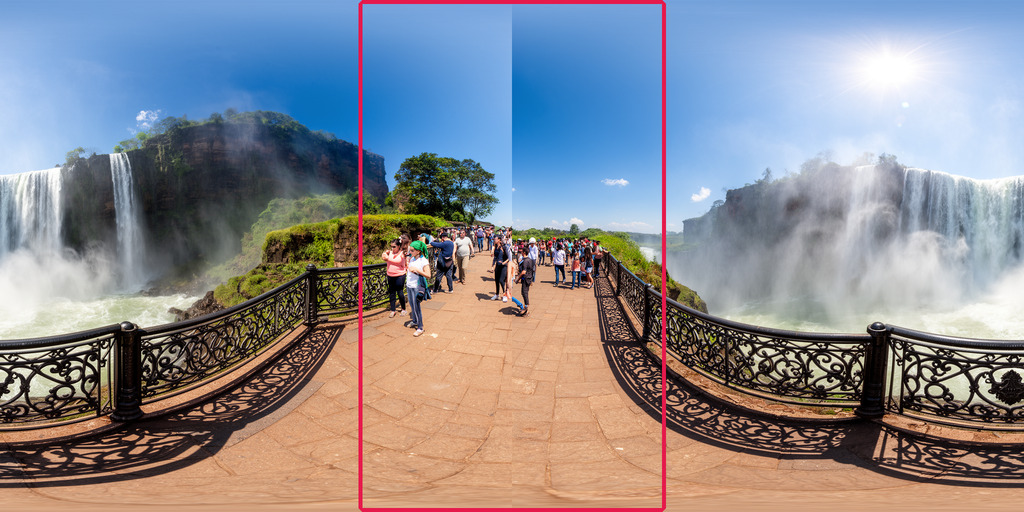} &
       \includegraphics[width=\panowidth]{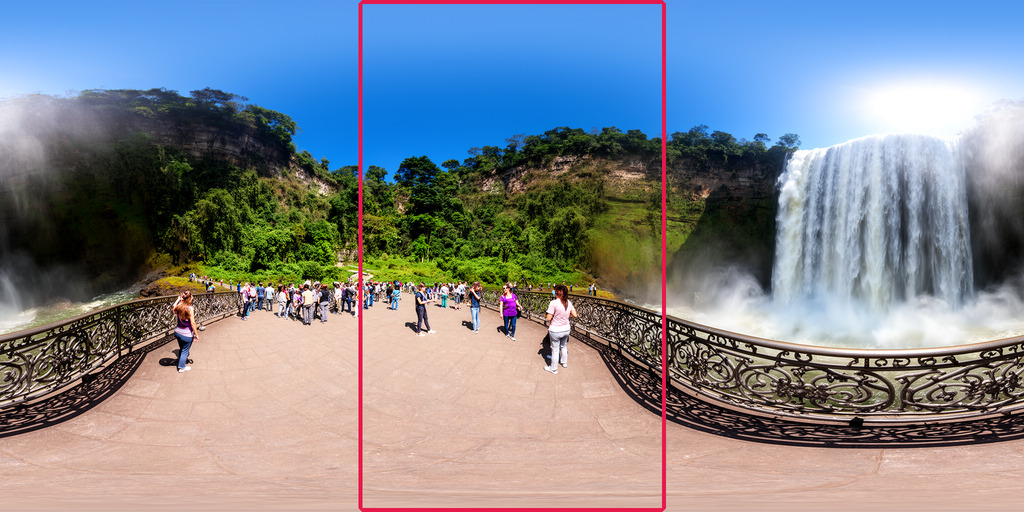} \\
       \multicolumn{2}{c}{\small (a)} \\[4pt]
       % --- (b) Studio Ghibli countryside (annotated + crops) ---
       \includegraphics[width=\panowidth]{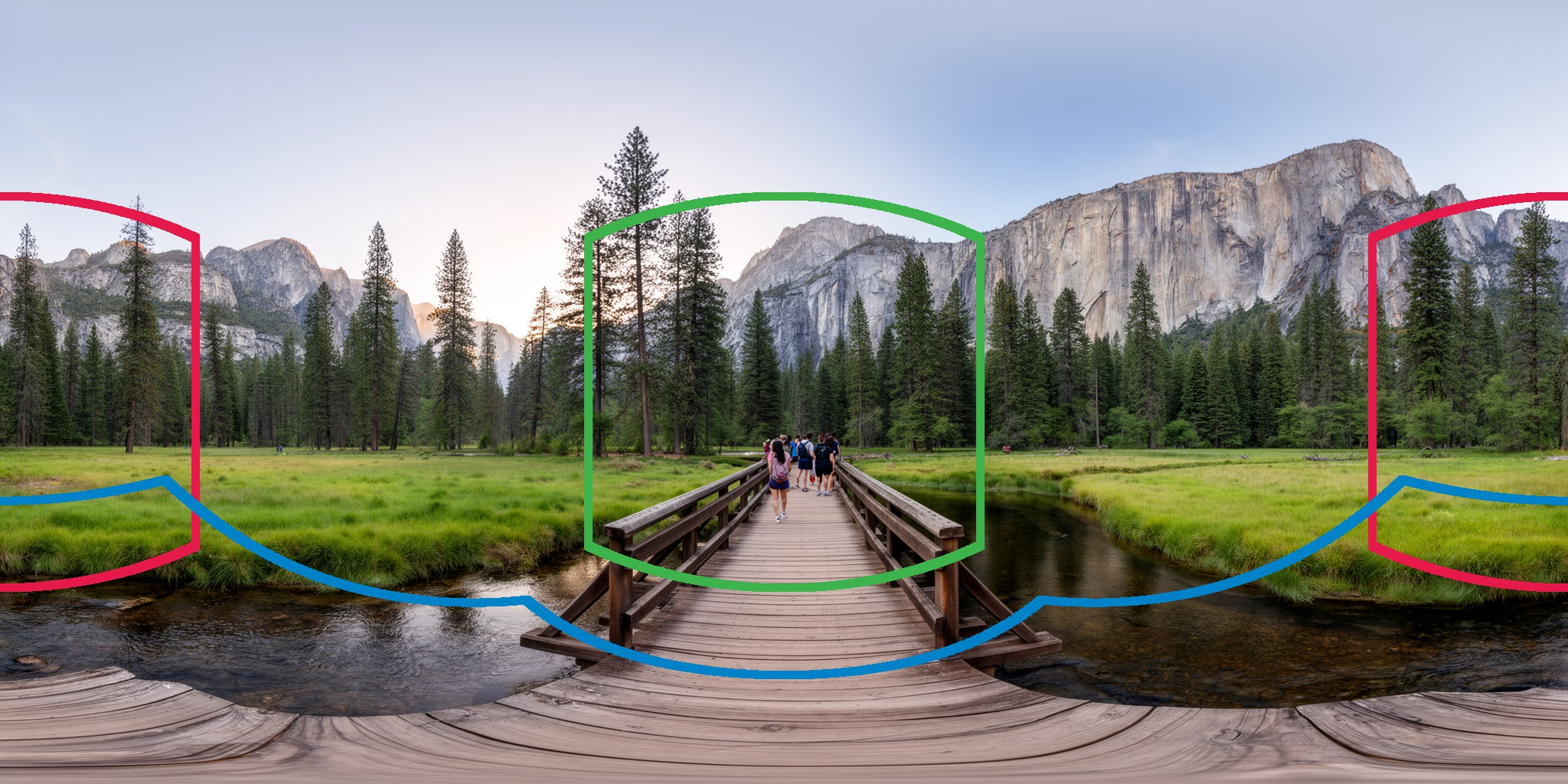} &
       \includegraphics[width=\panowidth]{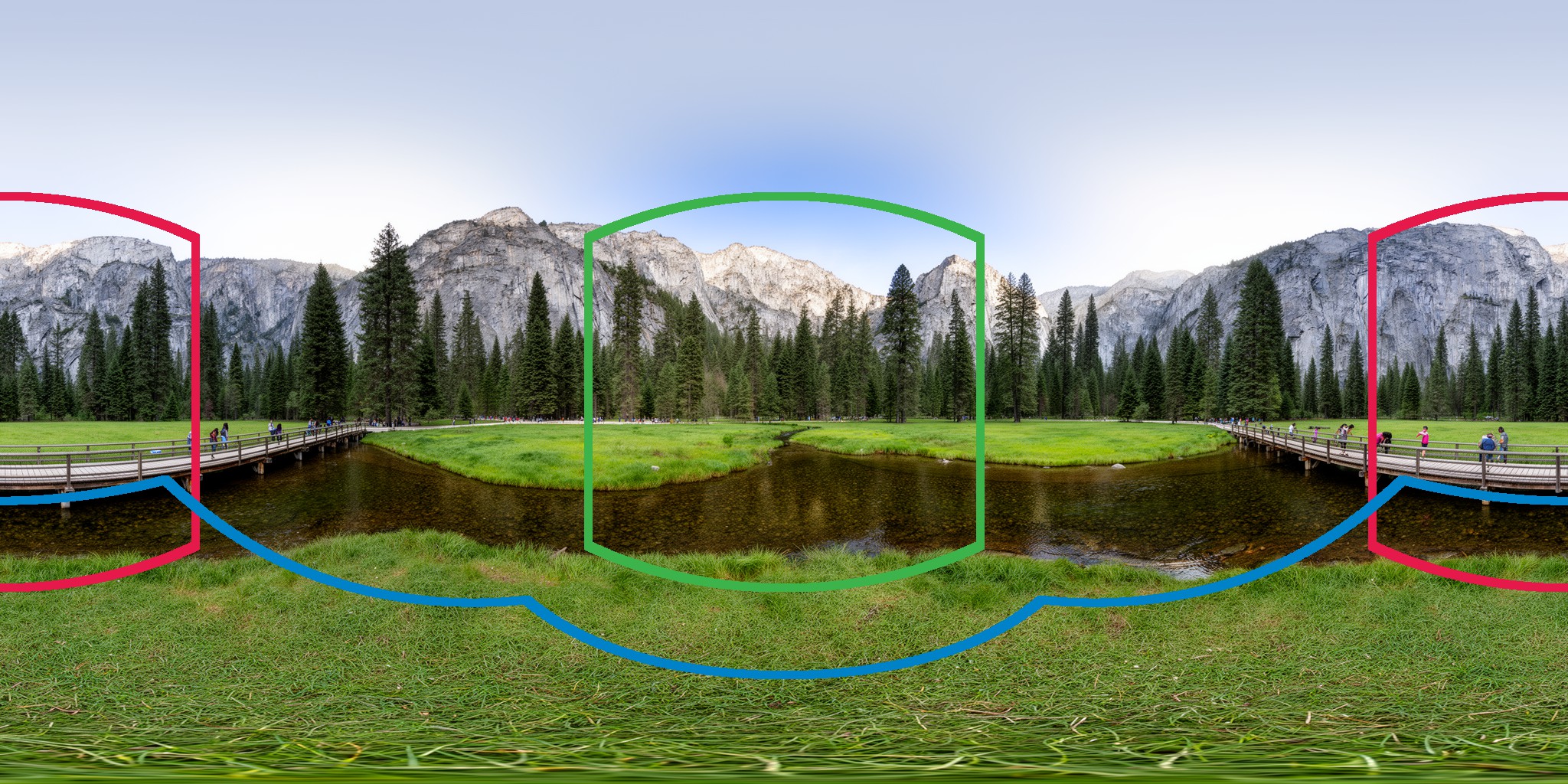} \\ %[-1pt]
       \includegraphics[width=\cropwidth]{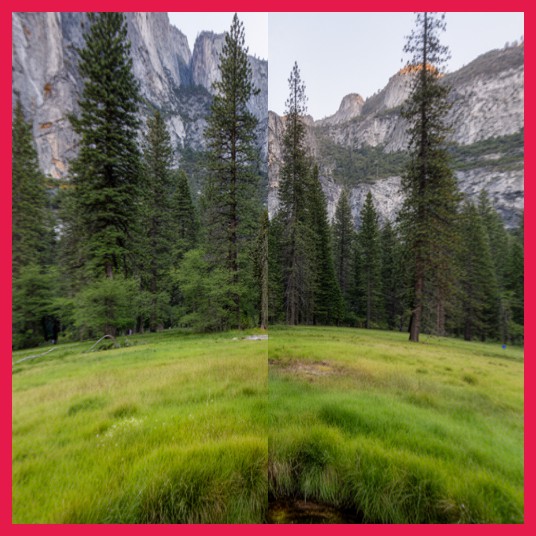}
       \includegraphics[width=\cropwidth]{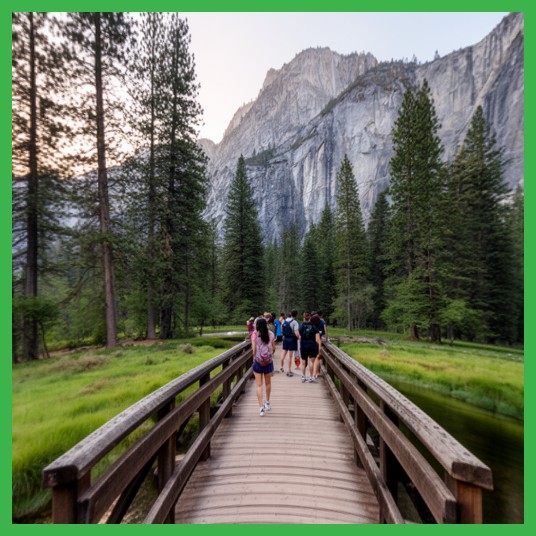}
       \includegraphics[width=\cropwidth]{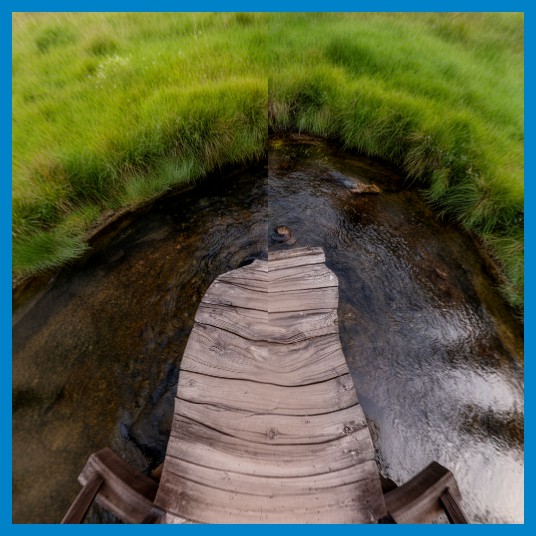} &
       \includegraphics[width=\cropwidth]{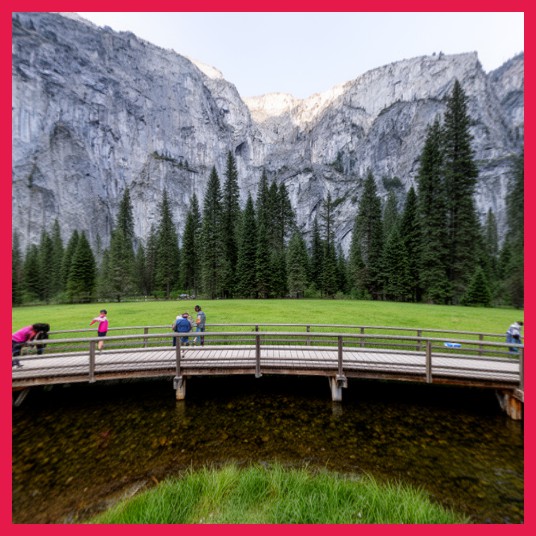}
       \includegraphics[width=\cropwidth]{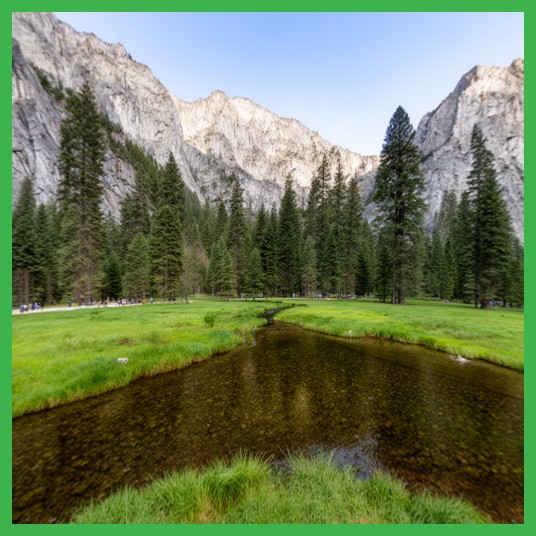}
       \includegraphics[width=\cropwidth]{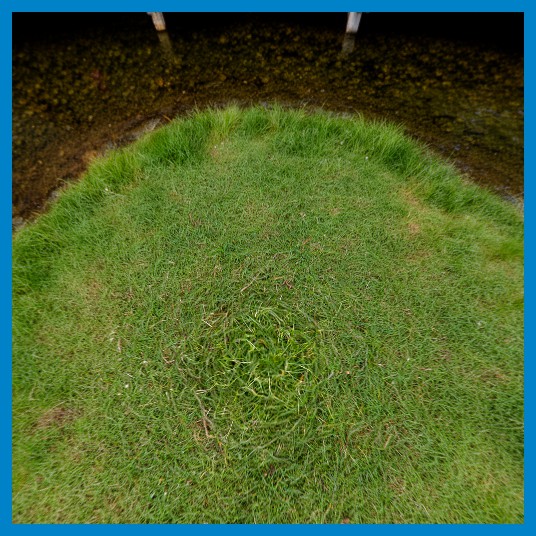} \\
       \multicolumn{2}{c}{\small (b)} \\
     \end{tabular}
     \caption{\textbf{Analysis of Implicit 360 Capabilities.} While current diffusion models capture some 360 panoramic image characteristics, they fail to satisfy spherical topology. (a) Shifting the panorama horizontally reveals a noticeable vertical seam in the base model, whereas our method maintains seamless periodicity. (b) Perspective crops show that the base model sometimes fails to properly model ERP distortions, highlighting the need for our CFG enhancement.}
     \label{fig:intuition}
   \end{figure}
\paragraph{Implicit 360 Capabilities.} Figure~\ref{fig:intuition} illustrates the core intuition behind our approach. As shown in the base model output (Flux.2), pre-trained diffusion transformers inherently capture the equirectangular projection (ERP) distribution, naturally synthesizing the stretched poles and wide aspect ratios characteristic of $360^\circ$ scenes. However, they lack the topological awareness to mathematically close the sphere, leaving disjointed boundaries and polar discontinuities. Our framework provides the critical missing link: Semantic Distortion CFG amplifies the model's latent panoramic priors, while Spherical RoPE explicitly enforces horizontal periodicity. Together, these mechanisms gently guide the base model's native ERP distribution into a seamless panorama.

\subsection{More Ablation Studies}
\label{supp:ablation}

   \newcommand{\ablfigpath}{figures/ablation_components}

   \begin{figure*}[t]
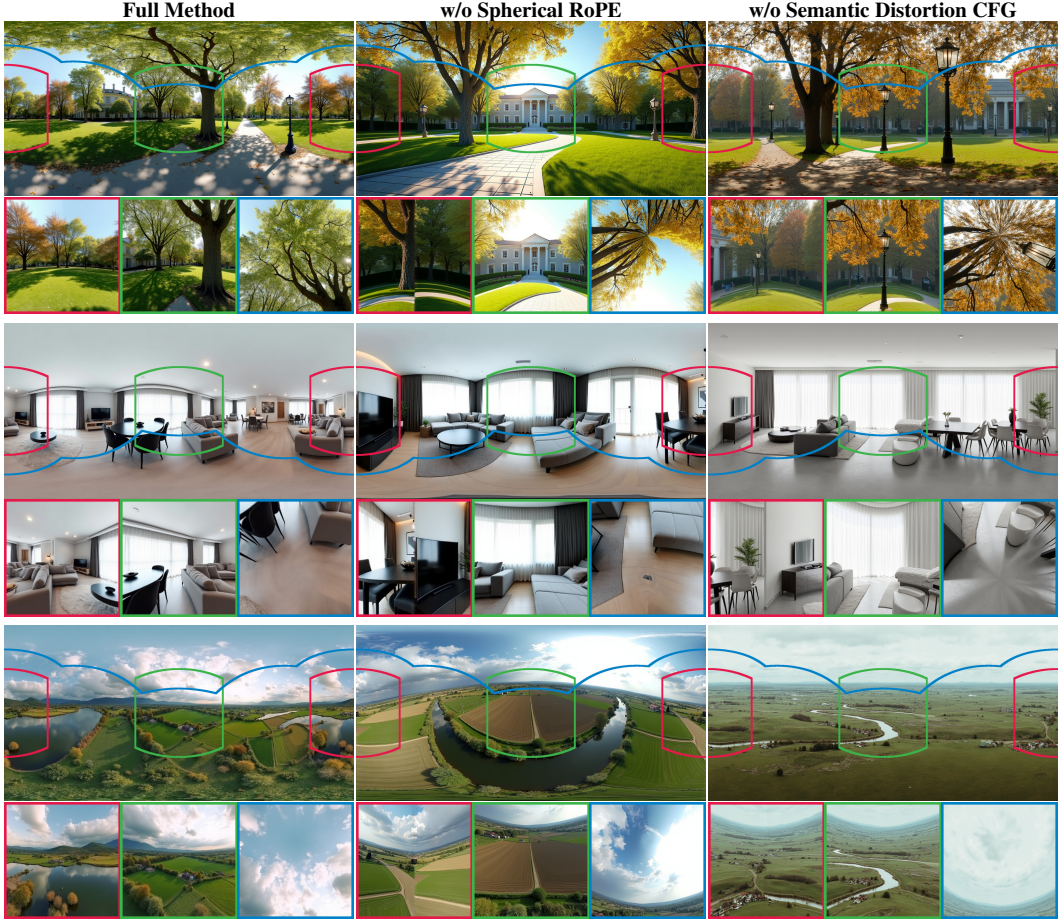

     \centering
     \setlength{\tabcolsep}{1pt}
     \resizebox{\linewidth}{!}{%
     \begin{tabular}{ccc}
     {\LARGE\textbf{Full Method}} & {\LARGE\textbf{w/o Spherical RoPE}} & {\LARGE\textbf{w/o Semantic Distortion CFG}} \\
     \methodcell{\ablfigpath}{full_1}
     & \methodcell{\ablfigpath}{no_rope_1}
     & \methodcell{\ablfigpath}{no_erpcfg_1} \\
     \noalign{\vspace{8pt}}
     \methodcell{\ablfigpath}{full_2}
     & \methodcell{\ablfigpath}{no_rope_2}
     & \methodcell{\ablfigpath}{no_erpcfg_2} \\
     \noalign{\vspace{8pt}}
     \methodcell{\ablfigpath}{full_3}
     & \methodcell{\ablfigpath}{no_rope_3}
     & \methodcell{\ablfigpath}{no_erpcfg_3} \\
     \end{tabular}%
     }
     \caption{\textbf{Component ablation.} Generated using Flux.1. Each column shows the full ERP panorama (top) and three perspective crops (bottom): seam region (\textcolor{red}{red}), horizon (\textcolor{green}{green}), and
   zenith/nadir (\textcolor{blue}{blue}). Without spherical RoPE, the output is a flat perspective image with a hard seam discontinuity and no polar convergence. Without Semantic Distortion CFG, RoPE ensures seamless boundaries but
  the content remains perspective-like, missing the pole stretching and horizon curvature expected in ERP.}
     \label{fig:ablation_components}
   \end{figure*}
\paragraph{Components Ablation.}  
We extend the component ablation from the main paper (Table~\ref{tab:ablation}) with qualitative results on FLUX.1 (Figure~\ref{fig:ablation_components}).
As shown, without spherical RoPE (middle column), the output lacks spherical characteristics - the seam crop (\textcolor{red}{red}) reveals a hard discontinuity where the left and right boundaries depict entirely different content, and the zenith crop (\textcolor{blue}{blue}) shows artifacts rather than the characteristic polar convergence of a valid ERP. Without Semantic Distortion CFG (right column), the seam is more continuous thanks to RoPE's periodicity encoding, and the zenith exhibits better convergence, but the overall scene lacks the geometric distortion patterns expected in equirectangular imagery, resulting in degraded distributional metrics.

\begin{figure}[t]
\centering
\setlength{\tabcolsep}{2pt}
\begin{tabular}{ccc}
{\small Cyclic linear} & {\small Spherical Cartesian} & {\small \textbf{Ours}} \\
\includegraphics[width=0.32\linewidth]{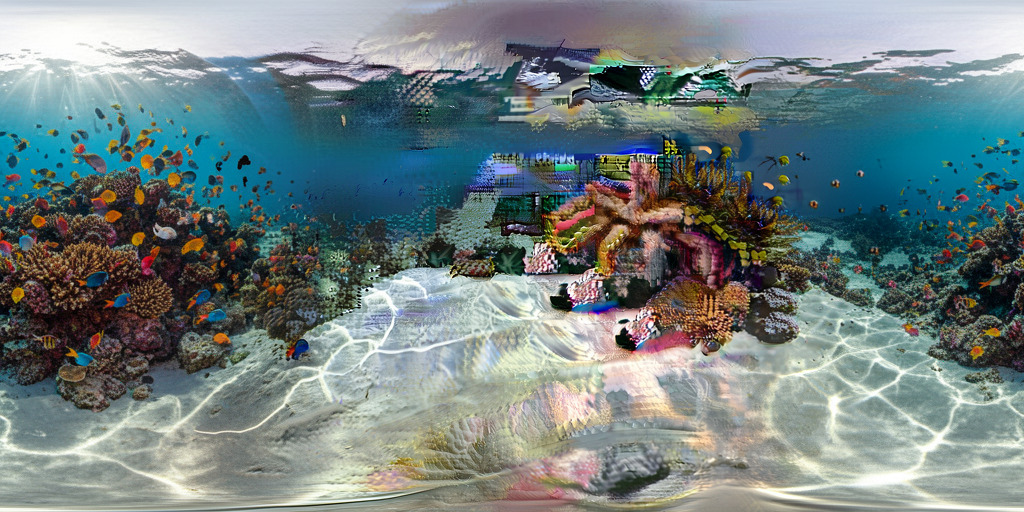} &
\includegraphics[width=0.32\linewidth]{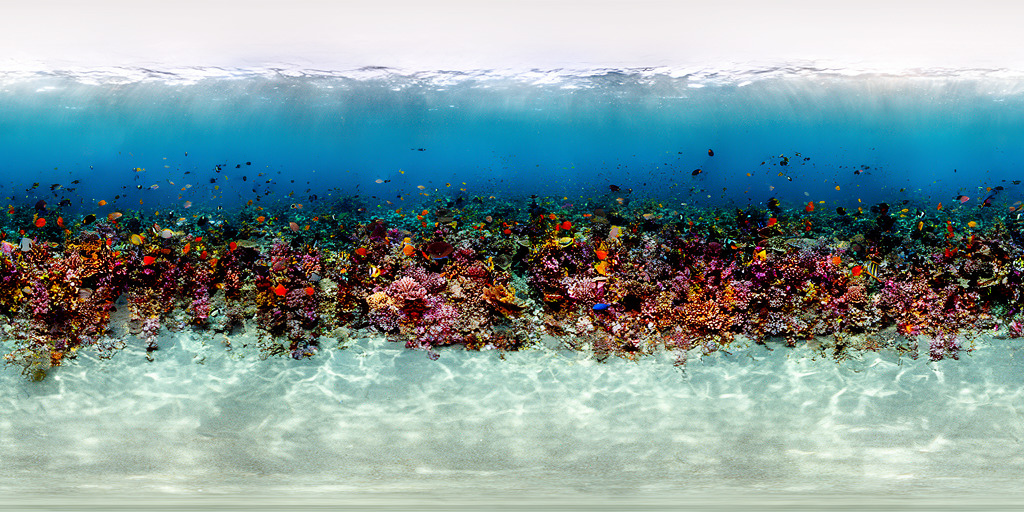} &
\includegraphics[width=0.32\linewidth]{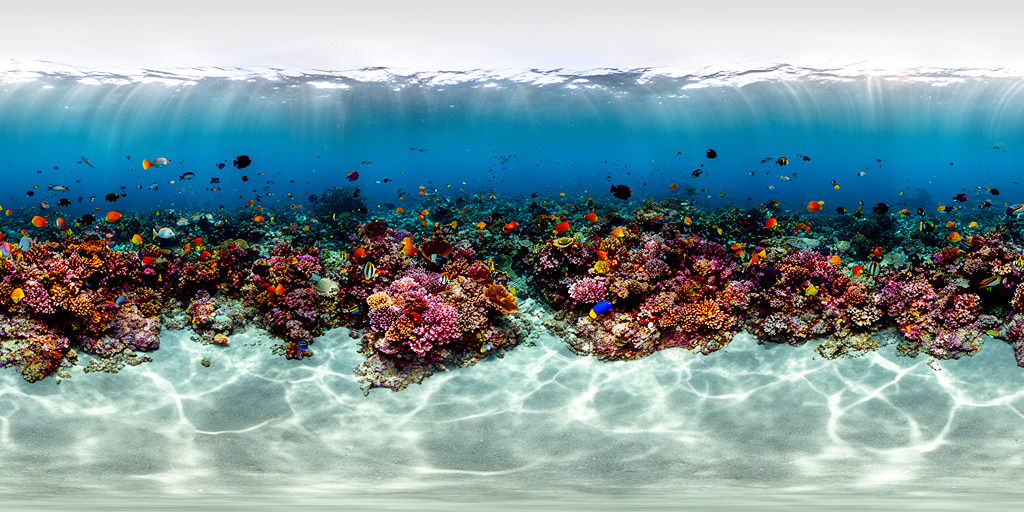} \\
\end{tabular}
\caption{\textbf{Spherical RoPE ablation.} (a) Using only Cyclic Linear enforces horizontal periodicity but yields out-of-distribution low-frequency values, resulting in artifacts. (b) Using only Spherical Cartesian satisfies global spherical topology and polar convergence but disrupts local distance metrics, causing blurry or aliased textures. (c) By partitioning the spectrum and using both strategies, we preserve high-frequency texture coherence while anchoring the global spherical manifold.}
\label{fig:ablation_rope}
\end{figure}

\paragraph{RoPE Ablation.} Figure~\ref{fig:ablation_rope} provides a comprehensive visual ablation that clearly demonstrates the necessity of our dual-path RoPE encoding strategy. When the model relies exclusively on cyclic linear encoding, it successfully achieves seamless horizontal wrapping. However, this comes at the cost of severe artifacts. This failure occurs because the large change in low frequencies RoPE is OOD for the pre-trained model, causing global structural priors to break down. 
On the other hand, applying only spherical Cartesian encoding successfully enforces the correct global topology and polar convergence. Yet, this approach severely disrupts the relative distance priors that are essential for fine-grained, high-frequency synthesis, ultimately degrading local texture quality and yielding blurry or heavily aliased details. Our proposed spectral partitioning mechanism elegantly resolves this dichotomy by assigning distinct functional roles to each encoding type. By routing the representations through a dual path, the linear encoding is leveraged exclusively to preserve crisp, sharp local textures, while the spherical encoding serves to firmly anchor the coherent global layout of the panorama.

\newcommand{\epspath}{figures/ablation_eps}

\begin{figure*}[t]
 \centering
 \setlength{\tabcolsep}{1pt}
 \renewcommand{\arraystretch}{0.5}
 \begin{tabular}{c @{\hspace{8pt}} cc}
 & {\small $\epsilon = 0.01$} & {\small $\epsilon = 0.03$} \\
 \rotatebox[origin=c]{90}{\small\textbf{FLUX.2}}
 & \includegraphics[width=0.4\linewidth]{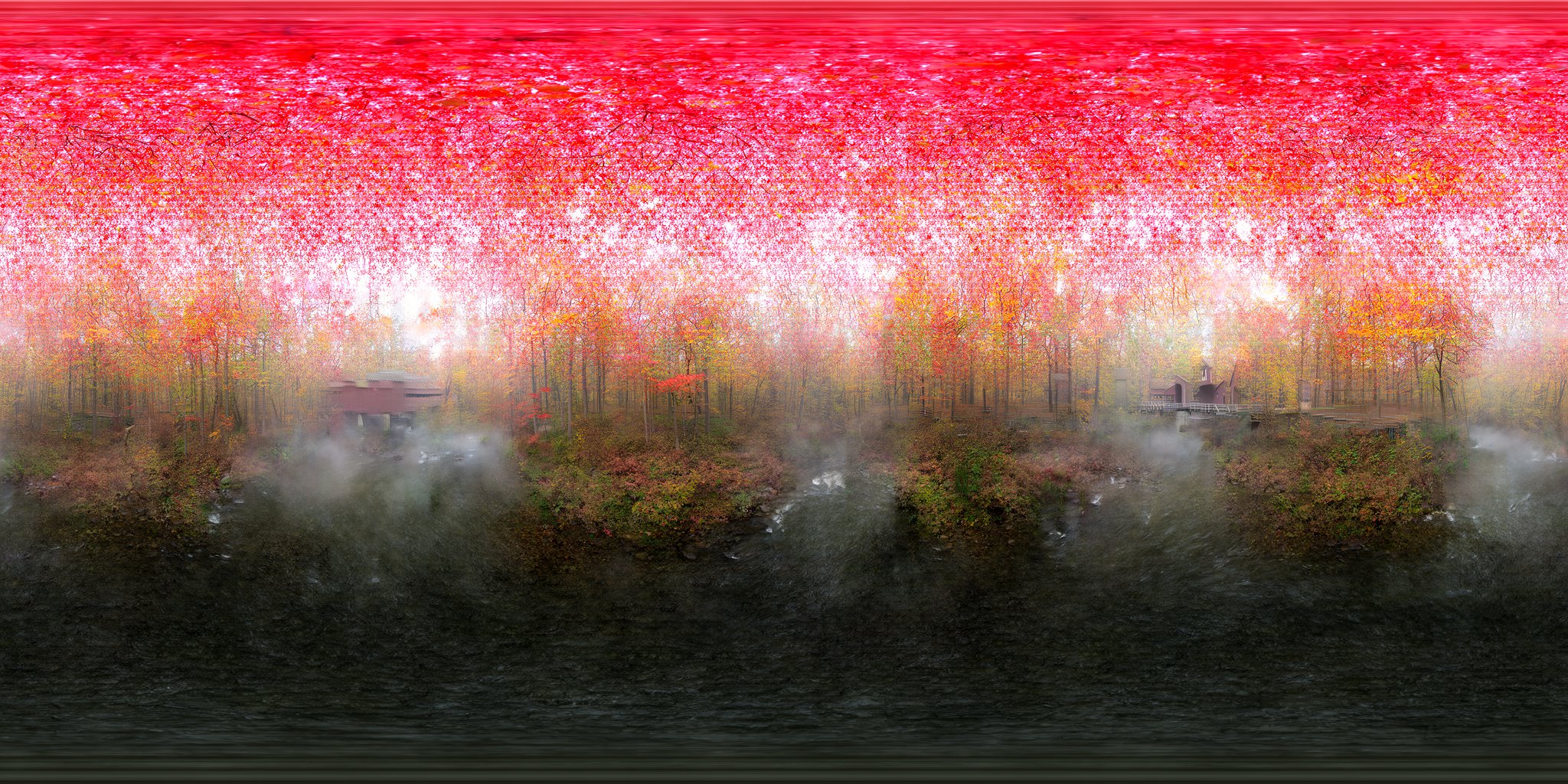}
 & \includegraphics[width=0.4\linewidth]{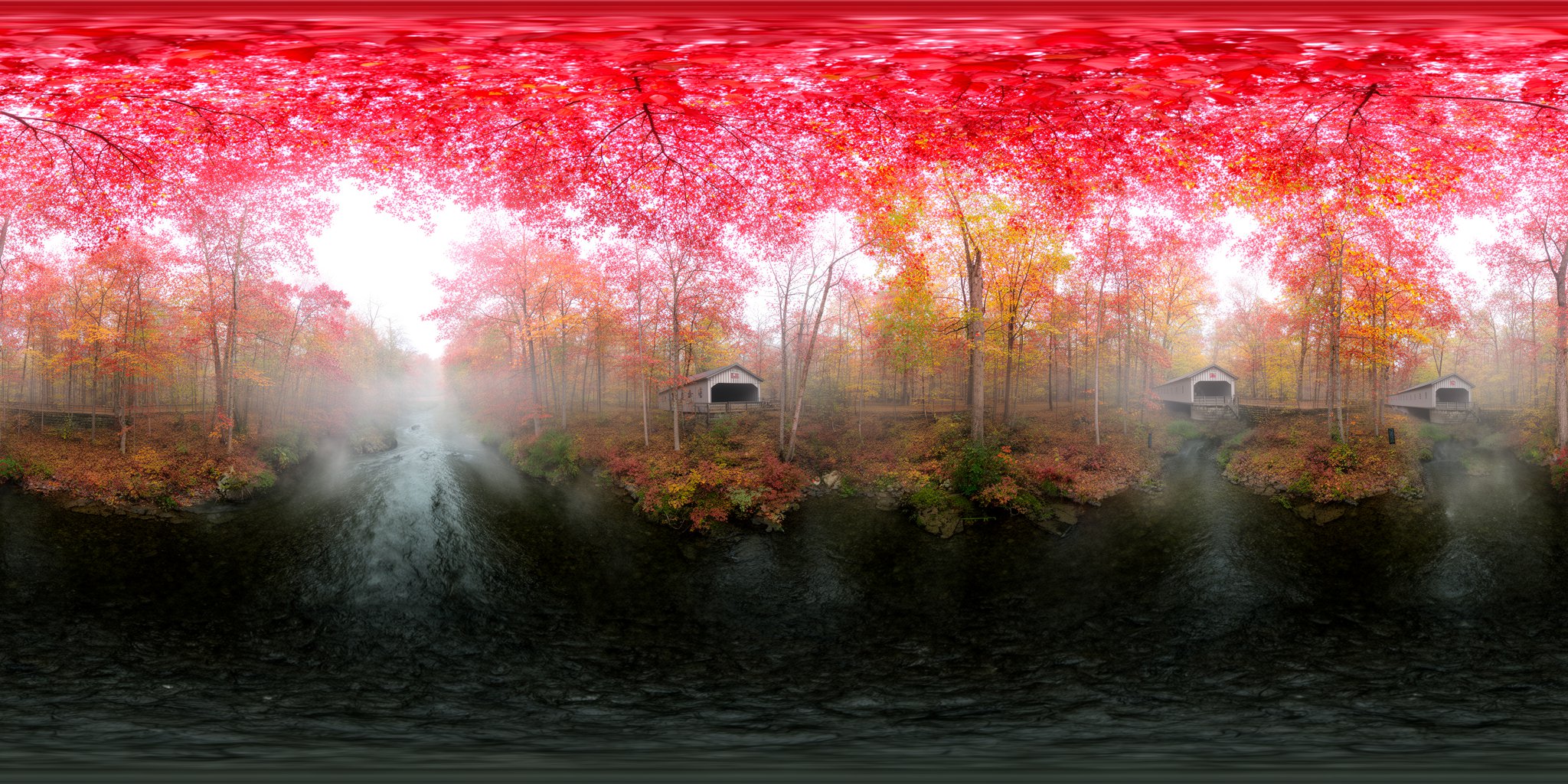} \\
 \noalign{\vspace{-2ex}}
 & {\small $\epsilon = 0.06$} & {\small $\epsilon = 0.10$} \\
 & \includegraphics[width=0.4\linewidth]{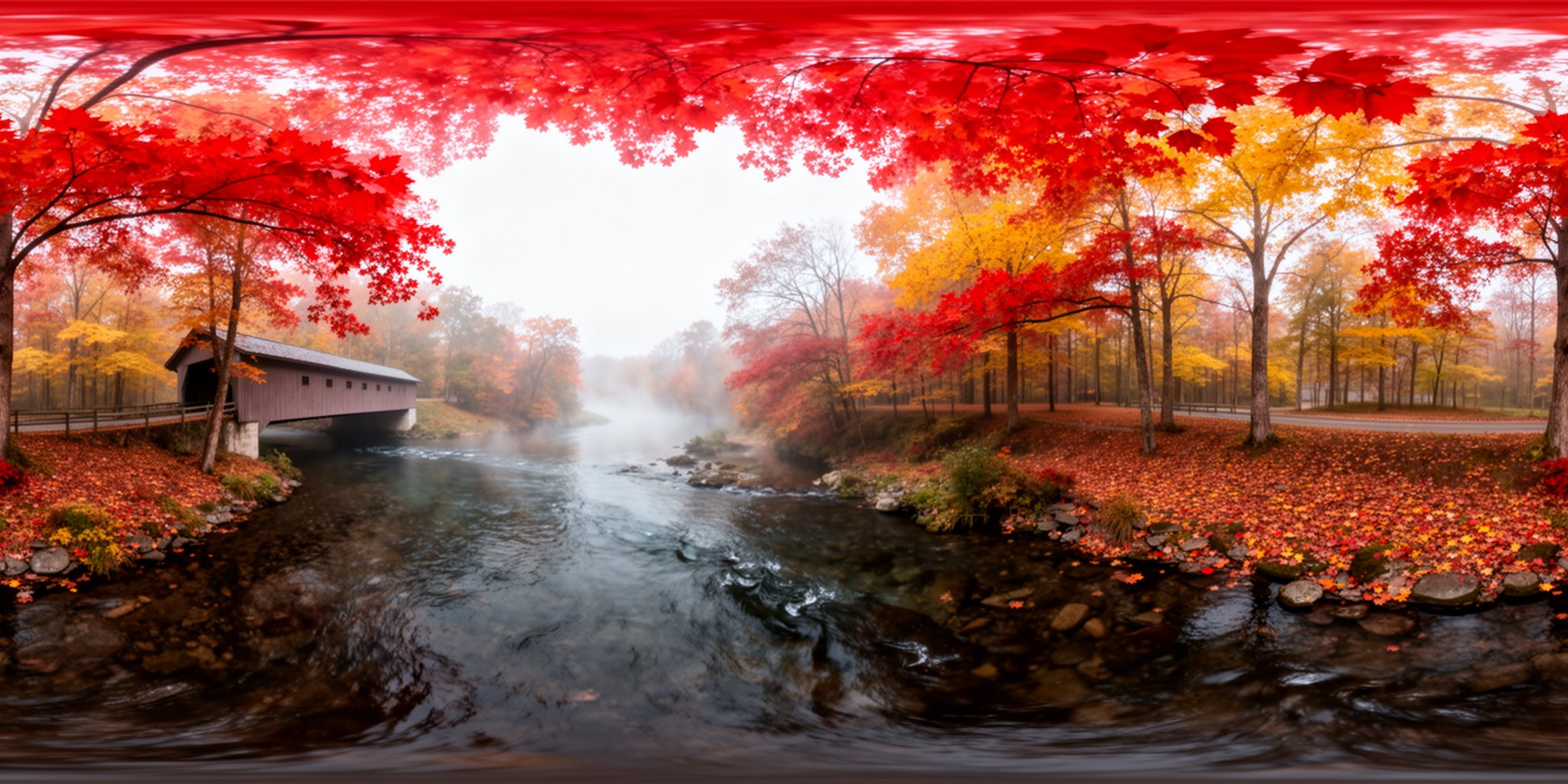}
 & \includegraphics[width=0.4\linewidth]{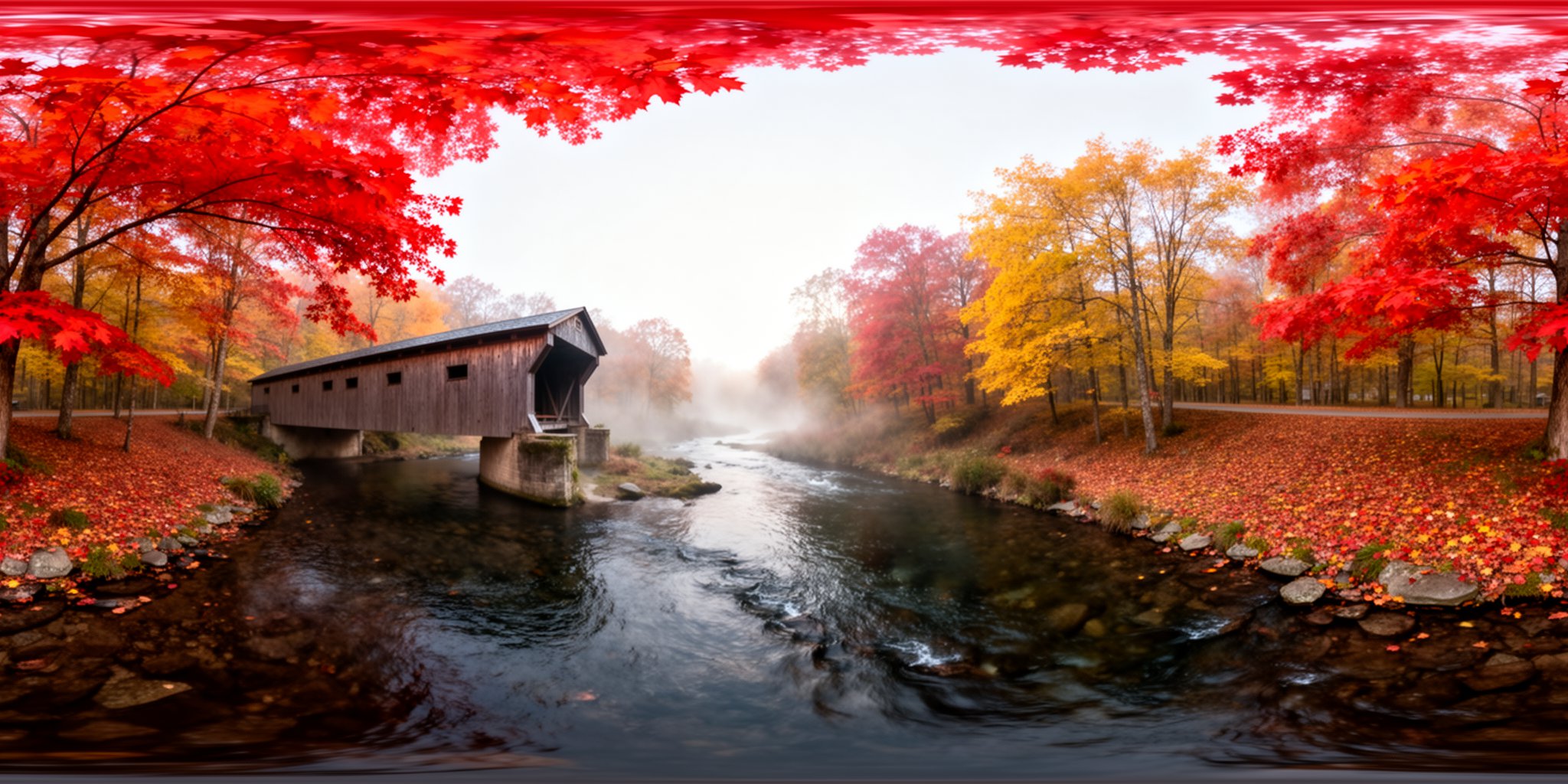} \\
 \noalign{\vspace{10pt}}
 & {\small $\epsilon = 0.01$} & {\small $\epsilon = 0.03$} \\
 \rotatebox[origin=c]{90}{\small\textbf{FLUX.1}}
 & \includegraphics[width=0.4\linewidth]{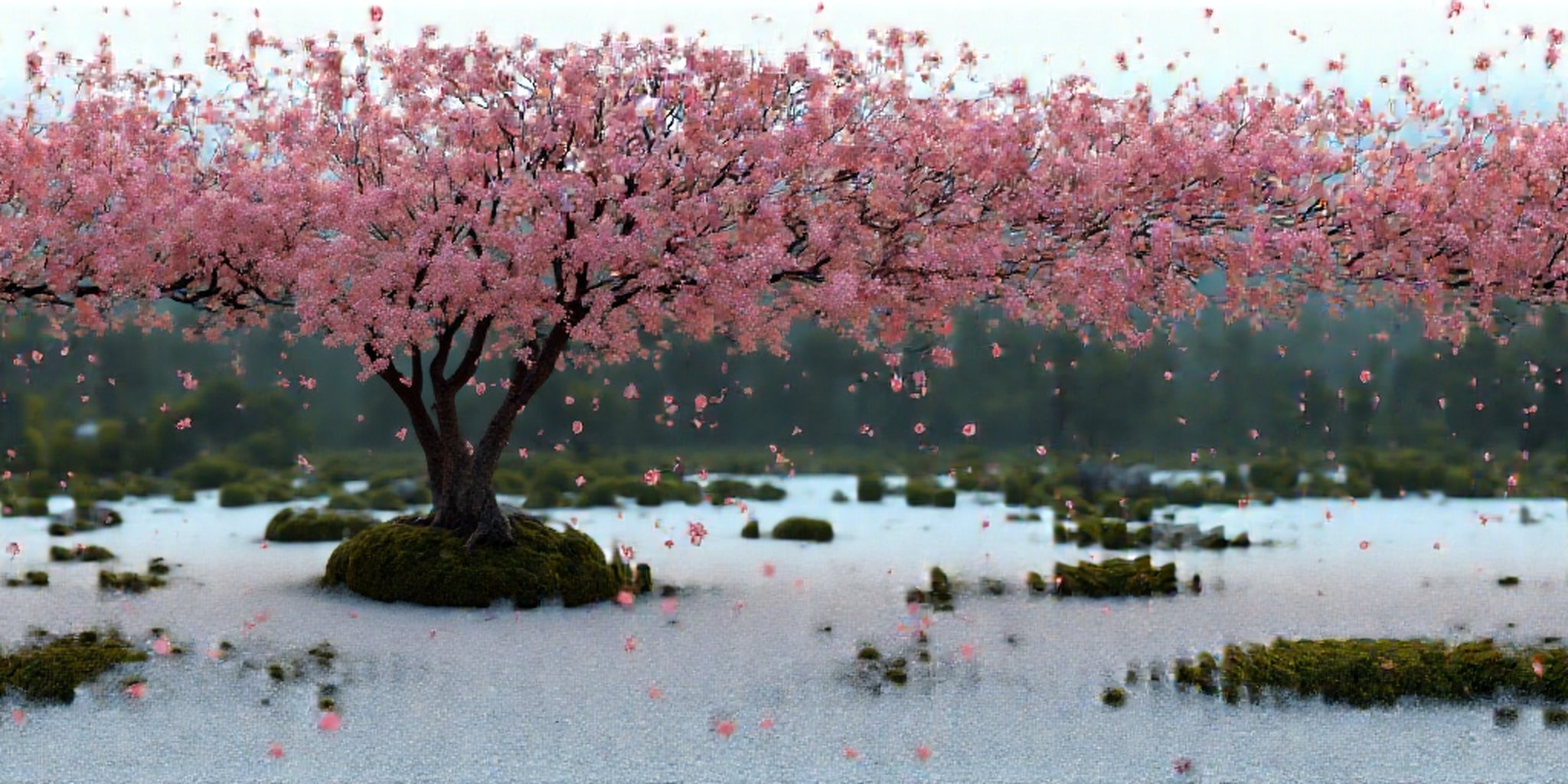}
 & \includegraphics[width=0.4\linewidth]{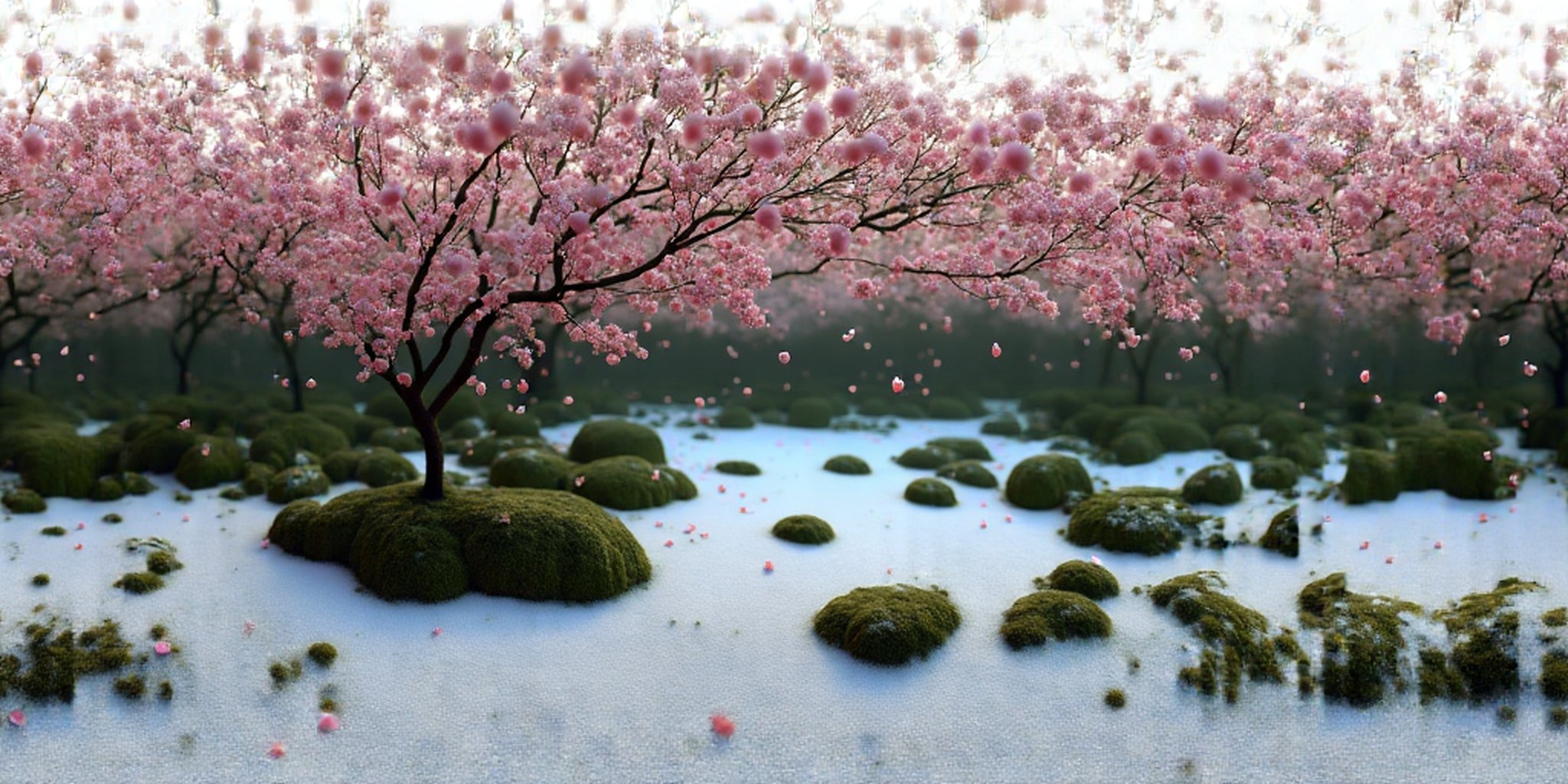} \\
 \noalign{\vspace{-2ex}}
 & {\small $\epsilon = 0.06$} & {\small $\epsilon = 0.10$} \\
 & \includegraphics[width=0.4\linewidth]{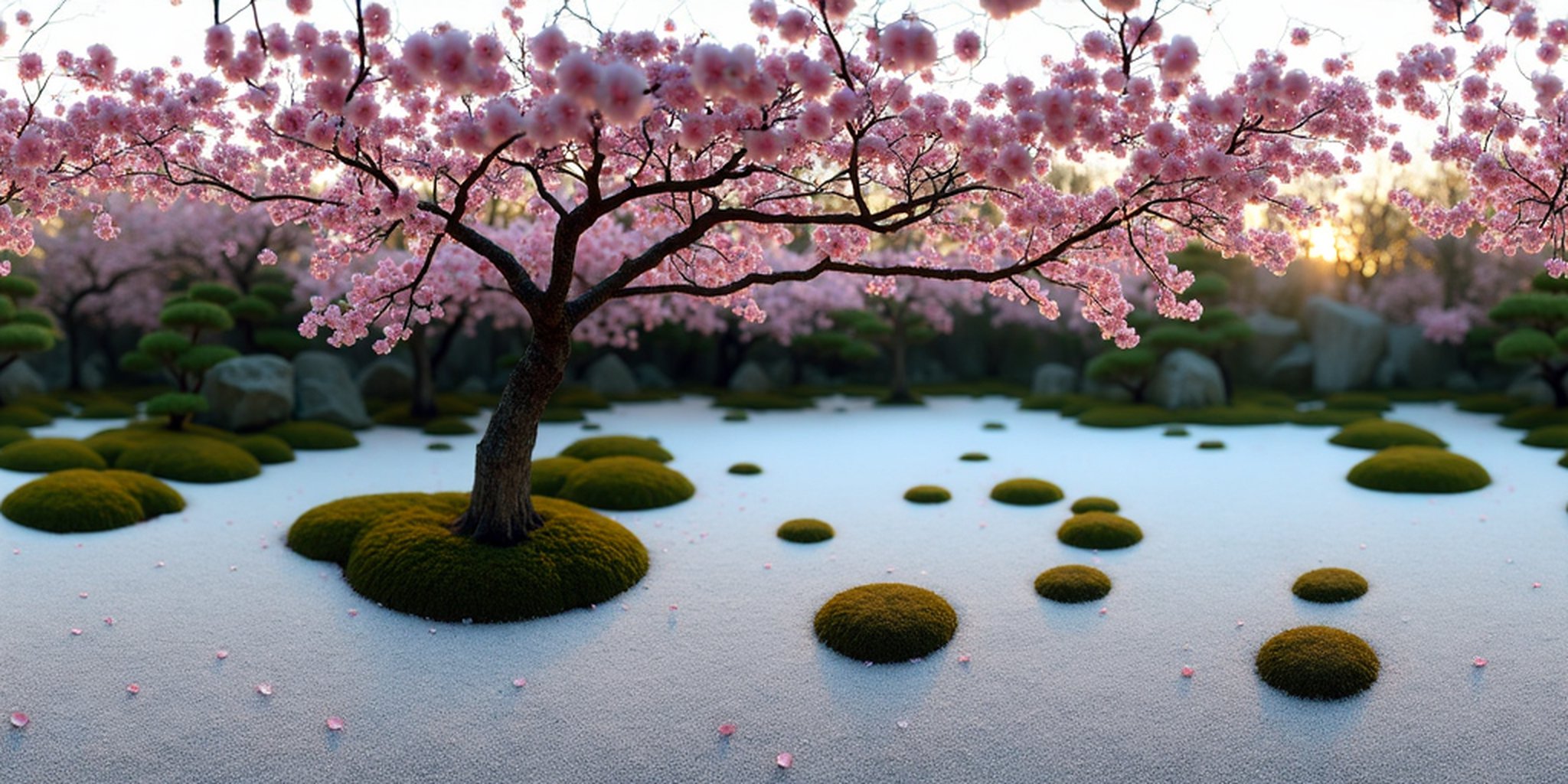}
 & \includegraphics[width=0.4\linewidth]{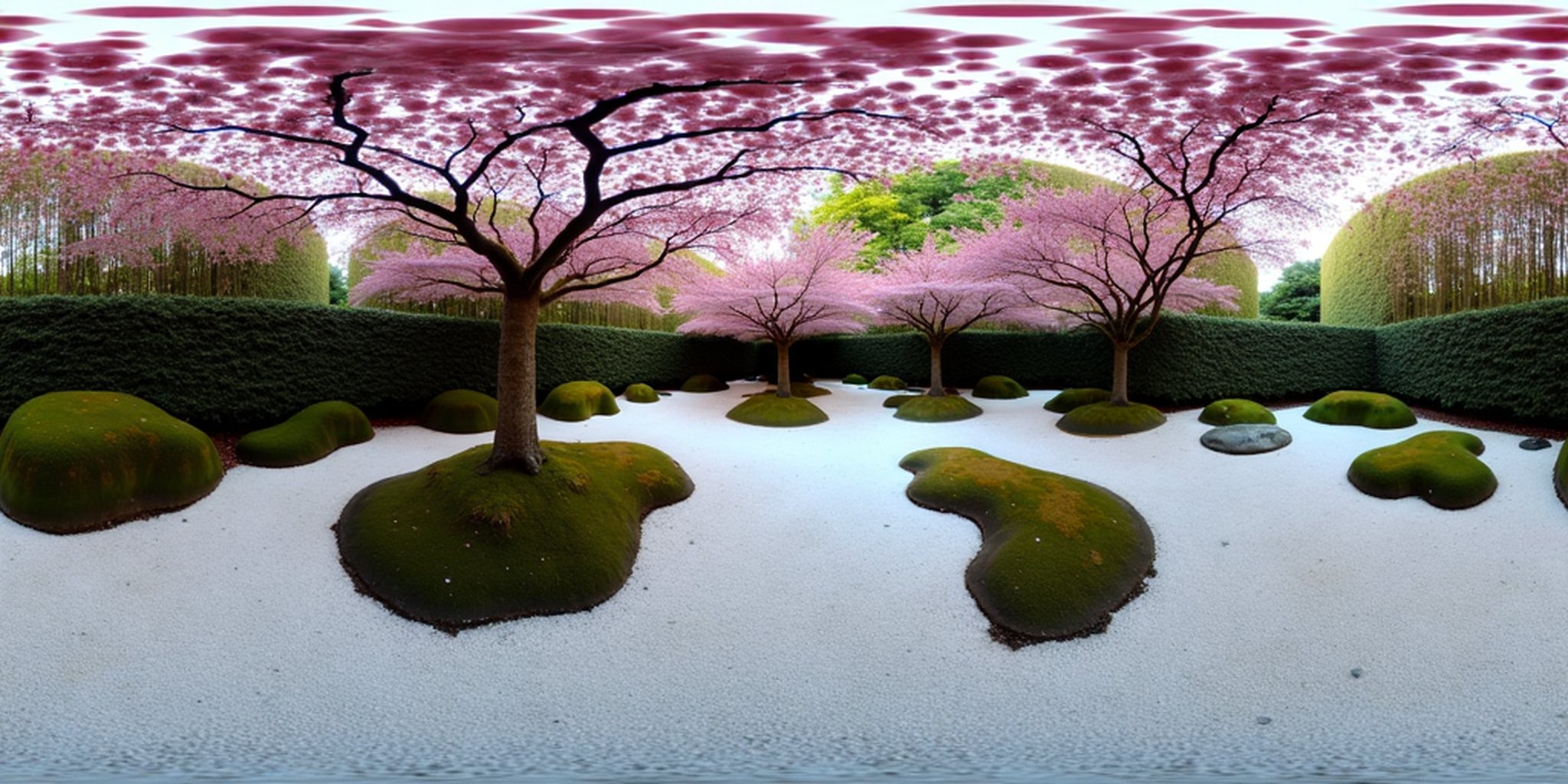} \\
 \end{tabular}
 \caption{\textbf{Harmonic quantization tolerance $\epsilon$.} We vary the tolerance used to partition RoPE frequencies between the cyclic linear path (harmonically quantizable within $\epsilon$) and the spherical Cartesian. Results shown for both FLUX.2 (top) and FLUX.1 (bottom) backbones.}
 \label{fig:eps_ablation}
\end{figure*}
\paragraph{Harmonic Quantization Tolerance.} SpheRoPE partitions the RoPE channels between the Cyclic Linear and Spherical Cartesian encodings based on whether frequencies $\omega_i$ can be harmonically quantized within a tolerance $\varepsilon$. Frequencies are assigned to the Cyclic Linear path as long as they complete at least one full cycle and are near-integer ($|k_i - \text{round}(k_i)| / k_i \leq \varepsilon$). The first channel index $i_{\text{split}}$ that violates this condition determines the boundary, with all subsequent lower frequencies ($i \ge i_{\text{split}}$) routed to the Spherical Cartesian encoding. 
The tolerance $\varepsilon$ therefore controls this split: a smaller value triggers the violation earlier, while a larger value extends the cyclic linear path.
Crucially, this mechanism has a natural saturation point. Increasing the tolerance beyond a certain threshold (e.g., $\varepsilon \ge 0.10$) yields an identical channel split, as all remaining low frequencies inherently fail the minimum cycle requirement ($k_i < 1.0$) and are strictly routed to the spherical path.
Figure~\ref{fig:eps_ablation} shows a sensitivity analysis across $\varepsilon \in \{0.01, 0.03, 0.06, 0.10\}$ on both FLUX.2 and FLUX.1. A very strict tolerance ($\varepsilon{=}0.01$) triggers an early split, forcing too many channels into the spherical encoding and producing blurrier textures due to the coarser positional resolution of the spherical path. A loose tolerance ($\varepsilon{=}0.10$) delays the split, however the minimum cycle requirement makes sure we avoid OOD artifacts. 
% Our default ($\varepsilon{=}0.06$) identifies the optimal $i_{\text{split}}$, consistently producing the sharpest and most geometrically faithful panoramas across both backbones.

% \begin{table}
% \centering
% \caption{\textbf{Semantic Distortion Strength.} We sweep the distortion gamma $\gamma$ controlling how aggressively ERP-aware CFG re-weights patches by their spherical distortion. Best in \textbf{bold}, second
% best \underline{underlined}.}
% \label{tab:flux2_gamma_sweep}
% \resizebox{\columnwidth}{!}{%
% \begin{tabular}{l|ccc|cccc}
% % \toprule
% & \multicolumn{3}{c|}{\textbf{Panorama-Level Metrics}} & \multicolumn{4}{c}{\textbf{Multi-View Metrics}} \\
% \textbf{Configuration} & FAED $\downarrow$ & OmniFID $\downarrow$ & DS $\downarrow$ & FID $\downarrow$ & KID$_{\times 10^2}$ $\downarrow$ & IS $\uparrow$ & CS $\uparrow$ \\
% \midrule
% $\gamma = 3$  & 34.48 & \textbf{83.05} & 0.90 & 29.05 & 1.15 & \textbf{13.43} & \textbf{18.46} \\
% $\gamma = 4$  & 34.08 & \underline{83.64} & 0.88 & \underline{28.58} & 1.23 & \underline{13.07} & \textbf{18.46} \\
% $\gamma = 8$  & \underline{32.20} & 86.87 & \textbf{0.85} & \textbf{28.06} & \textbf{0.98} & 11.95 & \underline{18.46} \\
% $\gamma = 10$ & \textbf{31.75} & 86.63 & \underline{0.85} & 28.70 & \underline{1.06} & 11.59 & 18.44 \\
% \bottomrule
% \end{tabular}%
% }
% \end{table}

  \begin{table}
  \centering
  \caption{\textbf{Semantic Distortion Strength.} We sweep the distortion gamma $\gamma$ controlling how aggressively ERP-aware CFG re-weights patches by their spherical distortion. Best in \textbf{bold}, second
  best \underline{underlined}.}
  \label{tab:flux2_gamma_sweep}
  \resizebox{\columnwidth}{!}{%
  \begin{tabular}{l|cc|cccc}
  % \toprule
  & \multicolumn{2}{c|}{\textbf{Panorama-Level Metrics}} & \multicolumn{4}{c}{\textbf{Multi-View Metrics}} \\
  \textbf{Configuration} & FAED $\downarrow$ & DS $\downarrow$ & FID $\downarrow$ & KID$_{\times 10^2}$ $\downarrow$ & IS $\uparrow$ & CS $\uparrow$ \\
  \midrule
  $\gamma = 3$ & 34.48 & 0.90 & 29.05 & 1.15 & \textbf{13.43} & \textbf{18.46} \\
  $\gamma = 4$ & 34.08 & 0.88 & \underline{28.58} & 1.23 & \underline{13.07} & \textbf{18.46} \\
  $\gamma = 8$ & \underline{32.20} & \textbf{0.85} & \textbf{28.06} & \textbf{0.98} & 11.95 & \underline{18.46} \\
  $\gamma = 10$ & \textbf{31.75} & \underline{0.85} & 28.70 & \underline{1.06} & 11.59 & 18.44 \\
  \bottomrule
  \end{tabular}%
  }
  \end{table}
\begin{figure}[t]
\centering
\setlength{\tabcolsep}{2pt}
\renewcommand{\arraystretch}{0.5}
\begin{tabular}{cc}
\includegraphics[width=0.42\columnwidth]{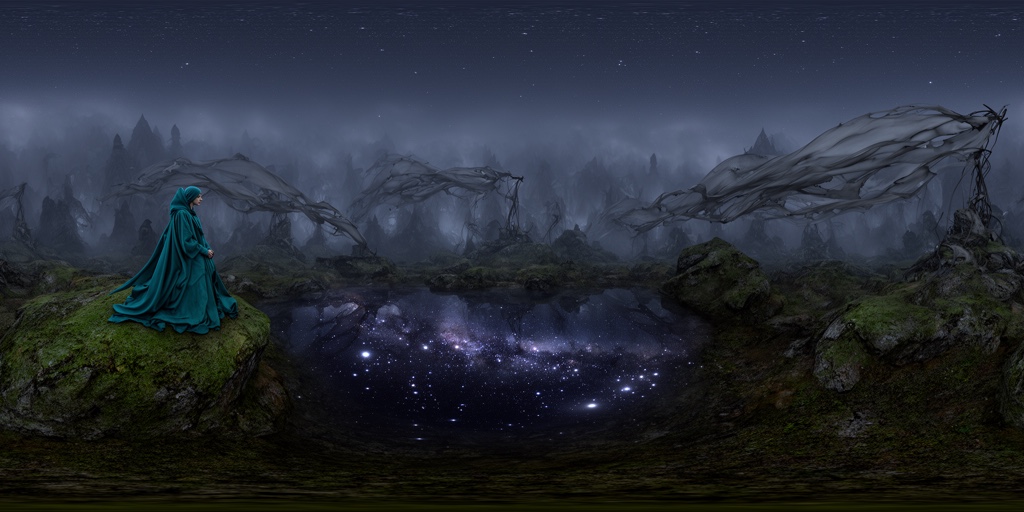} &
\includegraphics[width=0.42\columnwidth]{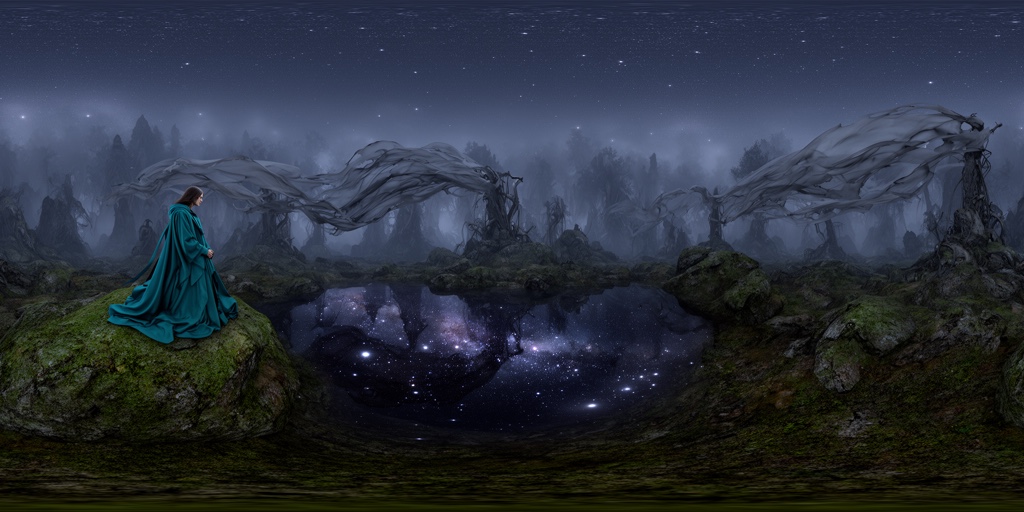} \\
{\small $\gamma = 3$} & {\small $\gamma = 4$} \\[2pt]
\includegraphics[width=0.42\columnwidth]{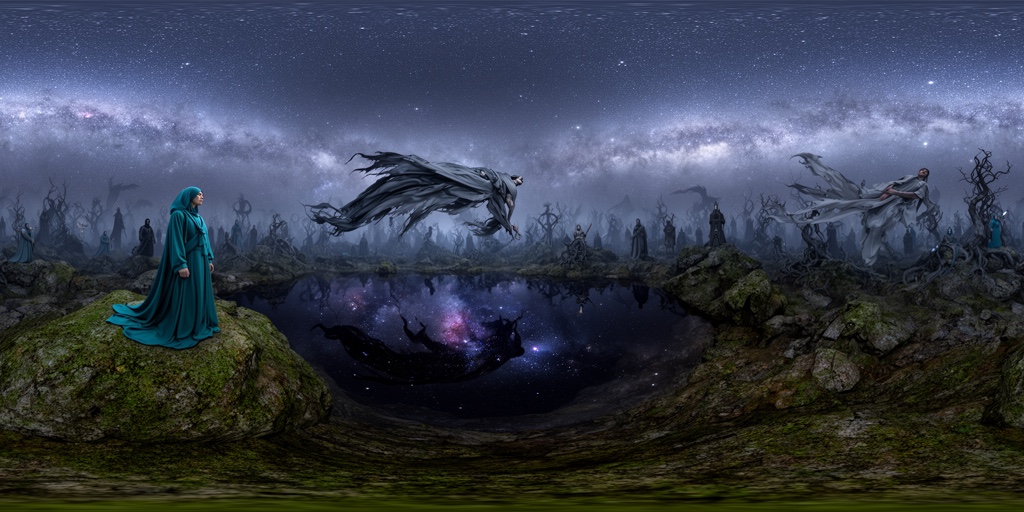} &
\includegraphics[width=0.42\columnwidth]{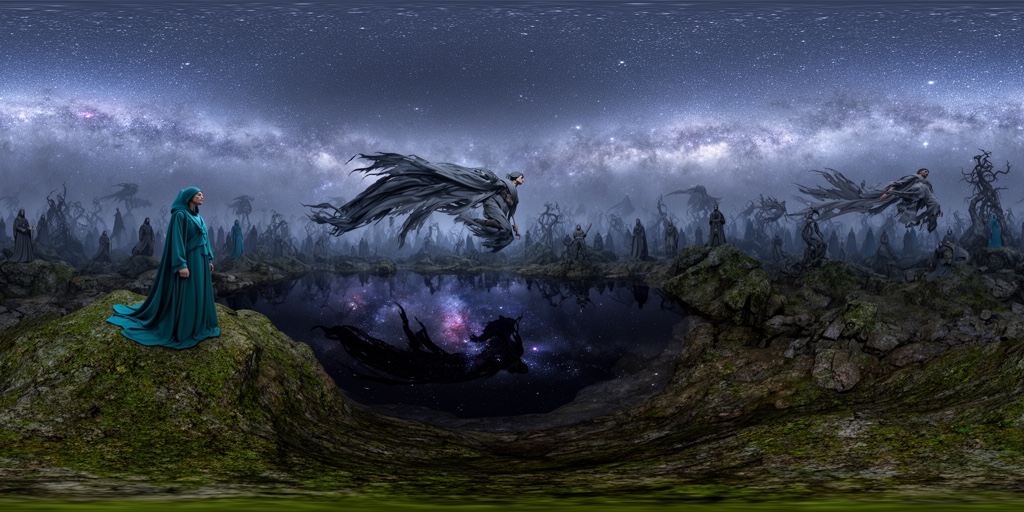} \\
{\small $\gamma = 8$} & {\small $\gamma = 10$} \\
\end{tabular}
\caption{\textbf{Visual effect of the distortion strength $\gamma$.} Panoramas generated with the
same prompt and seed under four different values of the ERP-aware CFG distortion parameter.
Visual quality and layout remain stable across the full range: differences are subtle and
primarily manifest as slightly sharper, more varied texture at lower $\gamma$ versus slightly
smoother, more geometrically-regularised content at higher $\gamma$.}
\label{fig:gamma_sweep_qualitative}
\end{figure}
\newcommand{\gammapath}{figures/ablation_gamma_scheduling}

\newcommand{\methodcells}[2]{% #1=figpath prefix, #2=method
 \parbox[c]{5cm}{%
   \includegraphics[width=5cm]{#1/#2/pano.jpg}\\[0pt]
   \includegraphics[width=1.67cm]{#1/#2/crop_0.jpg}%
   \includegraphics[width=1.67cm]{#1/#2/crop_1.jpg}%
   \includegraphics[width=1.67cm]{#1/#2/crop_2.jpg}%
 }%
}

\begin{figure*}[t]
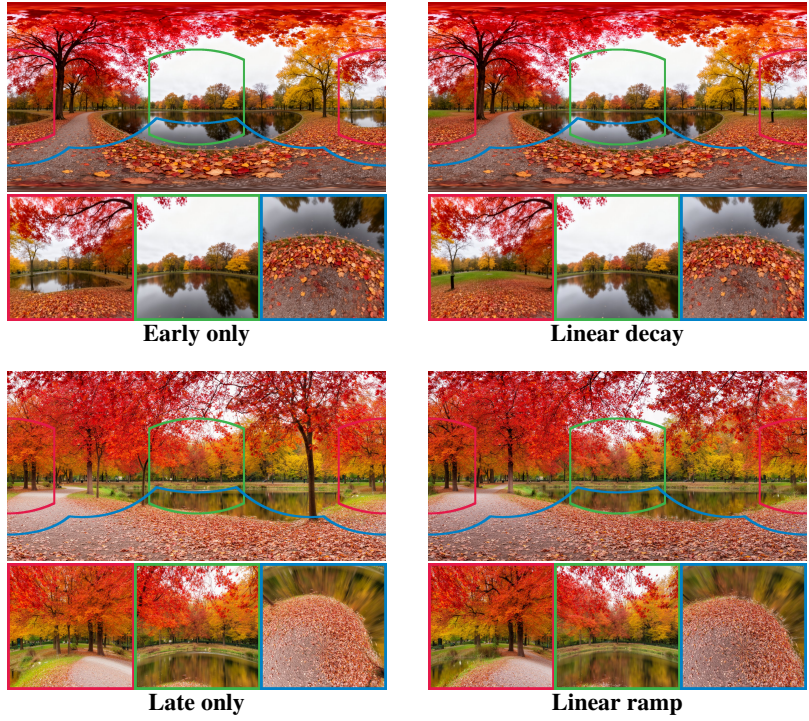

\centering
\setlength{\tabcolsep}{8pt}
\begin{tabular}{cc} 
\methodcells{\gammapath}{early_only}
& \methodcells{\gammapath}{linear_decay} \\
{\small\textbf{Early only}} & {\small\textbf{Linear decay}} \\
\noalign{\vspace{8pt}}
\methodcells{\gammapath}{late_only}
& \methodcells{\gammapath}{linear_ramp} \\
{\small\textbf{Late only}} & {\small\textbf{Linear ramp}} \\
\end{tabular}%

\caption{\textbf{Semantic Distortion CFG schedule.} We apply the geometric guidance term with different schedules over the $N$ denoising steps. The top row (\textit{early only}, \textit{linear decay})
concentrates guidance in the early steps and produces valid ERP geometry with proper polar convergence. The bottom row (\textit{late only}, \textit{linear ramp}) concentrates guidance in the late steps and yields
flat perspective-like panoramas with pole artifacts and incorrect distortion patterns.}
\label{fig:gamma_schedule}
\end{figure*}
\paragraph{Semantic Distortion CFG.}
We sweep $\gamma \in \{3, 4, 8, 10\}$ to examine how aggressively our ERP-aware CFG branch should re-weight latent patches according to their spherical distortion. Rather than producing a single clear winner, Table~\ref{tab:flux2_gamma_sweep} reveals a mild but consistent trade-off between panorama-level and multi-view metrics: weaker distortion ($\gamma = 3, 4$) maximises multi-view IS and CS, while stronger distortion ($\gamma = 8, 10$) minimises FAED, DS, and perspective FID, and KID. 
Critically, \emph{the absolute spread of each metric is narrow} - FAED varies by only $2.7$ points, FID by $1.0$, and CS is essentially flat (within $0.02$). 
This quantitative stability is mirrored qualitatively in
Figure~\ref{fig:gamma_sweep_qualitative}: panoramas generated under all four values of $\gamma$ preserve the same overall layout and perceptual quality, with differences amounting to slight local texture variations rather than structural changes. Together, these results show that our method is \emph{largely agnostic} to the precise choice of $\gamma$ within this range: any value between 3 and 10 yields comparable quality.
We adopt $\gamma = 6$ as a balanced default that sits inside this stable regime.

We further investigate the optimal scheduling for Semantic Distortion CFG across the denoising process. Figure~\ref{fig:gamma_schedule} compares four strategies: \textit{early only}, \textit{late only}, \textit{linear decay}, and \textit{linear ramp}. 
Results show that panoramic geometry is established during the initial denoising steps. Consequently, early-weighted schedules (\textit{early only}, \textit{linear decay}) successfully enforce valid ERP geometry and yield nearly identical, seamless panoramas. Conversely, late-weighted schedules (\textit{late only}, \textit{linear ramp}) fail to alter the already-crystallized global structure, resulting in the same pole artifacts and flat layouts seen in the unguided baseline. This confirms that Semantic Distortion CFG primarily shapes the early geometric trajectory rather than refining late-stage details. 
For simplicity, we choose to apply the Semantic Distortion CFG constantly across all denoising steps.

\newcommand{\radpath}{figures/ablation_r}

\begin{figure*}[t]
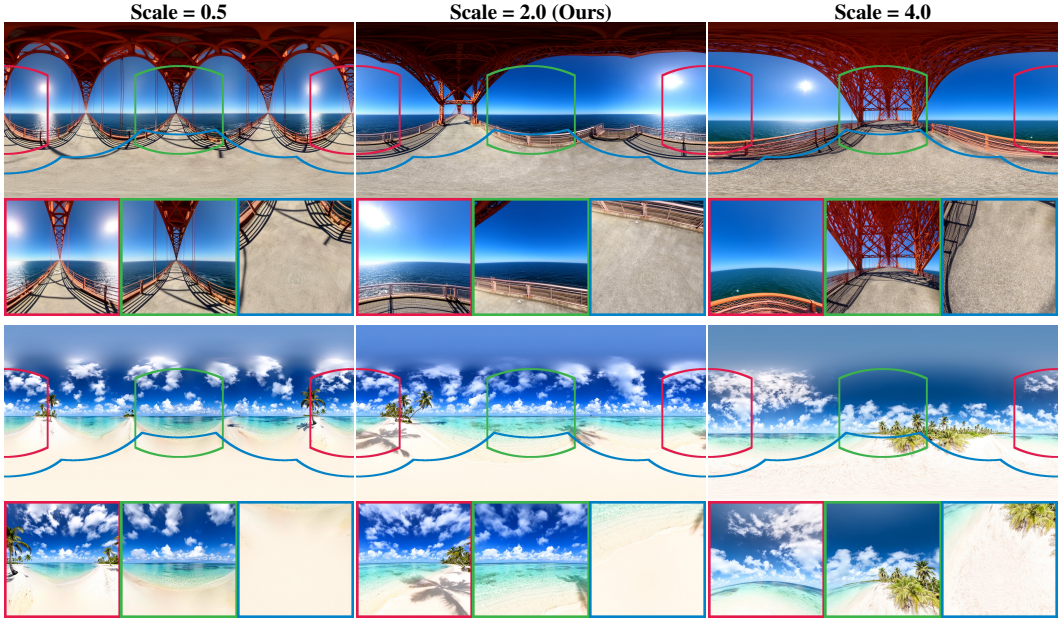

 \centering
 \setlength{\tabcolsep}{1pt}
 \resizebox{\linewidth}{!}{%
 \begin{tabular}{ccc}
 {\LARGE\textbf{Scale = 0.5}} & {\LARGE\textbf{Scale = 2.0 (Ours)}} & {\LARGE\textbf{Scale = 4.0}} \\
 \methodcell{\radpath}{golden_gate_bridge_scale0_5}
 & \methodcell{\radpath}{golden_gate_bridge_scale2_0}
 & \methodcell{\radpath}{golden_gate_bridge_scale4_0} \\
 \noalign{\vspace{8pt}}
 \methodcell{\radpath}{tropical_beach_scale0_5}
 & \methodcell{\radpath}{tropical_beach_scale2_0}
 & \methodcell{\radpath}{tropical_beach_scale4_0} \\
 \end{tabular}%
 }
 \caption{\textbf{RoPE scale ablation.} We vary the radius scale $s$ in $R_{\text{width}} = W/s$, which controls the range of positional values fed to RoPE. \textbf{Our default} ($s{=}2.0$) matches the expected input distribution, preserving both global geometry and local detail.}
 \label{fig:radius_ablation}
\end{figure*}
\paragraph{SpheRoPE Radius Scale Ablation.} The spherical encoding path in SpheRoPE (Eq.~\ref{eq:spherical_rope}) uses a radius $R_{\text{width}} = W/s$, where $s$ is a scale hyperparameter that controls the range of positional values fed to the pre-trained RoPE. Since the model was trained on a specific range of positional values (roughly $[0, W]$), this scale directly affects whether the RoPE frequencies operate in their learned regime. 
Figure~\ref{fig:radius_ablation} shows the failure modes at both extremes. 
Our default ($s{=}2.0$) matches the expected input distribution of the pre-trained model, producing panoramas with both coherent global structure and fine-grained texture.

\newcommand{\geopath}{figures/ablation_geo_prompt}

\begin{figure*}
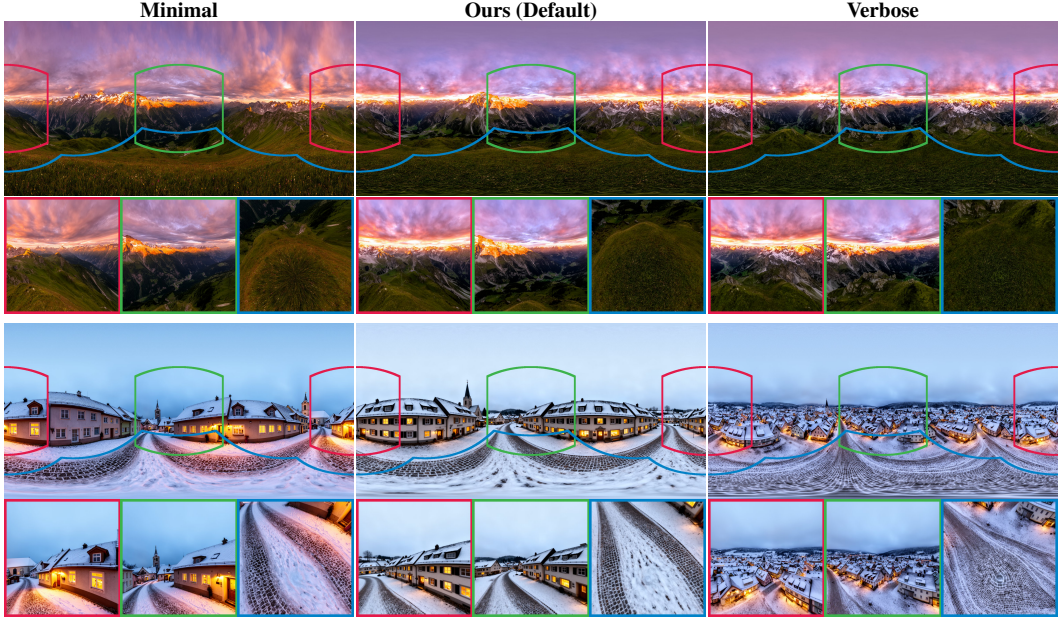

 \centering
 \setlength{\tabcolsep}{1pt}
 \resizebox{\linewidth}{!}{%
 \begin{tabular}{ccc}
 {\LARGE\textbf{Minimal}} & {\LARGE\textbf{Ours (Default)}} & {\LARGE\textbf{Verbose}} \\
 \methodcell{\geopath}{mountain_sunset_minimal}
 & \methodcell{\geopath}{mountain_sunset_default}
 & \methodcell{\geopath}{mountain_sunset_verbose} \\
 \noalign{\vspace{8pt}}
 \methodcell{\geopath}{snowy_village_minimal}
 & \methodcell{\geopath}{snowy_village_default}
 & \methodcell{\geopath}{snowy_village_verbose} \\
 \end{tabular}%
 }
 \caption{\textbf{Geometric prompt ablation.} We vary the geometric prompt $\mathbf{p}_{\text{geo}}$ used in Semantic Distortion CFG. The \textbf{minimal} prompt (``equirectangular panorama'') under-specifies
the desired geometry, producing flat panoramas with pole artifacts. The \textbf{verbose} prompt over-specifies, pushing the model toward extreme top-down curvature. \textbf{Our default} balances these extremes,
yielding valid ERP geometry with natural perspective.}
 \label{fig:geo_prompt_ablation}
\end{figure*}
\paragraph{Geometric Prompt Ablation.} Semantic Distortion CFG is steered by a geometric prompt $\mathbf{p}_{\text{geo}}$ that describes the desired ERP properties. Figure~\ref{fig:geo_prompt_ablation} shows the effect of varying this prompt. A minimal prompt (``equirectangular panorama'') under-specifies the desired geometry, and the model defaults to flat panoramas with visible pole artifacts. A verbose prompt, enumerating every panoramic property, over-constrains the generation and produces extreme top-down curvature. Our default prompt strikes a balance, providing enough specificity to enforce valid ERP geometry without distorting the natural perspective of the scene.

\begin{figure}
\centering
\setlength{\tabcolsep}{2pt}
\begin{tabular}{cc}
\includegraphics[width=0.48\linewidth]{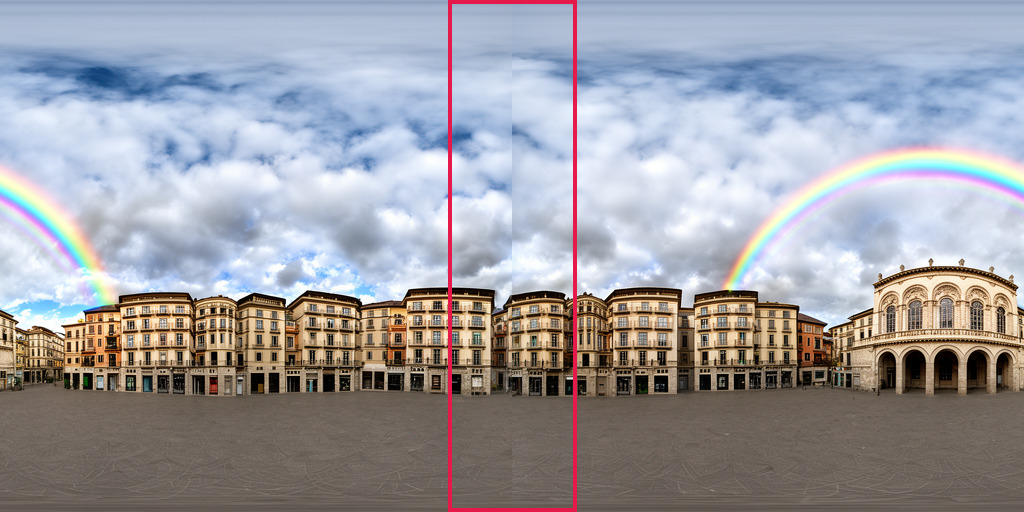} &
\includegraphics[width=0.48\linewidth]{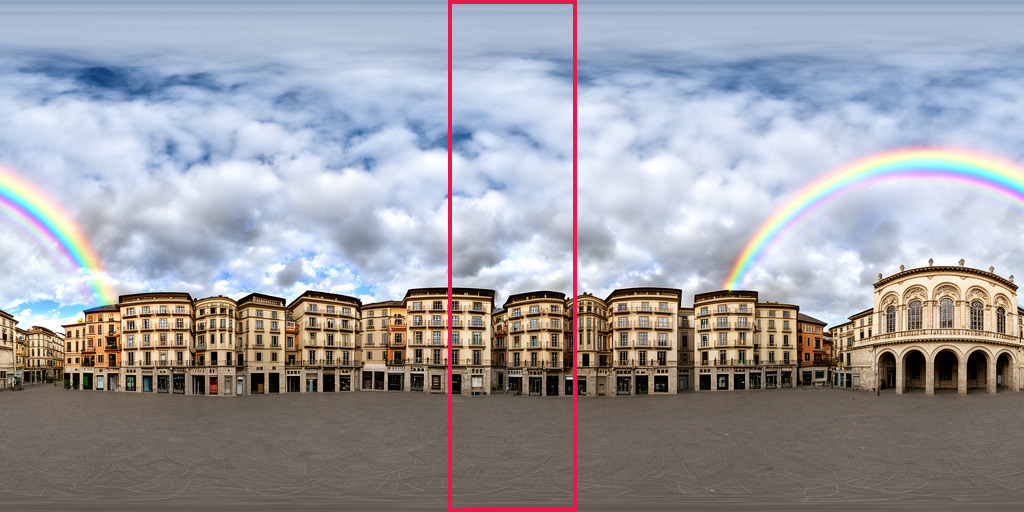} \\
{\small (a) Without circular padding} & {\small (b) With circular padding}
\end{tabular}
\caption{\textbf{Circular latent encoding ablation.} Panoramas are shifted by 180° to place the wrap boundary at the center of the frame.(a)~Without circular VAE padding: visible seam appears at the boundary. (b)~With circular padding: the seam is eliminated.}
\label{fig:ablation_circular}
\end{figure}
\paragraph{Circular Latent Encoding.}
Figure~\ref{fig:ablation_circular} shows the effect of disabling circular VAE padding. Without it, the VAE's zero-padded convolutions introduce a latent-space discontinuity at the horizontal boundary that manifests as a visible vertical seam in the decoded output. Enabling circular padding eliminates this artifact entirely, producing seamless wrap-around at no additional inference cost. 

  % \begin{table}[t]
  % \centering
  % \caption{\textbf{Semantic CFG Scale.} We vary the guidance scale $w_{sem}$.
  % Best in \textbf{bold}, second best \underline{underlined}.}
  % \label{tab:cfg_scale}
  % \resizebox{\columnwidth}{!}{%
  % \begin{tabular}{l|ccc|cccc}
  % & \multicolumn{3}{c|}{\textbf{Panorama-Level Metrics}} & \multicolumn{4}{c}{\textbf{Multi-View Metrics}} \\
  % \textbf{Configuration} & FAED $\downarrow$ & OmniFID $\downarrow$ & DS $\downarrow$ & FID $\downarrow$ & KID$_{\times 10^2}$ $\downarrow$ & IS $\uparrow$ & CS $\uparrow$ \\
  % \midrule
  % $w_{sem} = 2$ & 40.60 & 86.27 & \textbf{0.80} & 28.96 & 1.15 & 12.60 & \textbf{18.45} \\
  % $w_{sem} = 3$ & \underline{36.34} & \underline{83.69} & \underline{0.82} & \underline{28.53} & \underline{1.06} & 12.77 & 18.41 \\
  % $w_{sem}{=}4$ (Ours) & \textbf{25.40} & \textbf{80.66} & 0.94 & \textbf{25.33} & \textbf{0.90} & 12.75 & \underline{18.43} \\
  % $w_{sem} = 5$ & 36.64 & 83.50 & 0.83 & 29.67 & 1.22 & \textbf{12.88} & 18.37 \\
  % \bottomrule
  % \end{tabular}%
  % }
  % \end{table}

    \begin{table}[t]
  \centering
  \caption{\textbf{Semantic CFG Scale.} We vary the guidance scale $w_{sem}$.
  Best in \textbf{bold}, second best \underline{underlined}.}
  \label{tab:cfg_scale}
  \resizebox{\columnwidth}{!}{%
  \begin{tabular}{l|cc|cccc}
  & \multicolumn{2}{c|}{\textbf{Panorama-Level Metrics}} & \multicolumn{4}{c}{\textbf{Multi-View Metrics}} \\
  \textbf{Configuration} & FAED $\downarrow$ & DS $\downarrow$ & FID $\downarrow$ & KID$_{\times 10^2}$ $\downarrow$ & IS $\uparrow$ & CS $\uparrow$ \\
  \midrule
  $w_{sem} = 2$ & 40.60 & \textbf{0.80} & 28.96 & 1.15 & 12.60 & \textbf{18.45} \\
  $w_{sem} = 3$ & \underline{36.34} & \underline{0.82} & \underline{28.53} & \underline{1.06} & 12.77 & 18.41 \\
  $w_{sem}{=}4$ (Ours) & \textbf{25.40} & 0.94 & \textbf{25.33} & \textbf{0.90} & 12.75 & \underline{18.43} \\
  $w_{sem} = 5$ & 36.64 & 0.83 & 29.67 & 1.22 & \textbf{12.88} & 18.37 \\
  \bottomrule
  \end{tabular}%
  }
  \end{table}
\paragraph{Semantic CFG Scale.} We analyze the sensitivity of the semantic guidance scale $w_{sem}$ while maintaining all geometric components (Table~\ref{tab:cfg_scale}). Lower values ($w_{sem} = 2$) yield slightly better structural continuity (DS) and marginal gains in text alignment (CS), but significantly underperform in distributional image quality (higher FAED, and FID). Our default ($w_{sem} = 4$) achieves the optimal balance, yielding the highest overall image quality and panoramic fidelity across both panorama-level and multi-view metrics. Higher values ($w_{sem} = 5$) lead to regression in both visual quality and prompt adherence, suggesting that overly aggressive semantic guidance disrupts the balance between content fidelity and geometric validity.

\clearpage
\section{Geometric Prompts}
\label{sec:supp_prompts}

As described in Sec.~3.3 of the main paper, our Semantic Distortion CFG uses an anchored geometric prompt $\mathbf{p}_{\text{anchor}} = [\mathbf{p};\, \mathbf{p}_{\text{geo}}]$ to steer the denoising process toward valid ERP geometry. Below we list the fixed geometric prompts $\mathbf{p}_{\text{geo}}$ used across all experiments.

\paragraph{Image generation (Flux~1, Flux~2).}
\begin{quote}
\small\textit{``Single unified continuous environment, monolithic scene composition, solitary spatial layout, flawlessly stitched 360 panorama, true equirectangular projection, accurate spherical geometry, continuous horizontal wrap, zero parallax error.''}
\end{quote}

\paragraph{Video generation (LTX~2.3).}
\begin{quote}
\small\textit{``True 2:1 equirectangular projection, proper zenith/nadir pole geometry, seamless 360° horizontal wrap. Rigidly locked static tripod camera at a fixed nodal point. All geometry is permanently static: buildings, terrain, walls, floors, ceilings, roads, vegetation trunks, rocks, and all man-made structures maintain absolute pixel-locked positions across every frame. Zero structural deformation, zero background warping, zero surface drift. Only permitted motion: faint atmospheric haze, subtle light caustic shifts, microscopic dust particles. Flawless inter-frame coherence.''}
\end{quote}

The video prompt is longer than the image prompt because it must additionally enforce temporal stability: the camera must remain static and all scene geometry must be pixel-locked across frames, with only minimal atmospheric motion permitted. This prevents the video model from introducing camera movement or structural warping that would break the ERP constraints.

\subsection{LLM-as-a-Judge Evaluation Prompt}
\label{sec:supp_llm_prompt}

For the LLM-based evaluation reported in Sec.~4 of the main paper, we use the exact prompt introduced by SphereDiff~\cite{park2025spherediff} (Tab.~5 of their paper), which we reproduce verbatim in Tab.~\ref{tab:llm_prompt} for completeness. The prompt elicits a brief chain-of-thought~\cite{wei2022cot} justification before each score.

\begin{table}[h]
\caption{\textbf{Evaluation Prompt for VLM.} The evaluation prompt used to assess panorama quality based on four image and panoramic criteria, taken verbatim from SphereDiff~\cite{park2025spherediff}. We instruct the VLM~\cite{hurst2024gpt4o} to provide a score along with a brief reason to encourage chain-of-thought~\cite{wei2022cot}.}
\label{tab:llm_prompt}
\centering
\small
\begin{tabular}{|p{0.95\linewidth}|}
\hline
\textbf{Evaluation Prompt} \\
\hline
You are an evaluator assessing an image generation model based on a single image at a time. Your evaluation is based on the following four criteria: \newline
1. \textbf{Image Quality}: Assess the overall quality of the image. \newline
2. \textbf{Aesthetic Appeal}: Evaluate how visually pleasing the image is. \newline
3. \textbf{Distortion Level}: Determine whether the image appears distorted. If it does not resemble a photo taken with a normal camera, it will receive a lower score. \newline
4. \textbf{Connectivity}: Check if the middle of the image appears disconnected. If there is a noticeable break, the score will be lower. \newline
Each criterion is rated on a five-point scale: Excellent (5), Good (4), Fair (3), Poor (2), and Awful (1). \newline
You will receive one image at a time. For each criterion, provide a concise reason for the score before listing the rating. Format your response as follows: \newline
- \textbf{Image Quality}: (Brief reason) $\rightarrow$ Score \newline
- \textbf{Aesthetic Appeal}: (Brief reason) $\rightarrow$ Score \newline
- \textbf{Distortion Level}: (Brief reason) $\rightarrow$ Score \newline
- \textbf{Connectivity}: (Brief reason) $\rightarrow$ Score \newline
\{image\} \newline
Please evaluate the image with the given criteria. \\
\hline
\end{tabular}
\end{table}

\subsection{Stress-20 Benchmark Construction}
\label{sec:supp_stress20}

\paragraph{Construction protocol.}
To complement the SphereDiff-20 benchmark~\cite{park2025spherediff} with prompts that specifically target temporal coherence under challenging dynamics, we constructed the Stress-20 prompt set using the following protocol.
We provided Claude Opus~4.7~\cite{anthropic2026opus47} with the 20 SphereDiff-20 scene descriptions as reference examples of typical panoramic video prompts, and instructed it to generate 20 \emph{new}, diverse single-line prompts that stress-test $360^\circ$ video generation with high-motion content, rapid camera-implied dynamics, multiple interacting subjects, and complex environmental effects (e.g., explosions, weather, crowds).
The exact generation prompt was:
\begin{quote}
\small\textit{``Given these 20 panoramic video scene descriptions as reference for style and format, generate 20 new and diverse single-paragraph prompts (each 30--60 words) designed to stress-test 360-degree panoramic video generation. Focus on: rapid full-sphere motion, multiple interacting subjects with independent trajectories, dynamic lighting changes, particle effects (fire, water, debris), and scenes where temporal coherence is difficult to maintain. Each prompt should describe a different environment and action. Do not repeat or paraphrase the reference prompts.''}
\end{quote}
The 20 returned prompts (shown in Table~\ref{tab:stress_prompts}) were used verbatim without any filtering, reordering, or modification.
Crucially, the prompts were generated and finalized \emph{before} any method was evaluated on them, no prompts were added, removed, or edited after observing any method's outputs.
\begin{table*}[t]
   \centering
   \caption{\textbf{Stress-20 prompt set.} Twenty prompts designed to stress-test $360^\circ$ video generation with challenging motion, parallax, occlusions, and lighting transitions.}
   \label{tab:stress_prompts}
   \footnotesize
   \setlength{\tabcolsep}{6pt}
   \renewcommand{\arraystretch}{1.15}
   \begin{tabular}{@{}c p{0.92\linewidth}@{}}
   \toprule
   \textbf{\#} & \textbf{Prompt} \\
   \midrule
    1 & Rock concert in a packed $360^\circ$ arena, camera moving forward through the crowd toward the stage, people jumping and pushing past the camera, flashing strobe lights, smoke and confetti, performers
   moving rapidly across the full field of view with slight handheld roll and tilt. \\
    2 & Formula 1 race track from trackside perspective, camera moving laterally along the barrier while cars rush past at high speed in both directions, strong motion blur, rapid depth changes, vehicles crossing
   the full $360^\circ$ view. \\
    3 & Crowded subway platform, camera walking forward along the platform edge, trains arriving and departing, doors opening, dense crowd crossing left and right across the entire panorama, flickering overhead
   lights. \\
    4 & Busy outdoor marketplace with narrow aisles, camera weaving forward between stalls, people passing very close to the lens, fabrics and flags moving in the wind, sunlight shifting through overhead covers. \\
    5 & Dense forest trail, camera moving forward at walking speed, trees passing close on both sides, strong parallax between near trunks and distant background, sunlight flickering through leaves. \\
    6 & Underwater coral reef scene, camera drifting forward and slightly tilting, schools of fish swirling in all directions, particles floating, dynamic light caustics moving across the environment. \\
    7 & Stormy ocean deck of a ship, camera moving with the motion of the waves, strong vertical and lateral sway, water splashing across the scene, rain and lightning causing rapid lighting changes. \\
    8 & Urban city street at night after rain, camera moving forward along the sidewalk, pedestrians passing in both directions, neon reflections shifting on wet ground, cars driving past across the full view. \\
    9 & Airport terminal interior, camera moving forward through a crowd, travelers with luggage crossing in front of and behind the camera, escalators moving, dynamic occlusions and reappearances. \\
   10 & Large train station hall with repeating arches, camera dollying forward while slightly shifting sideways, people crossing symmetry lines, strong structural geometry with moving crowds. \\
   11 & Office interior with glass walls and corridors, camera walking through hallways, reflections on glass surfaces, people entering and exiting rooms, multiple occlusions and reappearances. \\
   12 & Library with tall dense bookshelves, camera sliding sideways while moving forward, strong parallax from near shelves and distant aisles, people occasionally crossing between rows. \\
   13 & Driving through a forest road from inside a car, camera moving forward with slight natural shake, trees passing quickly on both sides, sunlight flickering through branches. \\
   14 & Narrow cave interior transitioning to bright outdoor exit, camera moving forward, exposure adapting from dark to bright, irregular geometry and depth changes, turning slightly to look around. \\
   15 & Construction site with heavy machinery, camera moving through the site, cranes rotating, workers walking, dust and debris in the air, multiple moving objects at different depths. \\
   16 & Snowy mountain slope with skiers, camera moving downhill, people passing at different speeds and distances, snow particles flying, strong depth variation and motion. \\
   17 & Crowded street protest, camera walking through dense moving crowd, flags waving, smoke flares, people crossing the full panorama, chaotic motion and occlusions. \\
   18 & Theme park roller coaster POV, camera moving rapidly forward with turns and drops, fast changing depth, structures passing close to the camera, dynamic lighting transitions. \\
   19 & Shopping mall interior with multiple floors, camera moving along a corridor while looking around slightly, escalators and people moving at different levels, reflections and glass surfaces. \\
   20 & Beach shoreline with waves and people, camera walking along the waterline, waves moving in and out, people running and crossing near and far, strong motion in both foreground and background. \\
   \bottomrule
   \end{tabular}
   \end{table*}

\paragraph{Cross-set ranking agreement.}
To verify that Stress-20 does not selectively favor our method, we compare method rankings across the two prompt sets.
The ranking pattern remains stable: our method ranks first on all 6 VBench metrics on SphereDiff-20 and on 5/6 metrics on Stress-20.
The only exception is CLIP Mean on Stress-20, where DynamicScaler surpasses our method and we rank second, indicating that the self-designed benchmark does not uniformly inflate our scores.

\subsection{Prompt Format Conversion for Video Evaluation}
\label{sec:supp_prompt_conversion}

As described in Sec.~\ref{sec:video_eval}, our video evaluation spans two prompt sets - SphereDiff-20 and Stress-20 - and several methods that expect incompatible conditioning formats. 
Specifically, while our approach, along with training-based approaches expect a single semantic prompt, optimization-based methods (such as SphereDiff~\cite{park2025spherediff} and DynamicScaler~\cite{liu2025dynamicscaler}) require a per-elevation 5-line format with separate per-view conditioning.
To bring them onto a common ground without handicapping any method at generation time, we use two LLM-driven prompt conversions: a \textbf{5\,$\rightarrow$\,1} consolidation (for single-prompt methods) and a \textbf{1\,$\rightarrow$\,5} expansion (to lift Stress-20 single-paragraph prompts into SphereDiff's canonical per-elevation format). 
Importantly, our method and all training-based baselines receive only the consolidated single prompt, which necessarily discards some per-elevation detail, while the richer 5-line format is provided only to the optimization-based baselines that require it.

\paragraph{5\,$\rightarrow$\,1 consolidation.}
For the 5\,$\rightarrow$\,1 direction, we use GPT-4o~\cite{hurst2024gpt4o} to consolidate the five elevation lines into a single prompt. The exact consolidation prompt is given in Tab.~\ref{tab:prompt_5to1}.
For 360DVD~\cite{wang2024360dvd}, whose CLIP text encoder truncates inputs at 77 tokens, we further shorten the consolidated prompt to approximately 25 words using GPT-4o. The exact prompt is given in Tab.~\ref{tab:prompt_summarize}.

\begin{table}[h]
\caption{\textbf{5\,$\rightarrow$\,1 prompt consolidation.} Used to convert a 5-line per-elevation prompt into a single-paragraph prompt for methods that accept only one prompt (e.g., LTX~2.3).}
\centering
\small
\begin{tabular}{|p{0.95\linewidth}|}
\hline
\textbf{5\,$\rightarrow$\,1 Consolidation Prompt} \\
\hline
You are a prompt rewriter for a text-to-video model. You will receive a 360\textdegree\ panoramic scene described across five elevation-specific lines: \newline
1. \textbf{Top (skyward view)}: what is visible looking straight up. \newline
2. \textbf{Above the horizon}: what is visible in the upper hemisphere. \newline
3. \textbf{Eye level (horizon)}: the main subject and scene at eye level. \newline
4. \textbf{Below eye level}: what is visible in the lower hemisphere. \newline
5. \textbf{Bottom (ground-facing view)}: what is visible looking straight down. \newline
Your task is to consolidate these five lines into a single coherent paragraph (approximately 40-60 words) that preserves the main subject, the dominant camera motion, and the overall scene atmosphere. Prioritize the eye-level description as the anchor of the consolidated prompt. Drop per-region detail when it does not fit naturally into a single-paragraph description. Do not invent content that is not implied by the five input lines. Do not include section labels, bullet points, or the words ``top'', ``above'', ``below'', or ``bottom'' in the output. \newline
\{top\} \newline
\{above\} \newline
\{eye\} \newline
\{below\} \newline
\{bottom\} \newline
Return only the consolidated paragraph. \\
\hline
\end{tabular}
\label{tab:prompt_5to1}
\end{table}

\begin{table}[h]
\caption{\textbf{Prompt summarization for 360DVD.} Used to condense the consolidated prompt to $\leq$25 words for compatibility with CLIP's 77-token limit.}
\centering
\small
\begin{tabular}{|p{0.95\linewidth}|}
\hline
\textbf{Summarization Prompt} \\
\hline
You are a prompt rewriter for a text-to-video model with a strict token limit. You will receive a scene description for a 360\textdegree\ panoramic video. Your task is to rewrite it as a single sentence of at most 25 words that preserves the main subject, setting, and atmosphere. Do not add any detail not present in the input. Do not use bullet points or labels. \newline
\{prompt\} \newline
Return only the shortened sentence. \\
\hline
\end{tabular}
\label{tab:prompt_summarize}
\end{table}

\paragraph{1\,$\rightarrow$\,5 expansion.}
We use GPT-4o~\cite{hurst2024gpt4o} to expand Stress-20 single-paragraph prompts into the per-elevation 5-line format required by SphereDiff~\cite{park2025spherediff} and DynamicScaler~\cite{liu2025dynamicscaler}. The exact expansion prompt is given in Tab.~\ref{tab:prompt_1to5}.

\begin{table}[h]
\caption{\textbf{1\,$\rightarrow$\,5 prompt expansion.} Used to convert Stress-20 single-paragraph prompts into the per-elevation 5-line format used by SphereDiff~\cite{park2025spherediff} and DynamicScaler~\cite{liu2025dynamicscaler}.}
\centering
\small
\begin{tabular}{|p{0.95\linewidth}|}
\hline
\textbf{1\,$\rightarrow$\,5 Expansion Prompt} \\
\hline
You are a prompt rewriter for 360\textdegree\ panoramic video generation. You will receive a single-paragraph scene description. Your task is to expand it into five coherent, elevation-specific lines that together describe what a viewer would see across the full sphere at a fixed vantage point. The five lines must be plausibly consistent with each other and with the input paragraph, and must not introduce content that contradicts the input. \newline
Produce exactly the following five lines, each 1-2 sentences: \newline
1. \textbf{Top (skyward view)}: what is visible looking straight up at the zenith. \newline
2. \textbf{Above the horizon}: what fills the upper hemisphere between the horizon and the zenith. \newline
3. \textbf{Eye level (horizon)}: the main subject and scene at eye level; preserve the dominant motion. \newline
4. \textbf{Below eye level}: what fills the lower hemisphere between the horizon and the nadir. \newline
5. \textbf{Bottom (ground-facing view)}: what is visible looking straight down at the nadir. \newline
Keep all five lines mutually consistent in setting, lighting, time of day, and style. Do not use the labels in the output; return only the five descriptions, one per line. \newline
\{paragraph\} \\
\hline
\end{tabular}

\label{tab:prompt_1to5}
\end{table}

\subsection{User Study Protocol}
\label{sec:supp_user_study}

We conduct a blind pairwise preference study to evaluate perceptual quality across the full sphere.
Each trial presents two anonymized panoramas in interactive $360^\circ$ viewers with drag-to-pan navigation, allowing raters to freely explore the entire sphere rather than judge a single static viewpoint.
The two viewers are synchronized so that both display the same region simultaneously, ensuring direct comparison.
Navigation presets allow raters to jump to the seam boundary, poles, and front view, guiding inspection of known failure modes.
For each pair, raters make a three-way choice (A, B, or tie) on two questions: (1)~``Which panorama do you prefer?'' (overall quality) and (2)~``Which one has better text alignment?''.
The left-right presentation order is position-balanced to mitigate side bias.

We use 20 text prompts and 6 baselines (DIT360, PanFusion, PAR, SMGD, SphereDiff, and UniPano).
Each rater evaluates a subset of pairs with no repeated anchor images to avoid familiarity bias.
We collect 320 pairwise judgments from 18 annotators with no filtering applied.
% Results are reported in Table~\ref{tab:user_study}.

\clearpage
\section{Limitations and Future Work}
\label{limitations}

\begin{figure*}[t]
\centering
\begin{minipage}{0.49\textwidth}
\centering
\includegraphics[width=\textwidth]{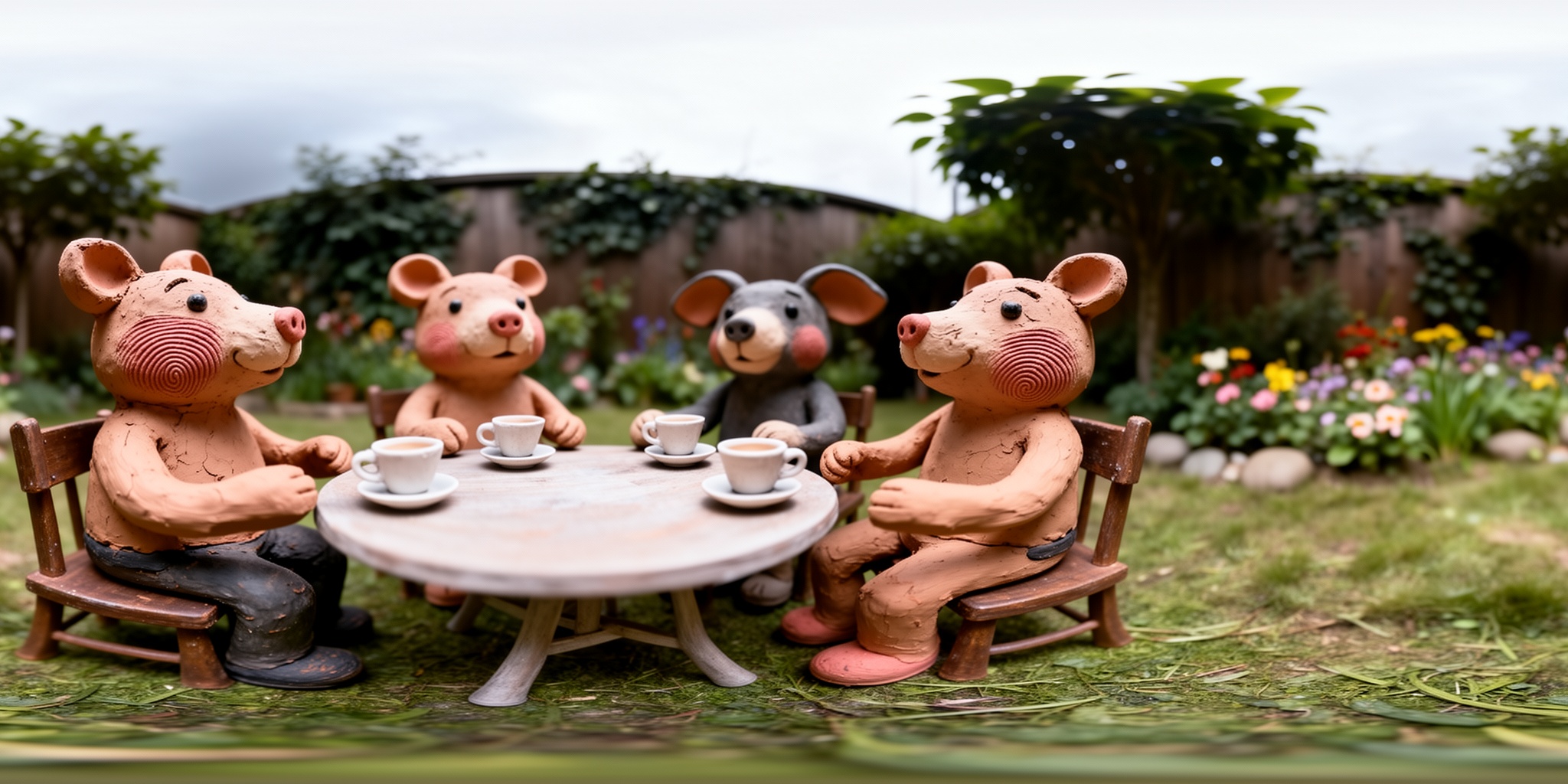}
\smallskip

{\small \textbf{(a)} ``... claymation-style garden party... soft diffused studio lighting and a shallow depth of field.''}
\end{minipage}
\hfill
\begin{minipage}{0.49\textwidth}
\centering
\includegraphics[width=\textwidth]{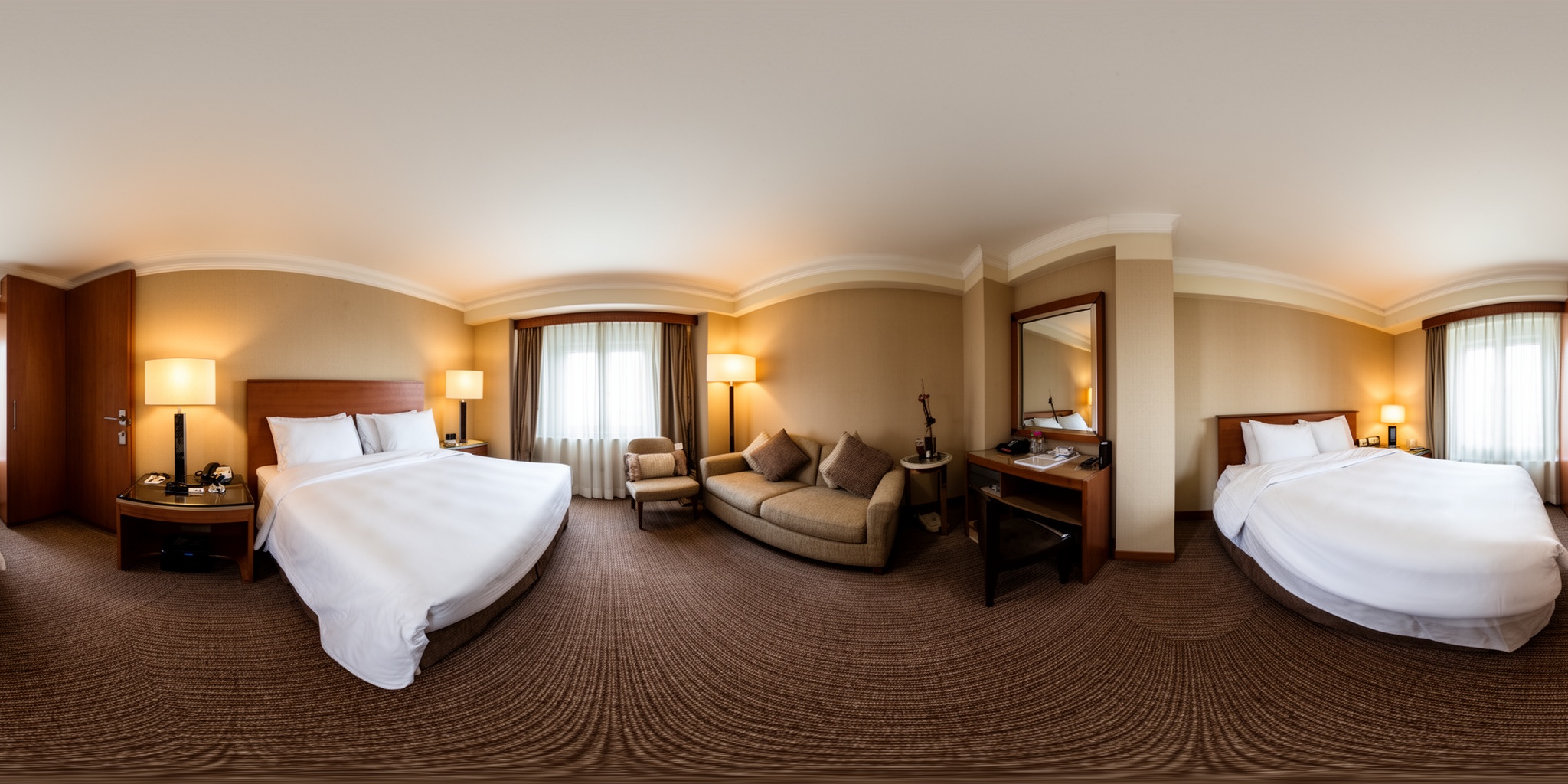}
\smallskip

{\small \textbf{(b)} ``A bedroom in a hotel room.''}
\end{minipage}
\caption{\textbf{Failure cases.} (a)~Prompts with strong perspective priors (e.g., studio lighting, shallow depth of field) can produce perspective-like outputs rather than true panoramas. (b)~The model may repeat structural elements to fill the full sphere.}
\label{fig:failures}
\end{figure*}

\paragraph{Failure Cases.}
We illustrate representative failure modes in Figure~\ref{fig:failures}. For prompts with strong perspective priors (e.g., front-facing compositions), the model may default to generating a conventional perspective image rather than a true equirectangular panorama. Additionally, the model may occasionally repeat structural elements to fill the full 360° field of view.

\paragraph{Limitations.}
Our method relies on two architectural assumptions. First, the backbone must use Rotary Position Embeddings (RoPE) for spatial encoding, as Spherical RoPE directly modifies the width-axis rotation angles. Architectures using other positional encoding schemes (e.g., learned absolute embeddings or additive sincos) would require a different adaptation strategy. Second, the pre-trained model must have been exposed to panoramic or ERP-like content during training. Our approach amplifies and steers latent panoramic priors that already exist in the model's distribution. But, if the backbone has no such priors, Spherical RoPE and Semantic Distortion CFG alone cannot induce panoramic generation from scratch. Representative residual failure modes are shown in Figure~\ref{fig:failures}. Additionally, our method inherits the resolution and quality ceiling of the backbone model. 
Furthermore, our extension to video generation is currently restricted by motion constraints. To prevent the model from breaking the enforced ERP topology across frames, we must rely on rigid prompting to ensure a strictly static camera and limit dynamics to minimal atmospheric motion. Consequently, generating 360$^\circ$ videos with complex subject movement or camera trajectories remains an open challenge.
Finally, because Semantic Distortion CFG computes three noise predictions per denoising step, it entails a $1.5\times$ increase in Network Function Evaluations (NFE) compared to standard CFG. This introduces a computational trade-off, exchanging inference-time efficiency for strict geometric adherence.

\paragraph{Broader impact.} By eliminating the need for model fine-tuning, our approach significantly reduces the computational burden and energy consumption typically associated with adapting large-scale diffusion models. This training-free paradigm not only lowers the carbon footprint of 360° content creation but also democratizes access, enabling independent researchers and smaller studios to generate high-quality immersive environments without requiring massive GPU clusters. However, as with all generative models, there is a potential for misuse in creating fabricated immersive content. We inherit the safety guardrails of the underlying backbone models and do not introduce new capabilities for generating harmful or deceptive content beyond what the base models already permit.

\paragraph{Future work.}
Our current audio-video generation produces mono audio natively output by LTX~2.3, with no spatial awareness. A natural extension is omnidirectional audio-video generation, where the audio is spatialized to match the $360^\circ$ visual content, enabling fully immersive audiovisual experiences. More broadly, the principle of reshaping RoPE to encode non-Euclidean geometry is not limited to the sphere - it could be extended to other domains such as cylindrical projections, hyperbolic spaces, or arbitrary manifolds, opening new directions for geometry-aware generation beyond panoramas.

%%%%%%%%%%%%%%%%%%%%%%%%%%%%%%%%%%%%%%%%%%%%%%%%%%%%%%%%%%%%

\end{document}